%% file: preprint.tex
\documentclass[11pt]{article}
\usepackage[preprint]{acl}
\usepackage{times,latexsym}
\usepackage[T1]{fontenc}
\usepackage[utf8]{inputenc}
\usepackage{microtype,inconsolata}
\usepackage{graphicx,xcolor}
\usepackage{amsmath,amssymb,mathtools,amsthm}
\usepackage{algorithm,algorithmic}
\usepackage{paralist,booktabs,array,tabularx}
\usepackage{placeins}
\usepackage[capitalize,noabbrev]{cleveref}
\input{math_commands.tex}

\title{Forget About It: Mitigating Forgetting \\ Without Replay or Auxiliary Objectives}
\author{
  Anat Kleiman\textsuperscript{1}\thanks{Correspondence to:
\texttt{anatkleiman@g.harvard.edu}}\quad
  Gintare Karolina Dziugaite\textsuperscript{2}\quad
  Jonathan Frankle\textsuperscript{3}\\
  \textbf{Sham Kakade\textsuperscript{1}\quad
  Mansheej Paul\textsuperscript{3}\thanks{Work done while at Databricks.}}\\
  \textsuperscript{1}Harvard University, Kempner Institute\\
  \textsuperscript{2}Google DeepMind\quad
  \textsuperscript{3}Databricks
}
\colorlet{blue}{black}
\newcommand{\coderepolink}{https://github.com/a-kleiman/forget-about-it}
\hypersetup{hidelinks,pdfauthor={Anat Kleiman, Gintare Karolina Dziugaite, Jonathan Frankle, Sham Kakade, Mansheej Paul},pdftitle={Forget About It: Mitigating Forgetting Without Replay or Auxiliary Objectives}}
\begin{document}

\maketitle

\input{body}
\FloatBarrier
\section*{Acknowledgements}

This work has been made possible in part by a gift from the
Chan Zuckerberg Initiative Foundation to establish the Kempner
Institute for the Study of Natural and Artificial Intelligence.
Thank you to Ben Deaner and Vincent Dumoulin for reading earlier versions of the draft and offering suggestions.
\bibliography{main}
\clearpage
\appendix
\raggedbottom
\input{appendix}

\end{document}

%% file: math_commands.tex
\usepackage{amsmath,amsfonts,bm}

\def\eqref#1{equation~\ref{#1}}
\def\1{\bm{1}}

\DeclareMathAlphabet{\mathsfit}{\encodingdefault}{\sfdefault}{m}{sl}
\SetMathAlphabet{\mathsfit}{bold}{\encodingdefault}{\sfdefault}{bx}{n}

%% file: body.tex
\begin{abstract}

Have we made mitigating catastrophic forgetting more complicated than it needs to be? Continual adaptation often degrades performance on earlier tasks, motivating rehearsal buffers, auxiliary objectives, and adaptive merging procedures. We investigate how far a simpler alternative-periodic parameter interpolation during fine-tuning-can go on its own. We introduce Sequential Fine-tuning with Averaging (SFA) and evaluate it alongside replay-free Lookahead, originally designed for optimization stability to find compelling results. In Qwen3-14B adapter continual learning, fixed configurations of both methods outperform the tested replay and adaptive-merging configurations in mean final accuracy using up to 36.3\% less task training time. Furthermore, across language and vision tasks, our results compete with more complex continual-learning methods. Together, these findings suggest that simple in-training interpolation deserves consideration before introducing more elaborate forgetting-mitigation mechanisms. 
Our code is available \href{\coderepolink}{here}.

\end{abstract}

\section{Introduction}\label{intro}

When a language model is fine-tuned on successive tasks,
its performance on earlier tasks can deteriorate
\citep{luo2023empirical,bai2025efficient,wang2025see}. This can affect both earlier fine-tuning tasks and more general capabilities \citep{liu2024generalcapabilities,wu2025translation}.
As models are repeatedly adapted to new domains, retaining those capabilities without substantially increasing training cost remains an important challenge.

One response is rehearsal: keeping earlier examples and revisiting them during training \citep{doi:10.1080/09540099550039318,lopezpaz2022gradient,dautume2019episodic}. Other approaches incorporate consistency objectives, memories of previous models, or adaptive merging procedures \citep{arani2022learningfastlearningslow,feng2025aim}. These mechanisms can be effective, but add storage, gradient computation, monitoring, or implementation complexity. This raises a more basic question: \emph{how much of their benefit can be obtained by simply averaging parameters during training?} Unlike endpoint merging \citep{wortsman2022robust,ilharco2023editing,kozal2024continuallearningweightinterpolation,marczak2024magmaxleveragingmodelmerging}, intermediate averaging can also change the subsequent optimization trajectory.

We investigate two lightweight forms of in-training interpolation. The first is \textbf{Lookahead}, originally introduced for optimization stability rather than sequential-task forgetting \citep{zhang2019lookahead}. It periodically interpolates the working weights with an evolving slow reference and resumes training. We also introduce the second, \textbf{Sequential Fine-tuning with Averaging (SFA)}, which instead interpolates the working model with a checkpoint from an earlier task that remains fixed throughout. Both methods require only a parameter reference and scalar mixing operations, without replay gradients or auxiliary losses. Comparing them allows us to study the role of the reference policy, while evaluating them alongside more elaborate continual-learning approaches lets us examine whether their additional mechanisms are always warranted.

We evaluate Math-to-Law and Math-to-Code domain shifts, language-task streams under full-parameter and LoRA fine-tuning, and longer image-classification sequences. In Qwen3-14B adapter continual learning, fixed configurations of both SFA and Lookahead achieve higher mean final accuracy than the more complicated baselines. SFA also reduces measured training-loop time relative to replay-based methods. As such, our results suggest that simple, carefully specified interpolation baselines should be considered for forgetting-mitigation before more complex alternatives. 

To summarize, our contributions are:
\begin{compactitem}
\item We evaluate \textbf{replay-free in-training interpolation} as a practical forgetting-mitigation strategy, including an application of the original Lookahead update to supervised continual language-model fine-tuning. We find that lightweight averaging can rival the tested rehearsal and adaptive-fusion configurations without replay or auxiliary objectives (Section~\ref{sec:adapterresults}).
\item We introduce \textbf{SFA} as a fixed-reference counterpart to moving-reference Lookahead and show how continuing training after averaging can improve upon endpoint merging. We examine fixed and moving references without assuming either is uniformly better (Sections~\ref{averaging:continual} and~\ref{sec:newdomain}).
\item We characterize the \textbf{accuracy-cost and retention-adaptation trade-offs} across domain shifts, task streams, model sizes, and averaging schedules, including cases where interpolation does not outperform rehearsal or tuned alternatives (Sections~\ref{sec:adapterresults}--\ref{sec:analysis}; Appendix~\ref{app:coefficient-frequency}).
\end{compactitem}

\section{Related Work}\label{sec:related}

\paragraph{Continual learning.}

Rehearsal mitigates forgetting by continuing to train on earlier examples \citep{dautume2019episodic,prabhu2023computationallybudgetedcontinuallearning}. Another line of work constrains the parameters to remain close to earlier solutions; EWC, for example, weights this constraint by estimated parameter importance \citep{Kirkpatrick_2017}. SFA shares the intuition of preserving an earlier solution, but incorporates it through periodic interpolation of the working model.

\paragraph{Averaging during training.}

Averaging appears within continual-learning systems: CLS-ER combines fast and slow EMA memories with rehearsal and prediction consistency \citep{arani2022learningfastlearningslow}; ESMER, SCoMMER, IMEX-Reg, and DUCA incorporate related memory mechanisms with replay or additional objectives \citep{sarfraz2023errorsensitivitymodulationbased,sarfraz2022sparsecodingdualmemory,bhat2024imexregimplicitexplicitregularizationfunction,gowda2023dualcognitivearchitectureincorporating}. Our question is what remains useful when such additions are removed. Lookahead \citep{zhang2019lookahead} is a particularly simple moving-reference update, originally proposed for optimization stability. SFA uses a fixed reference instead. \citet{abbes2026replay} evaluate replay-free Reptile-style updates in multilingual continual pretraining, while other retention and fast-slow approaches appear in Appendix~\ref{app:lookahead-context}. We contribute a controlled examination of replay-free interpolation for \emph{supervised} domain adaptation and adapter task streams, emphasizing comparison to rehearsal and adaptive fusion rather than claiming priority over Lookahead-style continual learning.

\paragraph{Model merging.}

There exist many techniques and applications for merging multiple models to create a single model with improved generalization on a given set of tasks. Model souping \citep{wortsman2022model} involves averaging the parameters of existing models to create a new model. Task Arithmetic \citep{ilharco2023editing} involves finding and adding task vectors to create a multi-task model. WiSE-FT~\citep{wortsman2022robust} merges the weights of a pretrained and a fine-tuned model.

TIES resolves sign conflicts when combining task updates \citep{yadav2023tiesmerging}, while continual weight interpolation and MagMax apply merging to sequential learning \citep{kozal2024continuallearningweightinterpolation,marczak2024magmaxleveragingmodelmerging}. Branch-and-Merge repeatedly combines parallel adaptation branches \citep{alexandrov2024bam}; SFA modifies a single training trajectory. More recently, Recurrent-KIF performs iterative fusion using dynamically updated parameter-importance estimates and replay \citep{feng2025recurrent}. AIMMerging also combines repeated fusion with replay, but adapts fusion timing and weights using learning and forgetting signals extracted from the training trajectory \citep{feng2025aim}. We compare SFA with AIMMerging under a shared adapter protocol. BECAME combines gradient projection with adaptive merging \citep{li2025became}, and SLAO uses orthogonal initialization and asymmetric LoRA merging \citep{qiao2026merge}. Further related work, including storage-budgeted adapter consolidation, appears in Appendix~\ref{app:recent}.

\section{Fixed-Reference In-Training Averaging}\label{averaging:continual}

We investigate two simple forms of in-training interpolation: Lookahead, which uses an evolving reference, and SFA, which we introduce as a fixed-reference alternative. During fine-tuning on a new task, SFA periodically averages the working parameters with a checkpoint obtained from previous tasks and continues optimizing from the averaged model, while keeping the reference fixed throughout the task.

More precisely, let $\theta_0$ denote the parameters of
the model at the beginning of the current task, and let
$a$ denote the reference checkpoint used for averaging.
Let $\theta_{t+1}^{*}$ be the parameters obtained after
taking a gradient step on the new task at time
$0 \leq t \leq T-1$, using the current parameters $\theta_t$.

At intervals of $pT$ iterations, as well as at the end
of fine-tuning, we interpolate the updated parameters
with the fixed reference $a$, using a weight
$0 \leq \beta \leq 1$:
\begin{equation}
\theta_{t+1}=\beta a+(1-\beta)\theta_{t+1}^{*}.
\label{eq:sfa}
\end{equation}
At other iterations, $\theta_{t+1}=\theta_{t+1}^{*}$.
The reference $a$ remains fixed throughout the current
task. The parameter $p$ specifies the interval between
merges as a fraction of task length: smaller values
produce more frequent averaging. Since $pT$ need not
be an integer, Algorithm~\ref{alg:sfa} rounds it to the
nearest integer, with a minimum interval of one step,
and applies the final merge exactly once.

\begin{algorithm}[t]
\caption{SFA for one task}
\label{alg:sfa}
\begin{algorithmic}[1]
\REQUIRE Task-start checkpoint $\theta_0$,
reference checkpoint $a$, optimizer $\mathcal{O}$,
interval fraction $0<p\leq1$, anchor weight
$0\leq\beta\leq1$, positive integer updates $T$
\STATE $\theta\gets\theta_0$
\STATE $k\gets\max(1,\lfloor pT+0.5\rfloor)$
\FOR{$t=1,\ldots,T$}
    \STATE $\theta\gets\mathcal{O}(\theta;\text{current-task batch})$
    \IF{$t\bmod k=0$ \textbf{or} $t=T$}
        \STATE $\theta\gets\beta a+(1-\beta)\theta$
    \ENDIF
\ENDFOR
\STATE \textbf{return} $\theta$
\end{algorithmic}
\end{algorithm}

By default, SFA refreshes the reference to the final model
after each task, so that $a=\theta_0$ at the beginning of
the next task. Alternatively, $a$ can remain fixed across
subsequent tasks to prioritize a particular earlier solution,
while the working model continues learning from the most
recent checkpoint (Appendix~\ref{app:reference-choice}).

The averaging interval and weight jointly determine the
influence of the reference. Our experiments examine both
a fixed weight per merge and weights adjusted for the
number of merges, so that more frequent averaging does
not automatically impose a stronger cumulative correction
toward the reference (Appendix~\ref{app:merge-schedule}).

\paragraph{Connection to Lookahead.}
SFA and Lookahead use the same parameter interpolation
operation but differ in how the reference evolves.
At a merge step, both interpolate the updated working
parameters with a reference, as in
Eq.~\ref{eq:sfa}. SFA keeps $a$ fixed throughout
the current task, whereas Lookahead updates its slow
reference to the interpolated parameters after each
synchronization. Both continue training from the
interpolated model. Thus, the distinction is not
the averaging operation itself, but whether its
reference remains fixed or evolves during training.

\paragraph{Intuition for averaging.}

We connect the fixed-reference update to penalty-based methods that constrain drift from an earlier solution to offer intuition for why merging is so effective. 

SFA has a related interpretation: writing $a=\theta_0$
for the reference and $z=\theta_{t+1}^{*}$ for the parameters
immediately before averaging, each merge with $0\leq\beta<1$
is the unique solution of a quadratic correction toward the
reference:

\begin{equation}
\begin{split}
(1-\beta)z+\beta a
=\operatorname*{arg\,min}_{x}\bigl\{
&\tfrac12\|x-z\|_2^2\\
&+\tfrac{\beta}{2(1-\beta)}\|x-a\|_2^2
\bigr\}.
\end{split}
\end{equation}

Each merge therefore contracts the current parameter displacement from the reference by a factor of $1-\beta$, while allowing learning to continue between corrections. Unlike Lookahead, which updates its reference after each synchronization, SFA repeatedly contracts toward the same checkpoint throughout the task. This provides a direct mechanism for limiting drift from a designated earlier solution. The derivation and regularization experiments appear in Appendix~\ref{sec:continual_merging}.

\paragraph{Implementation and cost.}

SFA retains one reference copy and adds no forward or backward passes. Unlike replay-based approaches, it neither stores past training examples nor computes their gradients; it also avoids the monitoring and additional replay updates used by AIMMerging. Under matched exposure to the current task, this reduces training work, as measured in Section~\ref{sec:adapterresults}. For $P$ trainable parameters and $K$ merges, SFA stores $P$ extra values and performs $O(KP)$ work, as does Lookahead. Their data movement differs: SFA interpolates directly into the working weights, leaving its reference unchanged, whereas Lookahead updates its slow weights and copies the result into the working weights. In our implementations, Lookahead's additional copy explains the measured merge-time difference (Appendix~\ref{app:merge-implementation}). SFA also requires no posterior estimation or curvature-weighted fusion, unlike variational approaches such as CoVON \citep{paul2026covon}.

\section{Experimental Setup}\label{sec:setup}

We investigate forgetting in two complementary language settings. The first examines a substantial domain shift between two successive fine-tuning tasks. The second follows a model through a sequence of text-classification tasks, testing whether SFA can preserve several learned tasks as new ones arrive. We also evaluate longer, 20-task image-classification sequences on Food-101 and CIFAR-100 (Appendix~\ref{image_tasks_SFA}).

\paragraph{Language-domain adaptation.}
To examine forgetting across distinct domains, we fine-tune Llama~2-7B and Qwen3-1.7B \citep{qwen32025} successively on Math and Law, with additional Pythia-2.8B and Qwen2.5-1.5B comparisons in Appendix~\ref{app:earlier-experiments}. Models first learn from MetaMathQA \citep{yu2023metamath}, before adapting to CaseHOLD, Terms of Service, and Overruling \citep{zheng2021does,Lippi_2019,tos_binary}. We evaluate mathematical retention on GSM8K \citep{cobbe2021training} and new-domain learning through accuracy on the three Law tasks. This allows us to examine the balance between acquiring domain-specific knowledge and preserving performance on a substantially different task. Training and evaluation details appear in Appendices~\ref{app:controlled-setup} and~\ref{app:experimental-details}.

\paragraph{Sequential text-classification tasks.}
Beyond a transition between two domains, we investigate whether interpolation can preserve knowledge across a sequence of tasks: entity-type classification on DBpedia, sentiment classification on Amazon reviews, question-topic classification on Yahoo Answers, and news-topic classification on AG News \citep{zhang2015character}. We evaluate all tasks seen so far after each stage.

We study this stream using full-parameter Qwen3-1.7B at a common total token budget (Appendix~\ref{app:fullparam-stream}). To extend the comparison to larger models, we use LoRA \citep{hu2022lora} on Qwen3-1.7B, 8B, 14B, and 32B. Updating a single adapter throughout the stream reduces fine-tuning cost while allowing continued adaptation. These adapter comparisons hold current-task exposure fixed, accounting separately for replay and monitoring (Appendix~\ref{app:controlled-setup}).

\paragraph{Comparison methods.}
Our primary comparison asks whether \emph{SFA and Lookahead}, as simple replay-free interpolation approaches, remain competitive with rehearsal and AIMMerging under a shared protocol. Lookahead also serves as a moving-reference control for analyzing SFA's fixed-reference choice. Endpoint interpolation and Task Arithmetic test whether updating the training trajectory matters, while EMA distinguishes training-time from evaluation-only averaging. Sequential fine-tuning provides an essential baseline, and Appendix~\ref{sec:continual_merging} compares with EWC.

\section{Results}\label{sec:results}

\subsection{Mitigating forgetting across language domains}\label{sec:newdomain}

By restricting ourselves to pairs of dissimilar tasks, we can clearly quantify the trade off between learning the second task and forgetting the first one by visualizing the results on a plane that measures the accuracy of the first task on the y-axis and the accuracy of the second task on the x-axis.

We begin with Qwen3-1.7B adapting from Math to Law. Within each seed, methods share the Math-trained checkpoint and Law example order. We measure retention on GSM8K and adaptation through mean accuracy across the three Law tasks. The Llama~2 experiments extend the comparison to another model family, using CaseHOLD for Law.

\begin{figure}[t]
\centering
\includegraphics[width=\columnwidth]{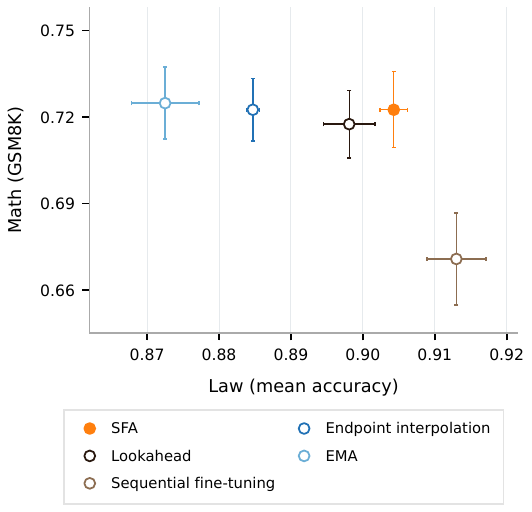}
\caption{Qwen3-1.7B Math$\to$Law: replay-free comparisons over three seeds (17, 42, 123). Both axes show mean $\pm$ sample SD. All underlying Law-training runs use learning rate $10^{-5}$. SFA uses the development-selected $\beta_{\mathrm{end}}=.5$. Appendix~\ref{app:m2l-protocol} gives the configurations; Appendix~\ref{app:m2l-tradeoffs} reports the SFA strength sweep. Table~\ref{tab:newm2l} additionally includes seeds 7 and 2026 and uses development-selected sequential fine-tuning at $5\cdot10^{-6}$.}
\label{fig:newm2l}
\end{figure}

\begin{figure*}[t]
\centering
\includegraphics[width=\textwidth]{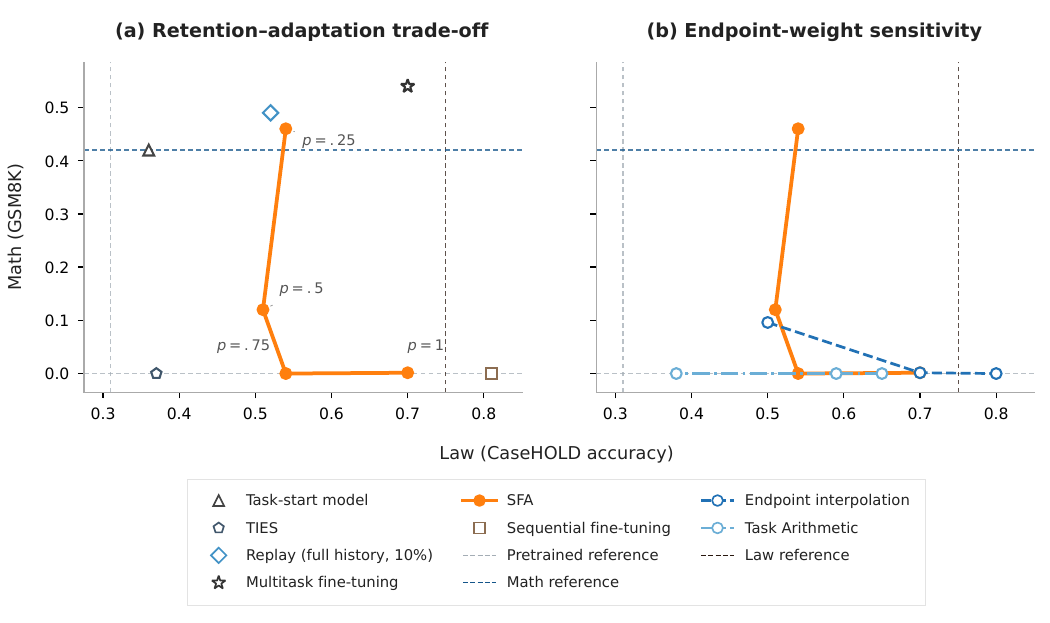}
\caption{Llama~2-7B Math$\to$Law experiments. \textbf{(a)} Intermediate SFA merges recover Math accuracy lost during sequential fine-tuning. \textbf{(b)} The SFA interval sweep is compared with endpoint-interpolation and Task Arithmetic weight sweeps, showing favorable intermediate-merge operating points beyond the equal-weight endpoint. Panel (b) sweeps weights of 0.25, 0.50, and 0.75 on the Law-adapted checkpoint for both endpoint-interpolation and Task Arithmetic. Replay samples from full history.}
\label{fig:llama7b_mathlaw}
\label{fig:llama7b_mathlawtaskarithsfasweep}
\end{figure*}

\paragraph{Comparison with sequential fine-tuning.}
We first examine whether incorporating earlier model weights improves upon ordinary sequential fine-tuning. On Qwen3-1.7B, SFA achieves slightly higher mean Math and Law accuracies across five seeds, even when compared with a lower-learning-rate baseline selected to preserve Math while maintaining Law performance (Table~\ref{tab:newm2l}). The contrast is more pronounced on Llama~2-7B: sequential fine-tuning loses nearly all Math performance as it learns Law, whereas SFA recovers much of this earlier knowledge while still improving Law accuracy over the Math-trained starting model (Figure~\ref{fig:llama7b_mathlaw}(a)). Across both model families, incorporating earlier checkpoints therefore offers a favorable balance between retaining past knowledge and learning the new domain.

\paragraph{Comparison with other averaging methods.}
We next examine whether this balance benefits from allowing the merged model to continue learning. We compare SFA with endpoint interpolation, EMA, and the Lookahead averaging control introduced in Section~\ref{sec:setup}. On Qwen3, SFA and endpoint interpolation at $\beta=.5$ retain essentially the same mean Math accuracy across three matched runs, but SFA achieves $1.95\pm0.24$ percentage points higher Law accuracy (Figure~\ref{fig:newm2l}). EMA retains slightly more Math but achieves lower Law accuracy, suggesting that maintaining averaged weights for evaluation does not necessarily provide the same benefit as continuing to train from them. We observe a consistent benefit of intermediate merging on Llama~2, where the SFA sweep contains favorable retention--adaptation trade-offs beyond those reached by the displayed endpoint-interpolation and Task Arithmetic weight sweeps (Figure~\ref{fig:llama7b_mathlaw}(b)).

Lookahead offers a second form of replay-free in-training averaging and likewise achieves a competitive retention--adaptation balance on Qwen3 (Table~\ref{tab:newm2l}). The fixed- and moving-reference methods produce nearby operating points. Appendix~\ref{app:m2l-protocol} reports the selected settings, paired differences, and independent replication seeds.

\paragraph{Controlling the retention-adaptation balance.}
SFA also allows greater emphasis on retention without returning to past examples. Increasing $\beta_{\mathrm{end}}$ from .25 to .75 raises Qwen3 Math accuracy from 69.93\% to 73.04\%, while Law accuracy decreases from 91.22\% to 88.05\%. At $.90$, Math finishes only $0.40\pm0.04$ points below its starting accuracy, while Law improves by $45.82\pm1.40$ points. Thus, users can retain nearly all starting Math accuracy while still learning the new domain. Appendix~\ref{app:m2l-tradeoffs} reports the full sweep (Table~\ref{tab:m2l-strength}).

\subsection{Replay-free averaging and rehearsal}\label{sec:fullparam-stream}

We next follow Qwen3-1.7B through DBpedia $\to$ Amazon $\to$ Yahoo Answers $\to$ AG News using full-parameter fine-tuning. We compare SFA with replay and their combination, SFA-R. SFA-R interleaves past-task examples with current-task data while periodically merging the working model with the fixed task-start checkpoint, combining rehearsal with SFA's averaging updates. Replay and SFA-R use the same 500-example reservoir and schedules, allocating approximately 10\% of input tokens to past examples after the first task within a common total token budget.

Table~\ref{tab:fullparam-replay-cost} shows that SFA approaches replay's accuracy with lower measured runtime. Adding replay to SFA (SFA-R) yields the highest mean accuracy, improving on replay by $1.0\pm0.4$ percentage points at similar measured cost (130.7 versus 133.0\,s). Thus, simple interpolation can serve either as an alternative to rehearsal or as a component alongside it; these results do not imply that rehearsal is never beneficial. Appendix~\ref{app:fullparam-stream} gives the timing protocol and per-task accuracies (Table~\ref{tab:fullparam-replay-tasks}).

\begin{table}[t]
\centering\small
\caption{Qwen3-1.7B full-parameter four-task learning at a common total token budget. Final average test accuracy is mean $\pm$ sample SD over three seeds; runtime averages two complete-stream training-loop measurements. Replay and SFA-R use approximately 10\% replay after the first task. Bold marks the highest accuracy and lowest measured runtime.}
\label{tab:fullparam-replay-cost}
\begin{tabular}{@{}lrr@{}}
\toprule
Method & Final average (\%) $\uparrow$ & Runtime (s) $\downarrow$\\
\midrule
SFA & $78.4\pm0.8$ & $\mathbf{117.5}$\\
Replay (10\%) & $78.8\pm0.3$ & $133.0$\\
SFA-R & $\mathbf{79.8}\pm0.2$ & $130.7$\\
\bottomrule
\end{tabular}
\end{table}

\subsection{Learning a sequence of tasks with adapters}\label{sec:adapterresults}

\begin{figure*}[t]
\centering
\includegraphics[width=\textwidth]{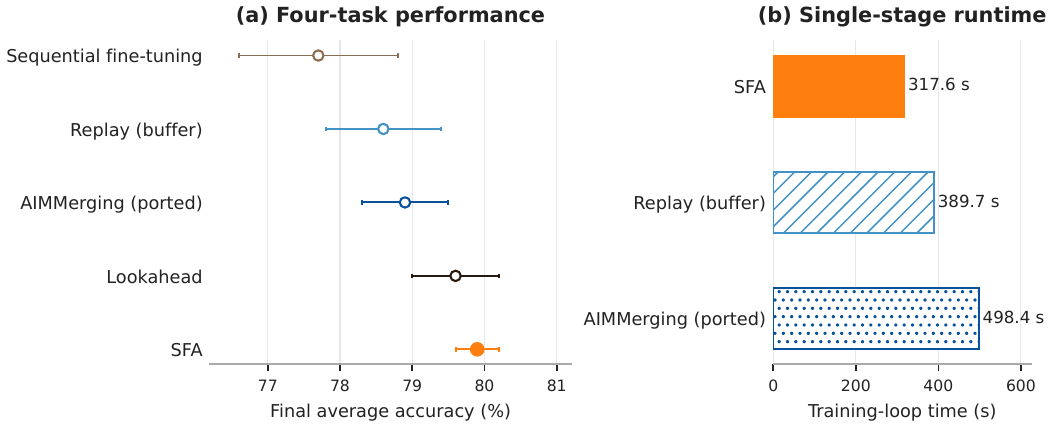}
\caption{Qwen3-14B adapter results under matched current-task exposure. \textbf{(a)}: final four-task accuracy for fixed configurations (three-seed mean and sample SD). Both SFA and Lookahead exceed the tested replay and AIMMerging means. \textbf{(b)}: controlled Amazon-stage training-loop time on one H200 (two repetitions); the measured SFA savings are specific to this stage. Peak GPU memory is comparable.}
\label{fig:cost}
\end{figure*}

We now test whether both simple interpolation approaches compete with methods that store past examples or adaptively monitor learning, using fixed configurations transferred across adapter model sizes. To match synchronization frequency, SFA and Lookahead each merge four times per task. Lookahead uses $\beta=.5$ per synchronization, an effective setting in \citet{zhang2019lookahead}; SFA uses cumulative $\beta_{\mathrm{end}}=.5$, motivated by earlier domain-adaptation results. Figure~\ref{fig:cost} shows the 14B comparison; Appendices~\ref{app:model-size} and~\ref{app:coefficient-frequency} report the full size sweep and tuning sensitivity.

On Qwen3-14B, the two interpolation methods obtain 79.9\% (SFA) and 79.6\% (Lookahead) mean final accuracy, compared with 78.6\% for replay and 78.9\% for the AIMMerging port (Figure~\ref{fig:cost}). SFA exceeds both replay-based configurations on all three seeds; its paired mean gains are $1.3\pm0.8$ and $1.0\pm0.7$ points, respectively. Replay adds past examples to each batch, while AIMMerging uses additional replay, monitoring, and an adaptive controller. Neither interpolation method requires these operations. In an alternative task order, SFA again has the highest fixed-configuration mean (Appendix~\ref{app:task-order}); at smaller scales, the replay-based methods can do better (Appendix~\ref{app:model-size}).

The additional mechanisms also matter for cost. In the controlled Amazon-stage comparison, SFA uses 18.5\% less training-loop time than replay and 36.3\% less than AIMMerging (Figure~\ref{fig:cost}).  Paired with the accuracy results, this motivates evaluating inexpensive interpolation before assuming that more elaborate continual-learning machinery is required (Appendix~\ref{app:timing-protocol}).

This result is conditional on model size. Relative to AIMMerging, SFA moves from $-2.7$ points at 1.7B to $-0.4$ at 8B and $+1.0$ at 14B; the 32B comparison narrows again (Appendix~\ref{app:model-size}). Thus, the main conclusion is not that interpolation always wins, but that the incremental value of replay and adaptive fusion must be demonstrated against these low-overhead baselines.

\subsection{Extending to longer image-classification sequences}
We also examine whether the benefits of intermediate averaging extend beyond language tasks. Across 20-task sequences from Food-101 and CIFAR-100, selected SFA configurations improve upon sequential fine-tuning, endpoint interpolation, and Task Arithmetic. In these settings, averaging near the end of each task allows the model to first acquire the new task before recovering earlier performance. The Food-101 learning curves suggest why: current-task accuracy improves over a substantial portion of training, so earlier corrections toward the prior checkpoint may hinder learning before the new task has been sufficiently acquired. Appendix~\ref{image_tasks_SFA} presents the results and comparisons with replay and CLS-ER.

\section{Understanding the effect of averaging}\label{sec:analysis}

We consistently find that \emph{simple interpolation during training is a substantive forgetting-mitigation baseline}. Lookahead was designed for optimization stability, yet it is competitive in our supervised task streams; SFA provides a fixed-reference alternative with similar simplicity. Both avoid replay gradients and auxiliary objectives. Their favorable results challenge the assumption that stronger retention necessarily requires an additional replay or monitoring mechanism.

The intervention remains sensitive to \emph{when} and \emph{toward what} the model is averaged. More frequent merges can favor Math retention in domain adaptation, whereas longer vision streams favor allowing substantial current-task learning before merging (Figure~\ref{fig:llama7b_mathlaw}; Appendix~\ref{image_tasks_SFA}). At matched cumulative strength $\beta_{\mathrm{end}}=.5$, two and four merges reach the same 79.9\% mean final accuracy at 14B (Appendix~\ref{app:coefficient-frequency}). The lower-learning-rate sequential controls also narrow SFA's gains (Appendix~\ref{app:othertransitions}). These findings argue for comparing against well-controlled simple alternatives, rather than treating the complexity of a method as evidence of its necessity.

Furthermore, the two reference policies serve different purposes. Lookahead updates its slow reference throughout training, making it a natural choice when no particular earlier checkpoint is prioritized; development-selected Lookahead also achieves higher final average accuracy in our adapter comparisons (Appendix~\ref{app:coefficient-frequency}). SFA instead permits a chosen checkpoint to remain the reference across later tasks. For example, retaining the post-Amazon checkpoint improves Amazon accuracy at the expense of Yahoo accuracy relative to refreshing the reference (Appendix~\ref{app:reference-choice}). Thus, the choice depends on whether the aim is to maintain an evolving reference or to protect a designated earlier solution, rather than on one rule uniformly outperforming the other.

\section{Conclusion}

Before adding machinery to mitigate forgetting, we ask how far simple in-training averaging can go. In supervised language adaptation and adapter task streams, both SFA and Lookahead can compete with more elaborate tested approaches; the SFA experiments also extend to longer vision streams. SFA contributes an explicit fixed-checkpoint reference; Lookahead illustrates that an existing optimization update can also mitigate forgetting. Our results motivate treating inexpensive interpolation as a serious baseline: additional complexity should justify itself through measurable gains in retention, adaptation, or cost.

\FloatBarrier\label{maintextend}

\section*{Limitations}

We do not suggest that in-training interpolation
universally replaces rehearsal or other forgetting-mitigation
methods. Its effectiveness depends on the learning setting,
model size, and desired balance between retention and
adaptation. Nevertheless, the fact that simple replay-free
interpolation outperforms the tested rehearsal and adaptive
merging configurations in our experiments
demonstrates that additional mechanisms are not always
necessary for competitive performance. These findings
motivate evaluating lightweight interpolation before
introducing more elaborate approaches, while broader
comparisons across tasks and training configurations
would further clarify when each approach is preferable.

Our computational results likewise illustrate the potential
benefits of simplicity. In a controlled adapter
comparison, SFA reduces training-loop time relative to
the tested replay-based methods, while its simpler parameter
update incurs less merge-operation overhead than Lookahead
in our implementation. The 36.3\% training-loop reduction
relative to AIMMerging was measured on one Amazon training
stage, rather than across the complete task sequence, with
two controlled repetitions on the evaluated hardware.
Moreover, merge-operation savings should not be interpreted
as equivalent end-to-end speedups. Both SFA and Lookahead
retain an additional parameter reference, so the simpler
update does not imply a memory advantage.

%% file: appendix.tex
\section{Controlled Experiment Protocols}
\label{app:controlled-setup}
{
\paragraph{Method names.}
We use ``Sequential fine-tuning'' for unregularized sequential training; the experiment specifies whether all parameters or only LoRA adapters are trained. ``Replay (full history)'' samples from retained past training sets, whereas ``Replay (buffer)'' uses a bounded memory. ``Endpoint interpolation'' averages the task-start and final checkpoints (SFA with $p=1$). ``SFA-R'' denotes SFA augmented with replay; buffer capacities or sampling fractions identify its variants. ``AIMMerging (ported)'' refers to the controller port described below.

\subsection{Merge schedules and parameterization}
\label{app:merge-schedule}
Writing $a=\theta_0$ for the task-start reference, all SFA experiments use Eq.~\ref{eq:sfa}. The Llama~2, Pythia, Qwen2.5, and vision interval sweeps use per-merge $\beta=.5$ unless a coefficient sweep is stated. In the Qwen3 controlled comparisons, we parameterize the weight by a cumulative contraction $\beta_{\mathrm{end}}$: for $K\geq1$ scheduled merges and $0\leq\beta_{\mathrm{end}}\leq1$,
\begin{equation}
\beta=1-(1-\beta_{\mathrm{end}})^{1/K}.\label{eq:contraction}
\end{equation}
If no training updates occurred between merges, their combined effect would multiply the parameter difference from the reference by $1-\beta_{\mathrm{end}}$. During training, this parameterization summarizes the strength of the averaging operations, rather than determining the model's final distance from the reference. It does not make the training trajectories equivalent. The fixed configuration uses $p=.25$ and $\beta_{\mathrm{end}}=.5$; when the rounded schedule gives $K=4$, each merge uses $\beta\approx.1591$. This is a hyperparameter setting of SFA, not a different update rule.

In order to specify the update schedule unambiguously, we convert the interval fraction $p$ into an integer $k=\max(1,\lfloor pT+0.5\rfloor)$ and merge at its multiples, with a final merge at $T$ applied exactly once. The resulting count is $K=\lceil T/k\rceil$, which determines the weight in Eq.~\ref{eq:contraction}. Rounding can give more than four merges at $p=.25$: for example, $T=45$ gives $k=11$ and merges at 11, 22, 33, 44, and 45. We retain Adam moments across within-task merges and reset the optimizer at task boundaries. Lookahead updates $s\gets(1-\alpha)s+\alpha z$ and initializes the working weights from $s$, with $\alpha=1-\beta_{\mathrm{LA}}$. SFA instead retains the same reference $a$ throughout a task. Equal numerical coefficients consequently do not imply equal regularization. In the LoRA experiments, both rules average the trainable $A$ and $B$ factors separately; this is not equivalent to averaging the dense product $BA$.

\subsection{Full-parameter domain adaptation}
We begin with Qwen3-1.7B (revision \texttt{70d244cc}) and fine-tune it successively on Math and Law. Throughout these experiments, we disable thinking in the chat template. We use response-only token-mean loss, AdamW ($\beta_1=.9$, $\beta_2=.999$, no weight decay), gradient clipping at 1.0, a batch of 32 sequences, 3\% warmup followed by cosine decay, fp32 master weights, and bf16 autocast. Each task receives $10^7$ input tokens. The optimizer and schedule restart at task boundaries, while optimizer moments persist across within-task merges.

The Math development set consists of 500 held-out GSM8K training questions. MetaMathQA examples derived from these questions are excluded from training, as are rows with 13-gram overlap with GSM8K test. The test set contains 1,319 questions and uses greedy generation and numeric answer extraction. The overlap screen removed 103 MetaMathQA rows; it is not a semantic contamination audit. CaseHOLD validation and label-stratified 5\% portions of the ToS and Overruling training sets supply Law development data. Law evaluation uses the fixed test files shared by all methods. Candidate answers are scored by mean token log-probability, and Law accuracy is the unweighted mean of the three component accuracies.

To isolate the effect of the update rule, all methods within a seed begin from the same Math-FT checkpoint and encounter the Law examples in the same order. The Law stage has 738--740 updates. At the four-merge setting, SFA and frequency-matched Lookahead act at updates 185, 370, 555, and the final update. Endpoint interpolation uses the task-start model and the final sequentially fine-tuned model. EMA produces an evaluation model without replacing the working parameters. Retention in this transition refers to preserving the Math-FT checkpoint: the pretrained model itself scores 79.9\% on GSM8K, above the Math-FT starts for seeds 17, 42, and 123 (74.1--75.7\%). We do not interpret the first SFT stage as a demonstrated improvement in mathematical capability.

\subsection{Shared-adapter continual learning}
For the adapter experiments, we continue training one rank-8 LoRA adapter as each new task is introduced. The adapter uses scaling $\alpha=32$, dropout .05, and query/value projection targets. The base remains frozen in fp32; adapter master weights are fp32 with bf16 autocast. We use Qwen3 revisions \texttt{70d244cc} (1.7B), \texttt{b968826d} (8B), \texttt{40c06982} (14B), and \texttt{9216db57} (32B). The four datasets are DBpedia, Amazon reviews, Yahoo Answers, and AG News \citep{zhang2015character}. Order 1 follows this sequence, whereas order 4 uses
DBpedia $\rightarrow$ Yahoo Answers $\rightarrow$
Amazon reviews $\rightarrow$ AG News. Each stage has 1,000 training examples, ten epochs, effective batch size 32, and 310 updates. Following the released gradient-accumulation schedule, ten per-epoch permutations are concatenated and the first 310 batches of 32 are used: 9,920 current-task examples per stage. Updates may cross epoch boundaries; the last 80 examples of the tenth permutation are unused. AdamW uses learning rate $3\cdot10^{-4}$, 50 warmup updates, linear decay, clipping at 1.0, and a fresh optimizer at each stage. Final weights initialize the following stage and are used for all evaluations.

Both methods are applied from the first stage, whose anchor/slow state is the initialized adapter ($B=0$). SFA and Lookahead merge the trainable $A$ and $B$ tensors at updates 78, 156, 234, and 310. Fixed-configuration SFA uses $\beta_{\mathrm{end}}=.5$ ($\beta\approx.1591$ per merge); fixed-configuration frequency-matched Lookahead uses $\beta_{\mathrm{LA}}=.5$. At every task boundary, the final Lookahead synchronization has already made the slow and working weights identical. Initializing the next stage's slow copy from those working weights therefore preserves its value. SFA instead refreshes its reference to its own preceding final adapter and holds that reference fixed during the next task. Neither method restores the model to the beginning of the complete stream; optimizer moments are reset for all methods at each task boundary. This synchronization frequency matches SFA, rather than the much shorter $k=5$ or $k=10$ intervals often used for Lookahead optimization. The adapter comparisons do not evaluate those frequent-sync configurations; Appendix~\ref{app:highfrequency} examines intervals of five and one updates in full-parameter domain adaptation.

Replay stores 20 examples per previous task: 20, 40, and 60 during the three later stages. Eight replay examples are added to each current-task batch. The loss is a token mean over the combined batch, so the replay contribution follows its share of supervised tokens. AIMMerging uses the same memory examples but its controller determines additional replay and monitoring operations. Thus current-task exposure is matched, while total tokens and optimizer steps are not. In the 1.7B stream, AIMMerging averages 629 additional replay steps and 36.7 merges, with 24.7 adaptive interval adjustments. It adaptively adjusted its merging intervals during our experiments, rather than using a fixed merge schedule. Our port uses final checkpoints for all methods instead of AIMMerging's released best-development-loss selection. All arms also use the shared supervised-token-mean loss rather than the released microbatch-mean reduction. AIMMerging performs no replay, monitoring, or merges in the first stage, where its memory is empty, as in the released controller. Controller transcription checks and agreement between adapter and exported dense-model predictions validate the implementation, not equivalence to the published experiments.

The released development files overlap the released test files. We remove those overlaps from development, keep the canonical test files, and do not select training checkpoints by development loss. Clean development sizes are 33 (DBpedia), 339 (Amazon), 182 (Yahoo), and 370 (AG News); test sizes are 7,000, 2,500, 5,000, and 2,000, respectively. Each training set has 1,000 examples and no recorded train--test overlap. The small DBpedia development set limits selection precision. Prompts and label verbalizers come from the released classification files, wrapped in the Qwen3 chat template with thinking disabled. Evaluation uses greedy generation, a 32-token limit, and exact-match classification accuracy.

The controller port is based on AIMMerging revision \texttt{90733c38e571}. It initializes the interval and outer replay count to eight, bounds the interval to $[2,128]$, and retains its adaptive fusion and monitoring rules. Software versions are PyTorch 2.13.0+cu130, Transformers 5.18.0, vLLM 0.30.0, and PEFT 0.21.2. The implementation logs merge events, controller changes, token counts, and optimizer steps; immutable split manifests and checkpoint hashes record provenance.

Memory-driven microbatch token caps are 16,384, 8,192, 4,096, and 2,048 at 1.7B, 8B, 14B, and 32B, respectively. The 32B runs additionally use activation checkpointing; no quantized training is used. A seed-17 five-arm 8B checkpointing control changes final accuracy by $-1.4$ to $+1.1$ points across methods. Focused checks verify configuration agreement and dropout-mask replay, but one control seed cannot estimate a general run-to-run noise floor. We make within-size comparisons and do not attribute differences between sizes solely to parameter count.

\subsection{Metrics and timing}\label{app:timing-protocol}
Let $a_{t,j}$ be accuracy on task $j$ after stage $t$, with $N=4$. Final average accuracy is $N^{-1}\sum_{j=1}^N a_{N,j}$, acquisition is $N^{-1}\sum_{j=1}^N a_{j,j}$, and final earlier-task accuracy is $(N-1)^{-1}\sum_{j=1}^{N-1}a_{N,j}$. Signed forgetting is
\begin{equation}
F=\frac{1}{N-1}\sum_{j=1}^{N-1}\left(\max_{j\leq t<N}a_{t,j}-a_{N,j}\right).
\end{equation}
It can be negative when subsequent training improves a task. Reporting acquisition prevents low initial learning from being mistaken for successful retention.

Controlled runtime measurements use one Amazon stage on one H200, with two repetitions in reversed method order, identical starting adapters, precision and microbatch policies, and matched instrumentation. Development-loss logging is disabled. Training-loop time includes optimization, merges, replay, and controller monitoring, but excludes loading, saving/export, and evaluation. Repetition spread is at most 2.7\% at 8B and 1.4\% at 14B. Table~\ref{tab:timingdetail} reports the measured means. These are not full-stream or end-to-end speedups.

\begin{table}[t]
\centering\small
\caption{Controlled single-stage training-loop time in seconds. Each entry averages two runs on one H200.}
\label{tab:timingdetail}
\begin{tabular}{@{}lrr@{}}\toprule
Method & 8B & 14B\\\midrule
SFA & 243.0 & 317.6\\
Replay (buffer) & 277.7 & 389.7\\
AIMMerging (ported) & 350.8 & 498.4\\\bottomrule
\end{tabular}
\end{table}

\begin{table}[t]
\centering\small
\caption{Additional training work in the 14B four-task adapter stream. Every method processes 6.01M current-task tokens. Replay and monitoring tokens are distinct from the single-stage timing in Figure~\ref{fig:cost}.}
\label{tab:tokens}
\begin{tabular}{@{}lrr@{}}
\toprule
Method & Replay-gradient & Monitoring\\
\midrule
SFA / Lookahead & 0 & 0\\
Replay (buffer) & 1.20M & 0\\
AIMMerging (ported) & 2.51M & 4.20M\\
\bottomrule
\end{tabular}
\end{table}

\subsection{Storage and compute accounting}\label{app:storage-cost}
SFA and Lookahead each retain one auxiliary trainable-parameter copy. In the full-parameter 1.7B classification stream this is 6.88\,GB in fp32, and both peak at about 89.7\,GiB versus 83.3\,GiB for sequential fine-tuning. A 500-example replay buffer occupies approximately 0.3\,MB of text in that experiment. In the adapter comparison, SFA and Lookahead instead retain only adapter copies: 6.1\,MiB at 1.7B and 20\,MiB at 14B. The measured 14B peak is essentially identical across all five arms (118.3\,GiB).

\paragraph{Why SFA merges cost less.}\label{app:merge-implementation}
Writing $w$ for working weights, $a$ for SFA's reference, and $s$ for Lookahead's slow weights, the measured implementations apply
\begin{align*}
\text{SFA:}\quad &w\gets(1-\beta)w+\beta a,\\
\text{Lookahead:}\quad &s\gets\beta_{\rm LA}s+(1-\beta_{\rm LA})w,\\
&w\gets s.
\end{align*}
The interpolation arithmetic is similar, but SFA writes its result directly into the working weights. Lookahead must leave both working and slow weights equal at synchronization: its working weights then continue optimizing while the slow weights remain available for the next interpolation. The tested implementation achieves this with an additional device-to-device copy over all trainable parameters. In the per-tensor implementation, SFA uses an in-place multiply and scaled add; Lookahead uses the same operations on its slow tensor followed by a copy into its working tensor. The batched implementation likewise uses one interpolation operation for SFA, and interpolation followed by copying for Lookahead. Neither method copies data to the CPU or allocates a third parameter bank during these merges. The extra read/write traffic and launch work explain the measured constant-factor cost difference, while both retain $O(P)$ storage and work per event. The percentage saving is specific to the tested implementations, precision, and hardware; it is not a lower bound on every possible Lookahead implementation.

We isolate merge cost using synthetic tensors with the exact full-parameter shapes of Qwen3-1.7B, 8B, and 14B, and the rank-8 query/value adapter shapes used with the 14B model. The full-parameter sets contain 1.72B, 8.19B, and 14.77B unique values; the adapter contains 5,242,880 values. Tied parameters are counted once. All sizes are measured in the same session on one H200 with fp32 values and $\beta=.5$, using two parameter banks without gradients, optimizer states, or model forward passes. Each of two blocks uses ten warm-ups and 50 synchronized CUDA-event measurements per operation; method order is reversed between blocks. Values are reset outside the timed region. We compare both the per-tensor operations used in training and batched tensor operations. These are timing repetitions, not training seeds.

\begin{table}[t]
\centering\small
\caption{Five-seed Qwen3-1.7B Math$\to$Law test accuracy (\%, mean $\pm$ sample SD; seeds 17, 42, 123, 7, 2026). The starting model is evaluated before Law fine-tuning; bold marks the highest adapted mean. Configurations maximize Math development accuracy while allowing at most a one-point Law reduction relative to reference sequential fine-tuning (Appendix~\ref{app:m2l-protocol}). Sequential fine-tuning uses learning rate $5\cdot10^{-6}$.}
\label{tab:newm2l}
\begin{tabular}{@{}lrr@{}}\toprule
Method & GSM8K & Law mean\\\midrule
Math-trained start & $74.80\pm0.65$ & $35.63\pm1.87$\\
\midrule
Sequential fine-tuning & $71.18\pm0.61$ & $90.07\pm0.24$\\
Lookahead, $\beta=.5$ & $71.45\pm0.95$ & $89.74\pm0.32$\\
SFA, $\beta_{\rm end}=.5$ & $\mathbf{71.83}\pm1.14$ & $\mathbf{90.18}\pm0.43$\\
\bottomrule\end{tabular}
\end{table}
\begin{table*}[t]
\centering\small
\caption{Isolated merge cost and recorded cumulative merging time. All times are milliseconds. SFA and Lookahead each use one auxiliary parameter copy; reductions are relative to Lookahead.}
\label{tab:merge-scaling}
\noindent\textbf{(a) Isolated operation benchmark: time per single merge}
\par\smallskip
\noindent\begin{tabular*}{\linewidth}{@{}l@{\extracolsep{\fill}}lrrr@{}}
\toprule
Parameter set & Implementation & SFA & Lookahead & SFA less time (\%)\\
\midrule
1.7B, full parameters & Per-tensor & 9.30 & 13.13 & 29.2\\
 & Batched & 5.09 & 8.60 & 40.8\\
8B, full parameters & Per-tensor & 39.91 & 56.09 & 28.8\\
 & Batched & 24.09 & 40.76 & 40.9\\
14B, full parameters & Per-tensor & 70.48 & 98.98 & 28.8\\
 & Batched & 43.44 & 73.55 & 40.9\\
\midrule
14B, adapters (5.24M) & Per-tensor & 1.092 & 1.657 & 34.1\\
 & Batched & 0.043 & 0.070 & 38.7\\
\bottomrule
\end{tabular*}
\par\smallskip
\noindent\footnotesize One H200, fp32; two blocks of 50 timed repetitions with method order reversed. CUDA-event means; initialization excluded. Per-tensor operations match the training implementation.
\par\bigskip
\noindent\normalsize\textbf{(b) Recorded training totals: sum across an entire four-task stream}
\par\smallskip
\noindent\small\begin{tabular*}{\linewidth}{@{}l@{\extracolsep{\fill}}crrr@{}}
\toprule
Training setting & Merges per run & SFA total & Lookahead total & Mean reduction (\%)\\
\midrule
1.7B full parameters & 16--17 & $153.0\pm5.7$ & $216.7\pm8.0$ & 29.4\\
1.7B adapters & 16 & $22.3\pm6.4$ & $22.4\pm0.1$ & 0.4\\
8B adapters & 16 & $20.3\pm0.1$ & $29.8\pm0.3$ & 32.0\\
14B adapters & 16 & $22.1\pm0.2$ & $32.8\pm0.8$ & 32.5\\
\bottomrule
\end{tabular*}
\par\smallskip
\noindent\footnotesize Mean $\pm$ sample SD over seeds 17, 42 and 123, original fixed configurations, order 1. Totals sum GPU-synchronized wall-clock merge timers from all four training stages; they are \emph{not} extrapolated from panel (a). Runs used different nodes/times, so reductions are descriptive accounting rather than a controlled speed comparison. The 1.7B full-parameter runs have 16, 16 and 17 merges; all adapter runs have 16.
\par\smallskip
\noindent\footnotesize The 8B and 14B full-parameter settings have operation benchmarks only, so no training totals are reported for them. Differences between panels also reflect timer and execution context. These totals count merging only, not the complete training loop.
\end{table*}

Table~\ref{tab:merge-scaling}(a) reports isolated operation costs. With per-tensor operations, SFA's absolute saving rises from 3.83\,ms at 1.7B to 28.50\,ms at 14B; batched operations save 3.51\,ms and 30.11\,ms, respectively. The 8B and 14B full-parameter rows measure operations only, not full-parameter training. Adapter costs depend on the trainable factors rather than the entire backbone. These absolute operation savings do not establish increasing fractional training savings with model size, since the cost of the rest of training also changes.

Panel (b) sums recorded merge timers across four-task streams using fixed configurations and seeds 17, 42, and 123. These are GPU-synchronized wall-clock totals from different nodes and times, not extrapolations from panel (a) or controlled runtime comparisons. The 1.7B adapter totals are nearly equal. Different timers and execution contexts also preclude treating panel (b) as a multiple of panel (a). Merging remains a small part of training: full-parameter 1.7B SFA merges occupy approximately 0.14\% of the loop. Appendix~\ref{app:highfrequency} separately measures cumulative merge cost during controlled, same-GPU Law training across much higher frequencies.
}

\section{Math-to-Law Selection and Replication}
\label{app:m2l-protocol}
{
\paragraph{Development selection.}
In order to compare retention without simply favoring a model that learns less of the new task, we first choose configurations under a common Law-accuracy requirement. The retention comparison evaluates three strengths for both SFA and Lookahead, $\{.25,.5,.75\}$, at $p=.25$ and learning rate $10^{-5}$. Sequential fine-tuning receives three learning rates, $\{10^{-5},5\cdot10^{-6},2.5\cdot10^{-6}\}$. For each method, we maximize development GSM8K subject to Law accuracy at least $L_{\mathrm{ref}}-\delta$, where $L_{\mathrm{ref}}=.90843$ is sequential fine-tuning's reference development score and $\delta\in\{.005,.010,.020\}$. Ties within one GSM8K development question (.002) prefer the fixed configuration, then higher Law accuracy, then the larger coefficient. An infeasible method remains infeasible. Table~\ref{tab:m2ldev} gives the full grid.

\begin{table}[t]
\centering\small
\caption{Development results underlying configuration selection for Table~\ref{tab:newm2l} (seed 17; accuracies in \%). Bold rows identify the configuration with the highest Math accuracy among those achieving at least 89.843\% Law accuracy. These scores determine selection given one seed; the five-seed test results provide the performance comparison.}
\label{tab:m2ldev}
\begin{tabular}{@{}llrr@{}}
\toprule
Method & Coefficient / LR & Math & Law\\
\midrule
SFA & $\beta_{\rm end}=.25$ & 73.2 & 91.523\\
 & $\mathbf{.5}$ & \textbf{76.4} & \textbf{90.339}\\
 & $.75$ & 75.4 & 88.722\\
Lookahead & $\beta=.25$ & 72.2 & 91.730\\
 & $\mathbf{.5}$ & \textbf{76.6} & \textbf{90.471}\\
 & $.75$ & 76.0 & 88.846\\
Sequential fine-tuning & $10^{-5}$ & 70.2 & 90.843\\
 & \boldmath$5\cdot10^{-6}$ & \textbf{75.4} & \textbf{90.321}\\
 & $2.5\cdot10^{-6}$ & 75.6 & 89.536\\
\bottomrule
\end{tabular}
\end{table}

The middle and loose requirements select SFA $\beta_{\mathrm{end}}=.5$, Lookahead $\beta=.5$, and sequential fine-tuning learning rate $5\cdot10^{-6}$. The tight requirement selects SFA $.25$, Lookahead $.5$, and sequential fine-tuning $10^{-5}$. SFA at $\beta_{\mathrm{end}}=.5$ misses the tight threshold by .00004 in accuracy (90.339\% versus 90.343\%); this is a formal threshold miss rather than a material adaptation deficit. Sequential fine-tuning at $5\cdot10^{-6}$ misses by .00021, and Lookahead $.75$ clears the loose threshold by .00003. The margins are therefore included alongside the selections, since crossing a threshold by this amount need not correspond to a meaningful difference in adaptation.

\paragraph{Search history.}
The seed-17 exploratory search evaluated six SFA candidates, six endpoint coefficients, three Lookahead frequencies, one sequential fine-tuning run, and three evaluation-only EMA decays. The shared $p=.25$ was chosen in this search. The retention comparison then evaluated equal three-point candidate grids for SFA, Lookahead, and sequential fine-tuning. Values for the fixed SFA and Lookahead configurations and the reference sequential fine-tuning development run were known when its selection rule was specified. For seeds 17, 42, and 123, selected SFA and Lookahead test evaluations were also available before the retention comparison was formulated, whereas lower-learning-rate sequential fine-tuning was trained under that comparison. We distinguish these runs from the prospectively specified replication seeds below.

\paragraph{Prospective replication.}
To test the selected configurations on independently specified runs, seeds 7 and 2026 were recorded before training and run to completion without retuning. Each seed shares one Math-FT checkpoint and one Law schedule across methods. Table~\ref{tab:m2lpaired} distinguishes the exploratory seed group (17, 42, 123) from these prospective replications. SFA improves Law over Lookahead in all five runs and Math in four of five. Relative to selected sequential fine-tuning, SFA improves Math in three of five runs and Law in four of five, with the highest pooled mean on both metrics.

\begin{table*}[t]
\centering\small
\caption{Math$\to$Law per-seed test accuracy (\%). Bold indicates the highest accuracy for each task within each seed. Seeds 7 and 2026 are prospective replications of the frozen middle-requirement configurations.}
\label{tab:m2lseeds}
\begin{tabular}{@{}lrrrrrr@{}}
\toprule
&\multicolumn{2}{c}{SFA}&\multicolumn{2}{c}{Lookahead}&\multicolumn{2}{c}{\shortstack{Sequential fine-tuning\\selected LR}}\\
Seed & GSM8K & Law & GSM8K & Law & GSM8K & Law\\
\midrule
17 & \textbf{71.42} & \textbf{90.22} & 71.19 & 89.59 & 71.11 & 89.92\\
42 & \textbf{71.57} & \textbf{90.49} & 70.96 & 90.23 & 70.20 & 90.39\\
123 & \textbf{73.77} & \textbf{90.57} & 73.09 & 89.61 & 71.42 & 90.23\\
7 (prospective) & 71.65 & \textbf{90.12} & 70.66 & 89.90 & \textbf{71.87} & 90.00\\
2026 (prospective) & 70.74 & 89.49 & \textbf{71.34} & 89.40 & 71.27 & \textbf{89.81}\\
\bottomrule
\end{tabular}
\end{table*}
\begin{table*}[t]
\centering\small
\caption{Paired Math$\to$Law differences in percentage points (mean $\pm$ sample SD). Positive values favor SFA. No significance or equivalence test is implied.}
\label{tab:m2lpaired}
\begin{tabular}{@{}llrr@{}}\toprule
Seed group & Comparison & GSM8K & Law\\\midrule
Seeds 17, 42, 123 & SFA minus Lookahead & $0.51\pm0.24$ & $0.62\pm0.35$\\
Seeds 17, 42, 123 & SFA minus sequential fine-tuning & $1.34\pm1.02$ & $0.25\pm0.13$\\
Seeds 7, 2026 & SFA minus Lookahead & $0.19\pm1.13$ & $0.16\pm0.10$\\
Seeds 7, 2026 & SFA minus sequential fine-tuning & $-0.38\pm0.21$ & $-0.10\pm0.31$\\
All five & SFA minus Lookahead & $0.38\pm0.61$ & $0.43\pm0.36$\\
All five & SFA minus sequential fine-tuning & $0.65\pm1.19$ & $0.11\pm0.26$\\
\bottomrule\end{tabular}\end{table*}

\paragraph{Endpoint and EMA operating points.}
On seeds 17, 42, and 123, endpoint interpolation at $\beta=.5$ achieves $72.25\pm1.09$\% GSM8K and $88.47\pm0.09$\% Law. EMA with decay .999 achieves $72.48\pm1.25$\% and $87.25\pm0.47$\%, respectively. SFA's paired Law gain over the endpoint is $1.95\pm0.24$ points, while the GSM8K difference is $0.00\pm0.35$ points. Endpoint and EMA use only the matched three-seed set (Table~\ref{tab:m2lendpoint}).
Figure~\ref{fig:newm2l} uses this same three-seed set for every method, whereas Table~\ref{tab:newm2l} pools five seeds for the selected SFA, Lookahead, and sequential fine-tuning configurations. All error terms are sample SD. Table~\ref{tab:m2lpaired} reports differences computed within each seed.
\begin{table}[t]
\centering\small
\caption{Matched Qwen3 Math$\to$Law averaging and fine-tuning comparison (\%, seeds 17, 42, 123, mean $\pm$ sample SD). These are the non-replay comparison points in Figure~\ref{fig:newm2l}; Table~\ref{tab:m2l-strength} gives the complete SFA sweep.}
\label{tab:m2lendpoint}
\begin{tabular}{@{}lrr@{}}\toprule
Method & GSM8K & Law mean\\\midrule
SFA, $\beta_{\rm end}=.5$ & $72.25\pm1.32$ & $90.43\pm0.19$\\
\shortstack[l]{Endpoint interpolation\\$\beta=.5$} & $72.25\pm1.09$ & $88.47\pm0.09$\\
EMA, decay .999 & $72.48\pm1.25$ & $87.25\pm0.47$\\
\shortstack[l]{Lookahead\\$\beta=.5$} & $71.75\pm1.17$ & $89.81\pm0.36$\\
\shortstack[l]{Sequential fine-tuning\\learning rate $10^{-5}$} & $67.07\pm1.60$ & $91.30\pm0.41$\\
\shortstack[l]{Sequential fine-tuning\\learning rate $5\cdot10^{-6}$} & $70.91\pm0.63$ & $90.18\pm0.24$\\
\bottomrule\end{tabular}
\end{table}
}

\subsection{Replay and SFA strength trade-offs}\label{app:m2l-tradeoffs}
{
These exploratory comparisons use Qwen3-1.7B and seeds 17, 42, and 123, with the same per-seed Math-trained checkpoints and Law schedules as Figure~\ref{fig:newm2l}. They are separate from the five-seed summary in Table~\ref{tab:newm2l}. Every reported setting is test-evaluated, rather than selecting the best test point. Values are means and sample standard deviations over the three seeds; differences from the starting checkpoint are calculated within seed before summarizing.

\paragraph{SFA strength.}
We evaluate $\beta_{\mathrm{end}}\in\{.25,.50,.75,.90,.99\}$ at four merges per task, using $\beta=1-(1-\beta_{\mathrm{end}})^{1/4}$ at each merge. The learning rate remains $10^{-5}$; data order, optimizer, precision, and training budget remain fixed. The final merge is applied exactly once. The $.25$ and $.50$ runs for all seeds, and the $.75$ run for seed 17, reuse existing checkpoints; the remaining eight runs are new. The candidate set was fixed before these new runs, and no test winner was selected. This follow-up was motivated by the preceding comparisons and is not an independent confirmation of reused results.

\begin{table*}[t]
\centering\small
\caption{Qwen3-1.7B SFA strength sweep (three seeds; accuracy in \%, changes in percentage points; mean $\pm$ sample SD). Positive Math forgetting denotes a decline from the Math-trained start. Law gain is relative to that same starting model.}
\label{tab:m2l-strength}
\begin{tabular}{@{}lrrrrr@{}}\toprule
$\beta_{\mathrm{end}}$ & Per-merge $\beta$ & GSM8K & Law mean & Math forgetting & Law gain\\\midrule
Math-trained start & --- & $74.78\pm0.86$ & $35.15\pm0.92$ & --- & ---\\
$.25$ & $.069$ & $69.93\pm0.57$ & $91.22\pm0.30$ & $4.85\pm1.42$ & $56.06\pm0.92$\\
$.50$ & $.159$ & $72.25\pm1.32$ & $90.43\pm0.19$ & $2.53\pm0.50$ & $55.27\pm1.01$\\
$.75$ & $.293$ & $73.04\pm1.26$ & $88.05\pm0.29$ & $1.74\pm0.66$ & $52.90\pm0.82$\\
$.90$ & $.438$ & $74.37\pm0.89$ & $80.98\pm2.25$ & $0.40\pm0.04$ & $45.82\pm1.40$\\
$.99$ & $.684$ & $74.96\pm1.15$ & $54.96\pm6.02$ & $-0.18\pm0.31$ & $19.81\pm5.53$\\
\bottomrule\end{tabular}
\end{table*}

Stronger anchoring reduces observed Math forgetting, but the Law cost rises sharply at $.90$ and $.99$. Terms of Service is especially sensitive: its mean accuracy falls to 80.29\% and 30.03\%, with sample SDs of 6.47 and 14.15 points, respectively, compared with 12.57\% at the start. Thus, reaching the starting model's Math accuracy does not imply preserving the same degree of Law adaptation. Table~\ref{tab:m2l-strength} reports the complete sweep, while Figure~\ref{fig:newm2l} compares the selected $\beta_{\mathrm{end}}=.5$ configuration with other replay-free updates.

\paragraph{Comparison with replay.}
We compare averaging with rehearsal. Rehearsal interleaves past-task examples with new-task data; on Qwen3, we sample from a fixed buffer of 500 Math examples to supply 1\%, 5\%, or 10\% of the combined input tokens during Law fine-tuning. SFA at $\beta_{\mathrm{end}}=.5$ achieves higher Law accuracy than all three replay configurations, without revisiting Math examples (Table~\ref{tab:m2l-replay}). Increasing the replay proportion further improves Math but reduces Law accuracy, widening SFA's mean Law advantage to 1.67 points at 10\% replay. SFA therefore offers a replay-free alternative when new-domain adaptation (Law) takes priority, while replay favors stronger retention (Math) at these settings. On Llama~2, SFA approaches replay's retention while achieving slightly higher Law accuracy at the displayed settings (Figure~\ref{fig:llama7b_mathlaw}(a)).

\paragraph{Bounded-memory replay.}
For each seed, a fixed reservoir of 500 examples is sampled reproducibly from the contamination-screened Math training pool and shared across replay proportions. No development or test examples enter this reservoir. We add Math replay to the unchanged Law schedule, targeting input-token fractions of 1\%, 5\%, and 10\% of the combined training inputs. Replay uses learning rate $5\cdot10^{-6}$, transferred from development-selected sequential fine-tuning, rather than independently tuned for replay. The other optimizer and evaluation settings match that baseline. The three existing 10\% runs are reused; all three proportions are reported without choosing a test winner.

Current-task exposure and optimizer-update counts are preserved, but total training compute is not held fixed. The schedule adds approximately 0.10M, 0.53M, and 1.11M Math input tokens to 10M Law tokens. At 1\%, roughly 400 of the 738--740 updates contain replay, distributed through a deterministic token-budget schedule. Gradients use a token-mean supervised loss over the combined examples, excluding masked and padding tokens. Because Law examples contain short supervised answers, replay's share of supervised tokens is approximately 11\%, 38\%, and 56\%, substantially larger than its input share. These are token shares, not measured gradient-norm contributions.

\begin{table*}[t]
\centering\small
\caption{Qwen3-1.7B bounded-memory replay (three seeds, fixed 500-example reservoir). GSM8K and Law mean show mean $\pm$ sample SD; individual Law columns show means. Input and supervised shares refer to different token counts. All accuracies and shares are percentages.}
\label{tab:m2l-replay}
\begin{tabular}{@{}lrrrrrr@{}}\toprule
Input share & Supervised share & GSM8K & Law mean & CaseHOLD & ToS & Overruling\\\midrule
1 & $\approx11$ & $74.50\pm0.97$ & $89.74\pm0.22$ & 81.02 & 93.09 & 95.11\\
5 & $\approx38$ & $74.93\pm0.82$ & $89.01\pm0.16$ & 80.12 & 92.34 & 94.56\\
10 & $\approx56$ & $75.76\pm0.85$ & $88.76\pm0.23$ & 79.67 & 91.83 & 94.78\\
\bottomrule\end{tabular}
\end{table*}

Relative to SFA at $\beta_{\mathrm{end}}=.5$, replay's paired Math gains are $2.25\pm1.31$, $2.68\pm1.08$, and $3.51\pm0.94$ points; its Law differences are $-0.69\pm0.24$, $-1.42\pm0.24$, and $-1.67\pm0.36$ points. Each direction holds on all three seeds. On the reported means, 1\% replay exceeds SFA at $.75$ and $.90$ on both tasks, and 10\% replay exceeds SFA at $.99$. The SFA sweep therefore illustrates replay-free control of the trade-off, rather than superiority over replay at high-retention settings.

\paragraph{Resource accounting.}
The replay reservoir occupies approximately 0.5\,MB of token IDs, or 0.36\,MB of text. Peak GPU memory is 84--85\,GB. The nine replay configurations used 2.54 GPU-hours, including evaluation and checks; the new SFA strength runs and evaluations used 2.23 GPU-hours. Avoiding stored examples does not always imply lower total memory than using a small replay buffer, because full-parameter SFA retains a model-sized reference. 
}

\section{Adapter Sensitivity and Hybrids}
\label{app:hybrids}
{
\subsection{Model size}
\label{app:model-size}
Table~\ref{tab:sizesfull} tests whether fixed SFA and Lookahead configurations transfer across model sizes, with replay and the AIMMerging port as more elaborate comparisons. At 14B, both interpolation methods exceed their means; at 1.7B the replay-based methods do better, and the differences narrow at 32B. Additional development-selected configurations are discussed in Appendix~\ref{app:coefficient-frequency}, separately from this no-retuning comparison. Each entry summarizes three seeds; error terms are sample standard deviations, not standard errors or confidence intervals.
\begin{table*}[t]
\centering\small
\caption{Final average test accuracy in adapter continual learning (\%, three seeds, mean $\pm$ sample SD). Configurations fixed at 1.7B and transferred across sizes, with matched current-task exposure. Bold marks the highest fixed-configuration mean at each size. The 32B runs use activation checkpointing.}
\label{tab:sizesfull}
\begin{tabular}{@{}lrrrr@{}}
\toprule
Method & 1.7B & 8B & 14B & 32B\\
\midrule
Sequential fine-tuning & $70.6\pm0.8$ & $76.2\pm0.5$ & $77.7\pm1.1$ & $78.1\pm1.2$\\
Replay (buffer) & $74.6\pm1.0$ & $77.5\pm0.3$ & $78.6\pm0.8$ & $78.7\pm0.3$\\
AIMMerging (ported) & $\mathbf{74.8}\pm0.2$ & $\mathbf{77.8}\pm0.6$ & $78.9\pm0.6$ & $79.0\pm0.3$\\
Lookahead, fixed coefficient & $72.8\pm0.6$ & $76.7\pm1.3$ & $79.6\pm0.6$ & $79.1\pm0.4$\\
\textbf{SFA}, fixed strength & $72.1\pm1.9$ & $77.4\pm0.9$ & $\mathbf{79.9}\pm0.3$ & $\mathbf{79.2}\pm0.4$\\
\bottomrule
\end{tabular}
\end{table*}

\subsection{Task order}
\label{app:task-order}

To examine whether SFA's benefits persist
under a different task sequence, we exchange the positions
of Amazon and Yahoo Answers. The resulting order is
DBpedia $\to$ Yahoo Answers $\to$ Amazon $\to$ AG News.
We use Qwen3-14B with a shared LoRA adapter, keeping
the method configurations unchanged from the original
order in Figure~\ref{fig:cost}.

As Table~\ref{tab:alternative-order} shows,
SFA attains 79.9\% final average accuracy on this alternative order, compared with 79.6\% for replay, 79.3\% for Lookahead, and 78.9\% for AIMMerging (Table~\ref{tab:alternative-order}); these remain fixed configurations. SFA also retains 97.4\% DBpedia accuracy after all tasks, versus 94.8--95.9\% for the other configurations. The result illustrates that replay-free averaging remains competitive beyond one ordering, rather than establishing an across-settings ranking between the simple methods.

\begin{table}[t]
\centering\small
\caption{Qwen3-14B shared-LoRA continual
learning with an alternative task order:
DBpedia $\to$ Yahoo Answers $\to$ Amazon $\to$ AG News.
Final average test accuracy (\%, mean $\pm$ sample SD
over seeds 17, 42, and 123). Methods retain the fixed
configurations used in Figure~\ref{fig:cost}:
SFA uses $\beta_{\mathrm{end}}=.5$ and Lookahead
uses $\beta=.5$. Bold marks the highest mean.}
\label{tab:alternative-order}
\begin{tabular}{@{}lr@{}}
\toprule
Method & Final average (\%)\\
\midrule
Sequential fine-tuning & $79.0\pm0.2$\\
Replay (buffer) & $79.6\pm0.2$\\
SFA & $\mathbf{79.9}\pm0.4$\\
Lookahead & $79.3\pm0.4$\\
AIMMerging (ported) & $78.9\pm1.0$\\
\bottomrule
\end{tabular}
\end{table}

\subsection{Coefficient and frequency selection}
\label{app:coefficient-frequency}
The main adapter comparison asks whether fixed, transferable interpolation settings can rival more elaborate methods, not whether either interpolation policy is optimally tuned. We therefore also examine sensitivity to coefficient selection. At 14B, a prospective three-point development grid for each method maximizes earlier-task accuracy subject to a current-task acquisition requirement; the reference is sequential fine-tuning's seed-17 development acquisition (.808), with allowable deficits of .005, .010, or .020. All thresholds select SFA $\beta_{\mathrm{end}}=.5$ and Lookahead $\beta=.75$. With coefficients selected on development data and held fixed for three test seeds, SFA obtains $79.9\pm0.3$\% and Lookahead $80.7\pm0.1$\% final average accuracy. The selected Lookahead result strengthens the empirical case for low-overhead interpolation.

The frequency study evaluates additional SFA merge counts $K\in\{1,2,8\}$ at strengths $\beta_{\mathrm{end}}\in\{.5,.75\}$. Selection uses seed-17 development results. The middle requirement selects $K=2$, $\beta_{\mathrm{end}}=.5$. Under the loose requirement, three candidates fall within the .002 tie margin, and the implemented selector chooses $K=2$, $\beta_{\mathrm{end}}=.75$. Interpreting the rule's ambiguous ``ties to .5'' as strength $\beta_{\mathrm{end}}=.5$ would instead select the middle-requirement configuration. We document the executed choice as a protocol deviation and retain the alternative for sensitivity analysis.

At 1.7B, seed-17 development selection chooses $\beta_{\mathrm{end}}=.75$ for SFA and $\beta=.75$ for Lookahead. Held fixed across three test seeds, these achieve $74.0\pm0.3$\% and $76.7\pm0.2$\% final accuracy, respectively. An exploratory learning-rate-weighted SFA schedule does not improve development accuracy. These settings are distinct from the fixed configurations transferred across model sizes; their stronger results reinforce that tuning can change conclusions about both simple methods.

\subsection{Task-wise retention}
\label{app:taskwise-retention}
A final average can obscure differences between the oldest and most recently learned tasks. We therefore examine each task after all four stages of the 14B DBpedia-first stream (Table~\ref{tab:taskwise14b}). SFA's reference is the incoming model at each stage, not the initial model throughout the stream. The reported values are a task-wise analysis of the completed test runs, with no additional configuration selection.

Under the fixed settings of Figure~\ref{fig:cost}, SFA has higher final Amazon accuracy on all three seeds, with paired gains of $3.00$, $0.04$, and $0.40$ points. Its DBpedia gains are $1.77$, $-0.23$, and $0.87$ points. On both tasks, signed forgetting is lower for SFA in every seed. Final accuracy and forgetting need not give the same ranking, since the latter also depends on the accuracy achieved before subsequent tasks were learned. Yahoo and AGNews favor fixed-coefficient Lookahead in mean final accuracy, so the first-task gains are partly obscured by averaging over the complete stream.

Under the fixed settings, SFA preserves DBpedia and Amazon better than fixed Lookahead, whereas coefficient-selected Lookahead performs better on these tasks. It reaches 77.63\% final earlier-task accuracy versus 76.52\% for SFA. This dependence on configuration is compatible with the broader result that both methods can mitigate forgetting without rehearsal. Appendix~\ref{app:coefficient-frequency} gives the coefficient-selection protocol.
\begin{table*}[t]
\centering\small
\caption{Task-wise retention after the complete 14B adapter stream, DBpedia $\to$ Amazon $\to$ Yahoo $\to$ AGNews (test, seeds 17/42/123; mean $\pm$ sample SD). Accuracy is measured after AGNews, not immediately after learning each task. Signed forgetting is the highest earlier accuracy minus final accuracy; lower is better. The last task has no subsequent stage and hence no forgetting entry. Fixed configurations are those in Figure~\ref{fig:cost}; selected Lookahead comes from the equal three-point coefficient study.}
\label{tab:taskwise14b}
\setlength{\tabcolsep}{5pt}
\begin{tabular}{@{}llrrrr@{}}\toprule
Metric & Configuration & DBpedia & Amazon & Yahoo & AGNews\\\midrule
Final accuracy (\%) & SFA, $\beta_{\rm end}=.5$ & $97.04\pm0.09$ & $61.41\pm0.41$ & $71.11\pm0.96$ & $90.00\pm0.30$\\
 & Lookahead, $\beta=.5$ (fixed) & $96.24\pm0.95$ & $60.27\pm1.22$ & $71.57\pm0.99$ & $90.28\pm0.54$\\
 & Lookahead, $\beta=.75$ (selected) & $98.06\pm0.19$ & $61.97\pm0.64$ & $72.87\pm0.48$ & $89.90\pm0.49$\\
\midrule
Forgetting (points) & SFA, $\beta_{\rm end}=.5$ & $1.37\pm0.11$ & $3.23\pm0.58$ & $1.53\pm0.48$ & ---\\
 & Lookahead, $\beta=.5$ (fixed) & $2.45\pm0.98$ & $4.79\pm1.12$ & $1.22\pm0.68$ & ---\\
 & Lookahead, $\beta=.75$ (selected) & $0.48\pm0.21$ & $2.36\pm1.52$ & $1.00\pm0.21$ & ---\\
\bottomrule\end{tabular}
\end{table*}

\paragraph{Interpreting the effect of the reference policy.}
Fixed anchoring does not necessarily produce less displacement than moving-reference averaging. For illustration, suppose each of four optimization blocks adds the same vector $d$ from its starting weights, beginning at anchor $a$. With $q=1-\beta$, SFA ends at $a+(q+q^2+q^3+q^4)d$, whereas Lookahead ends at $a+4\alpha d$. Our selected settings have $q=2^{-1/4}$ and $\alpha=.25$, giving displacements of approximately $2.64d$ and $d$, respectively. This simplified calculation does not model Adam dynamics, but shows why a moving reference need not move farther. Nor does Euclidean distance alone determine retained accuracy: updates in different directions can affect old-task predictions differently. In the reported task streams, SFA refreshes its reference between tasks. Before the final AGNews stage, its mean DBpedia accuracy is 97.70\%, versus 98.41\% for selected Lookahead, so the methods enter that stage with different retained solutions. Refreshing the reference is nevertheless a configuration choice rather than a requirement of the averaging update. A user wishing to prioritize mathematical knowledge, for example, can retain a Math-trained checkpoint as the reference while continuing to train the working model on subsequent tasks. SFA thus allows the reference checkpoint, alongside the averaging interval and weight, to reflect the user's retention priorities. Unlike standard Lookahead's evolving reference, this checkpoint remains an explicit target throughout the chosen sequence. The choice directs corrections toward a particular learned solution; its effect on retained skills and subsequent learning must still be evaluated.

\subsection{Replay hybrids}
To examine whether averaging also complements replay
under parameter-efficient training, we evaluate SFA-R on
Qwen3-1.7B with LoRA across the DBpedia $\to$ Amazon $\to$
Yahoo Answers $\to$ AG News stream. This extends the
full-parameter SFA-R comparison in
Table~\ref{tab:fullparam-replay-cost} to the adapter setting.

SFA-R combines the fixed-configuration adapter SFA rule with
the replay baseline's eight additional examples per update.
At 1.7B it reaches $75.5\pm.4$\% final average, 77.3\%
acquisition, and 2.5 points forgetting. Its paired
final-average gains are $0.6\pm0.2$ points over AIMMerging
and $0.85\pm1.15$ over replay; the gain over replay is
positive in two of three seeds. SFA-R adds 1.20M
replay-gradient tokens to the 6.01M current-task tokens,
and no monitoring tokens.

Together with the full-parameter results, this
comparison shows that intermediate averaging can complement
rehearsal under both training regimes. The adapter result
provides a smaller and less consistent gain over replay,
while remaining competitive with the tested adaptive
merging implementation.

\subsection{Choosing a reference to prioritize earlier knowledge}\label{app:reference-choice}

Beyond the averaging interval and strength, SFA allows
the reference checkpoint to reflect which earlier knowledge the
user wishes to preserve. We examine this choice using Qwen3-14B
with LoRA, designating Amazon sentiment classification as the
task to protect. Within each seed, both variants begin from the
same checkpoint obtained by sequentially learning DBpedia and
Amazon, then continue training on Yahoo Answers and AG News.
Standard SFA refreshes its reference at each task boundary,
whereas SFA-Ref retains the post-Amazon checkpoint throughout
the continuation without resetting the working model.

Table~\ref{tab:sfa-reference-choice} shows that the
selected SFA-Ref configuration achieves higher mean accuracy
on both earlier tasks. Its paired Amazon gain is
$3.3\pm3.1$ percentage points, accompanied by a
$1.8\pm0.3$-point reduction on Yahoo Answers. This illustrates
a configuration that favors earlier-task retention over
subsequent-task performance. The comparison includes
development-selected strengths of $\beta_{\mathrm{end}}=.25$
for refreshed SFA and $.5$ for SFA-Ref, so the test difference
reflects both strength and reference policy. At matched
strengths of $.25$, $.5$, and $.75$, the seed-17 development
comparisons also favor SFA-Ref on Amazon, by 3.2, 3.8, and
0.9 points, respectively.

\begin{table*}[t]
\centering\small
\caption{Reference selection in Qwen3-14B LoRA
continual learning. Both variants continue from an identical
post-Amazon checkpoint within each seed and subsequently learn
Yahoo Answers and AG News. Values are final test accuracies
(\%, mean $\pm$ sample SD over seeds 17, 42, and 123).
Amazon is the designated protected task; its starting accuracy
is $64.4\pm0.7$\%. Strengths are selected on seed-17 development
data. Bold marks the higher mean in each column.}
\label{tab:sfa-reference-choice}
\setlength{\tabcolsep}{4pt}
\begin{tabular}{@{}lcrrrrr@{}}
\toprule
Method & $\beta_{\mathrm{end}}$
& DBpedia & Amazon & Yahoo Answers & AG News & Final average\\
\midrule
Refreshed SFA & $.25$
& $96.7\pm0.6$
& $58.0\pm3.8$
& $\mathbf{71.3}\pm0.9$
& $\mathbf{89.9}\pm0.4$
& $79.0\pm1.0$\\
SFA-Ref & $.50$
& $\mathbf{97.4}\pm0.2$
& $\mathbf{61.3}\pm1.6$
& $69.5\pm1.0$
& $89.6\pm0.3$
& $\mathbf{79.4}\pm0.7$\\
\bottomrule
\end{tabular}
\end{table*}

\subsection{Merge diagnostics}
In order to better understand what happens at an averaging step, we measure development loss on fixed examples immediately before and after each merge in one 1.7B stream. Later merges often reduce loss on both past and current tasks, while an early merge can briefly increase some losses. We also compare factor-wise interpolation with interpolation of the effective dense update at three merge points. The factor-wise version performs at least as well on the measured losses. These observations do not support attributing the adapter results to harmful late merges or to factorization alone.

}

\subsection{Synchronization frequency and merge cost}\label{app:highfrequency}
{
\begin{table}[t]
\centering\small
\caption{Full-parameter Math$\to$Law frequency controls: SFA minus Lookahead in accuracy (points, five-seed mean $\pm$ SD) and measured cumulative merge-time savings for SFA on one H200. These pairwise differences diagnose reference-policy and implementation effects; they are not the main comparison to rehearsal-based methods.}
\label{tab:highfrequency}
\setlength{\tabcolsep}{3pt}
\begin{tabular}{@{}rrrr@{}}
\toprule
$k$ (interval) & $\Delta$Math & $\Delta$Law & Merge saving\\
\midrule
185 & $+0.03\pm0.46$ & $+0.43\pm0.26$ & 0.015\,s (28.9\%)\\
5 & $+0.14\pm0.68$ & $+0.46\pm0.28$ & 0.563\,s (28.9\%)\\
1 & $+0.39\pm0.51$ & $+0.30\pm0.23$ & 2.877\,s (29.3\%)\\
\bottomrule
\end{tabular}
\end{table}

\paragraph{Protocol and selection.}
We study full-parameter Qwen3-1.7B Math$\to$Law using the same per-seed Math checkpoints and Law schedules as the domain comparison. The training budget, learning rate $10^{-5}$, warmup and decay, precision, batches, and optimizer handling remain unchanged. We merge every 185, 5, or 1 updates, always applying the final merge once. This gives 4, 148, and 740 merges for seeds 17, 123, 7, and 2026. Seed 42 has 738 updates and consequently 738 every-update merges; its other schedules still have 4 and 148 events. Intermediate held-out loss evaluations are disabled for both methods.

At each higher frequency, seed-17 development selection gives SFA $\beta_{\mathrm{end}}\in\{.25,.5,.75\}$ and Lookahead per-sync $\beta\in\{.25,.5,.75\}$. We maximize GSM8K subject to Law mean at least .898426, carrying over the middle adaptation requirement. Ties within one GSM8K example (.002) prefer coefficient .5, then higher Law, then the larger coefficient; infeasibility is not relaxed. Selection chooses .5 for both methods at both higher frequencies. These coefficients do not represent equal regularization: SFA uses per-merge $\beta=1-(1-\beta_{\mathrm{end}})^{1/K}$, whereas Lookahead retains its selected weight at each synchronization. Holding SFA's cumulative contraction fixed redistributes corrections across training, rather than simply increasing total anchoring strength.

Before replication or test evaluation, we freeze the selected configurations, four-merge controls, and analysis. Seeds 42, 123, 7, and 2026 are then run without retuning. Every pair shares its Math checkpoint and Law data order. The five-seed test summary includes seed 17, whose development checkpoints are reused; Figure~\ref{fig:highfrequency} distinguishes this seed from the four replication seeds. No training stops or changes according to intermediate rankings. Checks verify initialization, immutable SFA references, and exact merge counts.

\begin{table*}[t]
\centering\small
\caption{Qwen3-1.7B full-parameter Math$\to$Law fine-tuning: controlled cumulative merge time and complete Law-stage training-loop time (seconds), with two repetitions on the same H200 per merge interval using seed 17. Training-loop time excludes loading, saving, and evaluation. Merge time uses queued CUDA events; loop time uses synchronized wall-clock timing.}
\label{tab:hf-timing}
\begin{tabular*}{\textwidth}{@{}lr@{\extracolsep{\fill}}rrrr@{}}
\toprule
 & & \multicolumn{2}{c}{Cumulative merging} & \multicolumn{2}{c}{Training loop}\\
\cmidrule(lr){3-4}\cmidrule(lr){5-6}
Method & Interval & Rep.\ 1 & Rep.\ 2 & Rep.\ 1 & Rep.\ 2\\
\midrule
SFA & 185 & 0.037 & 0.037 & 585.3 & 586.1\\
Lookahead & 185 & 0.052 & 0.053 & 585.0 & 586.1\\
\addlinespace[3pt]
SFA & 5 & 1.388 & 1.385 & 583.9 & 589.9\\
Lookahead & 5 & 1.960 & 1.939 & 585.2 & 588.1\\
\addlinespace[3pt]
SFA & 1 & 6.931 & 6.925 & 596.2 & 596.7\\
Lookahead & 1 & 9.799 & 9.810 & 599.4 & 601.1\\
\bottomrule
\end{tabular*}
\end{table*}

\begin{figure*}[t]
\centering
\includegraphics[width=\textwidth]{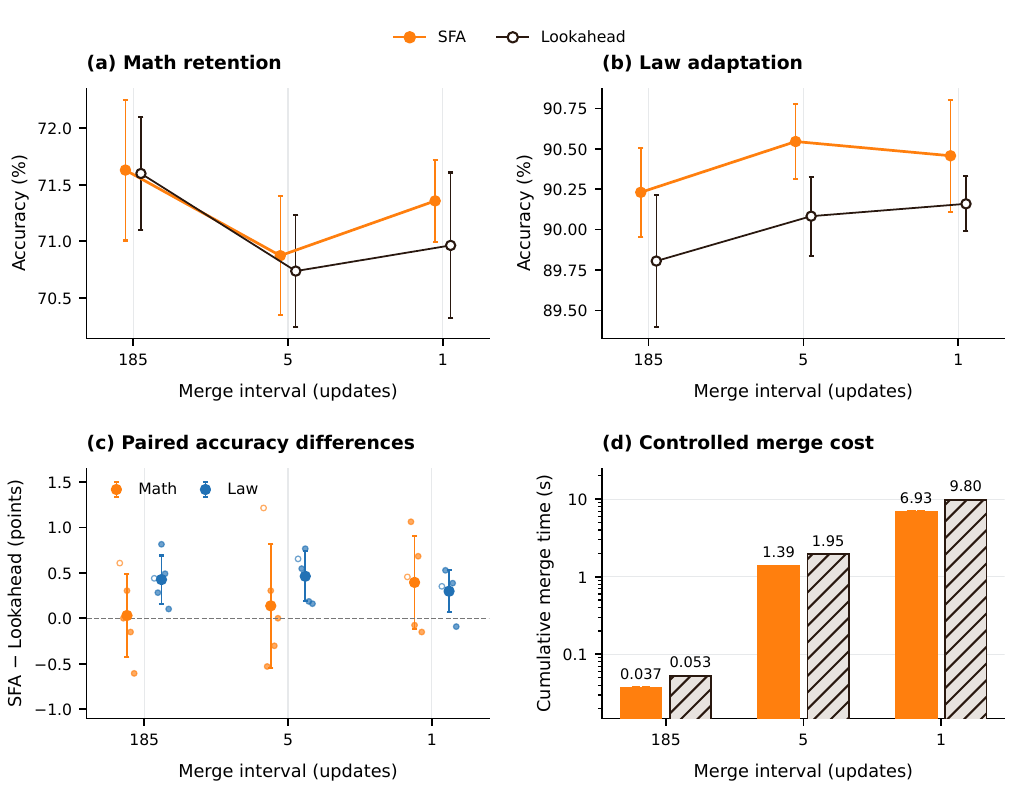}
\caption{Full-parameter Qwen3-1.7B Math$\to$Law across merge intervals. Top: test accuracy (five-seed mean $\pm$ sample SD). Bottom left: paired SFA-minus-Lookahead differences; small marks show individual seeds, with seed 17 hollow because it supplied development selection. Bottom right: controlled cumulative merge time in seconds on a logarithmic vertical scale (two repetitions per method). SFA uses $\beta_{\mathrm{end}}=.5$ and Lookahead uses per-sync $\beta=.5$. Timings cover only merging.}
\label{fig:highfrequency}
\end{figure*}

\paragraph{Retained accuracy and adaptation.}
Figure~\ref{fig:highfrequency} and Table~\ref{tab:highfrequency} provide a finer-grained diagnostic of the simple interpolation rules. SFA has higher Law accuracy in 14 of 15 seed-frequency pairs, while Math differences are small and mixed. Four held-out replication seeds preserve the Law trend in 11 of 12 comparisons.

Higher frequency does not reliably increase the difference between the two rules. Among the four replication seeds, moving from four to every-update merges changes the SFA-minus-Lookahead gap by $0.49\pm0.63$ Math points and $-0.14\pm0.31$ Law points. Both methods lose some Math accuracy relative to their four-merge controls while gaining Law accuracy. 

\paragraph{Controls and experiment history.}
The four-merge controls are new training runs under this study's code and instrumentation. They are kept separate from Table~\ref{tab:newm2l}, whose runs use synchronized merge timers and intermediate held-out evaluations. Across five seeds, the paired Law gain reproduces ($0.43\pm0.26$ points here versus $0.43\pm0.36$ there), while the small Math gain is less stable ($0.03\pm0.46$ versus $0.38\pm0.61$). 

\paragraph{Controlled timing.}
For each interval, one H200 allocation runs four complete Law trainings from the same seed-17 start, in the order SFA, Lookahead, Lookahead, SFA. Queued CUDA-event pairs measure each merge without inserting per-event synchronization; we sum their device times over the run. Synchronized wall-clock measurements separately cover the complete training loop, excluding model loading, saving, and evaluation. Table~\ref{tab:hf-timing} reports both repetitions. The percentage reduction remains close to 29\%; it is the absolute saving that grows with the number of events (Table~\ref{tab:highfrequency}).

At every-update merging, the controlled mean training-loop times are 596.5\,s for SFA and 600.2\,s for Lookahead; merging occupies 1.16\% and 1.63\%, respectively. The two loop-time differences are 3.2\,s and 4.4\,s, while lower-frequency differences are inconsistent. All methods in each paired timing comparison use identical instrumentation.
}

\FloatBarrier
\section{Additional Domain and Learning-rate Controls}
\label{app:othertransitions}
{
\subsection{Direct pretrained-to-Law adaptation}

Our main domain experiment begins from a model already fine-tuned on Math. To examine retention from a different starting point, we also fine-tune Qwen3-1.7B directly from its pretrained checkpoint to Law using full-parameter updates (Table~\ref{tab:p2l}). All methods share the starting checkpoint, Law data, and training budget. Sequential fine-tuning achieves the highest Law accuracy, but reduces GSM8K accuracy from 79.9\% to 28.2\%. Averaging substantially mitigates this decline. Among the averaging methods, SFA achieves the highest mean Law accuracy, while its Math accuracy overlaps Lookahead's within the reported standard deviations. In contrast, endpoint interpolation preserves the most Math at the cost of lower Law accuracy, illustrating the different balances these methods offer.

Separately, we test whether explicitly requesting a worked solution can recover mathematical performance. This diagnostic uses a different prompt from the standard evaluation reported in Table~\ref{tab:p2l}. Under this instruction, sequential fine-tuning achieves only $12.9\pm0.7$\% on GSM8K, generating approximately 17--19 tokens per answer. At most 1.4\% of responses reach the generation limit across the measured conditions, and inspecting generated answers reveals no extraction error explaining the decline. Requesting a worked solution therefore does not recover performance, although this diagnostic does not establish the cause of the degradation.

\begin{table}[t]
\centering\small
\caption{Direct pretrained-to-Law full-parameter fine-tuning of Qwen3-1.7B. Test accuracy (\%, mean $\pm$ sample SD over three seeds) uses the standard evaluation prompt, not the worked-solution diagnostic. The pretrained model scores 79.9\% on GSM8K and 51.6\% on Law.}
\label{tab:p2l}
\setlength{\tabcolsep}{4pt}
\begin{tabular}{@{}lrr@{}}
\toprule
Method & GSM8K & Law mean\\
\midrule
Sequential fine-tuning & $28.2\pm4.0$ & $91.1\pm0.5$\\
Endpoint interpolation, $\beta=.5$ & $75.8\pm1.1$ & $87.5\pm0.7$\\
Lookahead & $72.9\pm2.7$ & $89.2\pm0.6$\\
SFA & $71.7\pm3.4$ & $89.9\pm0.6$\\
\bottomrule
\end{tabular}
\end{table}

\subsection{Full-parameter replay comparison}\label{app:fullparam-stream}
Table~\ref{tab:fullparam-replay-tasks} complements the final averages in Table~\ref{tab:fullparam-replay-cost}. Qwen3-1.7B learns DBpedia, Amazon, Yahoo Answers, and AG News in this order, with approximately 300K input tokens per stage and seeds 17, 42, and 123. All three methods use learning rate $10^{-5}$ and a fresh optimizer at each stage. SFA uses $p=.25$ and $\beta_{\mathrm{end}}=.5$, refreshing its reference between tasks. Replay uses a global 500-example reservoir, displacing approximately 10\% of current-task input tokens after the first task; SFA-R applies SFA to the identical replay schedules. Unlike the adapter comparison, the total token budget is held fixed rather than current-task exposure.

For timing, one H200 runs the complete seed-17 streams in the order SFA, replay, SFA-R, then in reverse. The two training-loop measurements are 113.1/121.8\,s for SFA, 122.8/143.1\,s for replay, and 121.6/139.8\,s for SFA-R. Model loading and setup are excluded. SFA is faster in both repetitions, while SFA-R and replay have similar measured costs; the small difference between their means is within repetition variability. Replay schedules process 4.9\% more padded tokens despite comparable input-token budgets. This accounts for an additional source of work, rather than extra replay tokens beyond the total budget.

\begin{table*}[t]
\centering\small
\caption{Per-task test accuracy after the complete Qwen3-1.7B full-parameter stream, DBpedia $\to$ Amazon $\to$ Yahoo Answers $\to$ AG News (\%, mean $\pm$ sample SD over seeds 17, 42, and 123). Every entry measures final accuracy, rather than accuracy immediately after learning that task. Replay and SFA-R use identical schedules with a 500-example reservoir and approximately 10\% replay input tokens after the first task, under the same total token budget as SFA. Bold indicates the highest mean in each column.}
\label{tab:fullparam-replay-tasks}
\begin{tabular}{@{}lrrrr@{}}\toprule
Method & DBpedia & Amazon & Yahoo Answers & AG News\\\midrule
SFA & $96.76\pm0.11$ & $\mathbf{60.40}\pm0.21$ & $66.63\pm3.19$ & $89.85\pm0.22$\\
Replay & $97.00\pm1.26$ & $57.52\pm1.60$ & $70.15\pm1.08$ & $\mathbf{90.52}\pm0.39$\\
SFA-R & $\mathbf{98.07}\pm0.23$ & $59.13\pm1.32$ & $\mathbf{72.37}\pm0.81$ & $89.72\pm0.23$\\
\bottomrule\end{tabular}
\end{table*}

\subsection{Sensitivity to the learning rate.}
\label{app:other-domain-controls}
The extent of forgetting also depends on the optimization settings. We examine this in the Qwen3-1.7B full-parameter four-task classification stream, using approximately 300K input tokens per stage. At a shared learning rate of $10^{-5}$, SFA improves final average accuracy over sequential fine-tuning by $10.3\pm0.8$ percentage points. To assess how much of this difference remains after tuning, we select each method's learning rate from $\{10^{-5},5\cdot10^{-6},2.5\cdot10^{-6}\}$ using seed-17 development data. This selects $5\cdot10^{-6}$ for sequential fine-tuning and $10^{-5}$ for SFA and Lookahead.

Under the selected learning rates, final test accuracy is $77.4\pm1.1$\% for sequential fine-tuning, $78.4\pm0.8$\% for SFA, and $78.2\pm0.6$\% for Lookahead. SFA retains the highest mean among these three configurations and the lowest forgetting: 2.8 points, compared with 4.7 for sequential fine-tuning and 3.7 for Lookahead. Its paired accuracy gain over sequential fine-tuning is $1.1\pm1.8$ points, positive in two of three seeds. Thus, selecting a suitable learning rate substantially narrows the difference, while SFA retains a modest mean advantage in final accuracy and lower mean forgetting.

}

\section{Further Related Work}
\label{app:recent}
{
The growing literature on model merging includes several settings that share the goal of preserving earlier knowledge but differ in how new models or data arrive. \citet{tang2025opcm} use orthogonal projection and adaptive scaling to merge a stream of completed models without retraining. \citet{sokar2025aligned} encourage alignment while learning incoming vision-language tasks, then merge their parameters. \citet{phan2025holistic} combine tangent-space training, optimizer-state information, and post-merge correction. These works reinforce the importance of distinguishing intervention during training from consolidation of independently learned updates.

For language models, K-Merge explicitly budgets adapter storage \citep{shenaj2026kmerge}, while SLAO combines a single LoRA adapter with orthogonal initialization and asymmetric factor merging \citep{qiao2026merge}. SAFE-Merge, a recent preprint, studies data-free continual merging with sparse masking and low-rank recovery to protect general knowledge \citep{qiu2026safe}. These methods address related retention problems under different information and storage constraints. Our purpose is to test whether the additional mechanisms are needed to obtain the observed accuracy-cost trade-off in our supervised streams.}

Related NLP work also clarifies the scope of the comparison. STAR truncates task-vector spectra and rescales them before consolidating completed models \citep{lee2025star}; it does not intervene in a single ongoing trajectory. MERA merges a multimodal backbone and realigns lightweight modality connectors with replay \citep{zhang2025mera}. \citet{wu2026interpolation} study post-hoc interpolation between instruction and reasoning models to control reasoning cost. These objectives differ from retaining earlier capabilities while optimizing new supervised data. Longer language streams are studied by \citet{guan2025multistage}, who address evolving knowledge with preference-based learning and self-distillation; their seven-stage results do not predict the relative performance of SFA and Lookahead.

\paragraph{Lookahead and continual learning.}
\label{app:lookahead-context}
Lookahead was introduced as a general optimization method for improved stability, convergence, and generalization \citep{zhang2019lookahead}. Its experiments cover image classification, language modeling, and machine translation, rather than sequential-task forgetting. Subsequent work nevertheless connects fast--slow learning with continual adaptation. La-MAML uses a meta-objective and episodic memory \citep{gupta2020lamaml}; the lifelong-learning experiments of Lookbehind-SAM include a replay-based Lookahead-C-MAML comparator alongside replay-free Lookbehind-SAM experiments \citep{mordido2024lookbehind}. These are distinct from applying the original Lookahead interpolation rule to AdamW without replay. More recently, CoVON develops replay-free fast--slow posterior updates for continual learning, including LLM fine-tuning, and relates their fixed-variance isotropic special case to Lookahead \citep{paul2026covon}. There are also direct applications to language-model retention: \citet{li2024continually} compare Lookahead as a fusion rule in sequential language-model editing.

\citet{abbes2026replay} study continual pretraining on successive language corpora and explicitly evaluate forgetting, retained validation loss, and downstream performance. Their Reptile-only variant uses no replay: every 500 updates it interpolates toward a block-start checkpoint with step size .1 and refreshes that reference. They relate this moving-reference update to Lookahead; gradient alignment describes its effect rather than an additional gradient-alignment computation. Our experiments address supervised specialization across domains and sequences of labeled tasks, using both full-parameter adaptation and shared LoRA adapters. We compare a reference fixed within each task with a moving reference, and assess retained task accuracy alongside acquisition and additional training work. Thus, the novelty of this empirical analysis is in the controlled supervised setting, reference-policy and schedule comparisons, and training-work accounting---not in claiming that Lookahead-style forgetting mitigation or parameter interpolation was previously unknown. We treat both SFA and Lookahead as meaningful simple alternatives to replay-dependent or adaptively controlled baselines.

Reference selection provides an interpretable way to specify which learned solution should continue to influence adaptation. When preserving mathematical ability is a priority, a checkpoint validated on Math provides a natural reference; a checkpoint trained across several tasks can instead represent broader capabilities. The averaging weight and interval determine how strongly the chosen solution influences adaptation, separating reference choice from correction strength. The task-boundary policy in our main experiments preserves the incoming model throughout each task. The scheduled-reference variant in Algorithm~\ref{alg:sfpa} further separates merge timing from reference refresh timing; refreshing after every merge recovers the Lookahead parameter update under matching coefficients and optimizer handling. Keeping a particular checkpoint across several tasks is another possible policy; the reference-selection comparison in Appendix~\ref{app:reference-choice} illustrates the resulting balance between earlier-task retention and subsequent learning.

The simplicity distinction concerns the additional learning mechanism. CoVON maintains Gaussian posterior information, estimates curvature, and combines posterior means using precision weights. SFA and Lookahead instead interpolate parameter vectors using scalar coefficients and leave the underlying optimizer unchanged. They require no additional posterior parameterization, uncertainty estimation, or curvature-weighted fusion. This distinction concerns the additional learning mechanism; direct resource comparisons with CoVON remain open.

The distinction between full fine-tuning and LoRA is also substantive: \citet{biderman2024less} find differences in both adaptation and forgetting in mathematical and programming domains. Our adapter-size experiments hold rank and target projections fixed, not the fraction of trainable parameters. Consequently, they characterize a practical configuration transfer across sizes rather than a causal test of model scale alone.
\FloatBarrier

\section{Domain and Vision Experiments}
\label{app:earlier-experiments}
  \label{app:earlier}
 
 \textbf{Datasets and Models:} In order to test the efficacy of our method and compare it to other baselines, we fine-tune our models on a variety of datasets. Specifically, we use multiple tasks from 3 distinct language domains: Law, Math and Code, as well as, streams of image classification tasks from Food-101 \citep{bossard14}, and CIFAR-100 \citep{Krizhevsky09learningmultiple}. Details about which datasets we use in our language domains, and how we construct the image tasks are in~\cref{general:tasks}. Finally, fine-tuning model descriptions, and details about hyperparameter selection can be found in~\cref{model_descriptions} and \cref{hyperparameters} respectively.\\
 \textbf{Evaluation Metrics for Data:} In our work, we reference the \textit{forgetting} of various tasks. We define forgetting specific knowledge as a decrease in performance on a given evaluation task between the current model and the original model before fine-tuning. For example, if evaluation performance on Task A drops when a model fine-tunes on Task B, given that the model has already fine-tuned on task A, we consider the model to forget Task A.  To evaluate performance on our fine-tuning data, we use the metrics and holdout sets described in \cref{appendix:eval-metrics}. In summary, we focus on accuracy when evaluating performance on holdout sets, as it is a common metric in both continual learning and model merging~\citep{garg2024ticclipcontinualtrainingclip,wortsman2022robust,scialom2022finetuned}. This helps bridge the gap between the two fields and clearly highlight performance differences.

\subsection{Mitigating Forgetting from Cross Language Domains}
\label{averaging:results}
\begin{figure}[t]
     \begin{center}
    \includegraphics[clip,width=\linewidth]{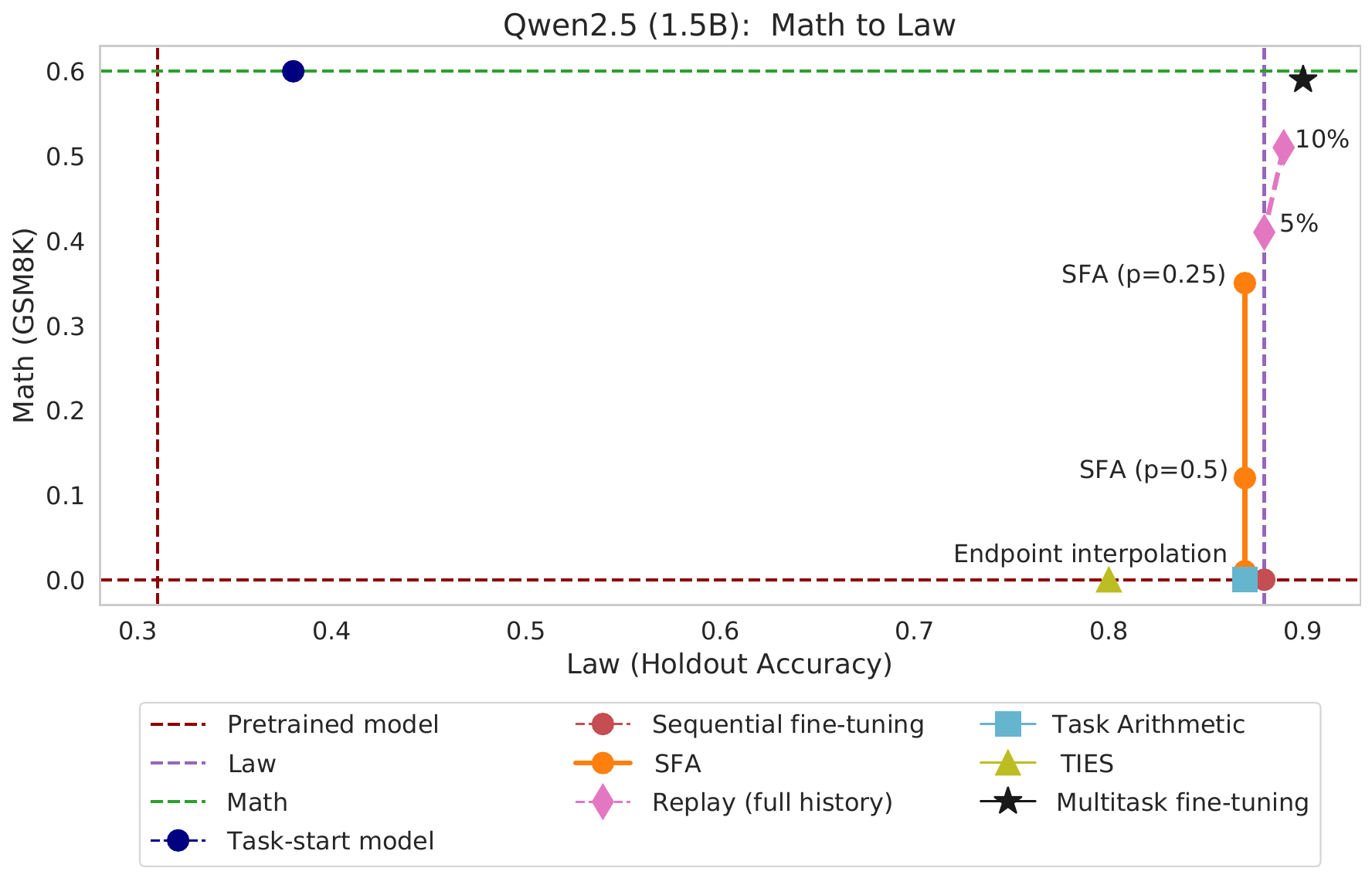}
     \caption{Qwen2.5-1.5B Math$\to$Law comparison. SFA at $p=.25$ retains more Math than the displayed Task Arithmetic and equal-weight endpoint configurations.  (Table~\ref{tab:legacycompact}).}
     \label{fig:qwen1.5b_mathlaw}
     \end{center}
\end{figure}

\begin{figure}[t]
     \begin{center}
    \includegraphics[clip,width=\linewidth]{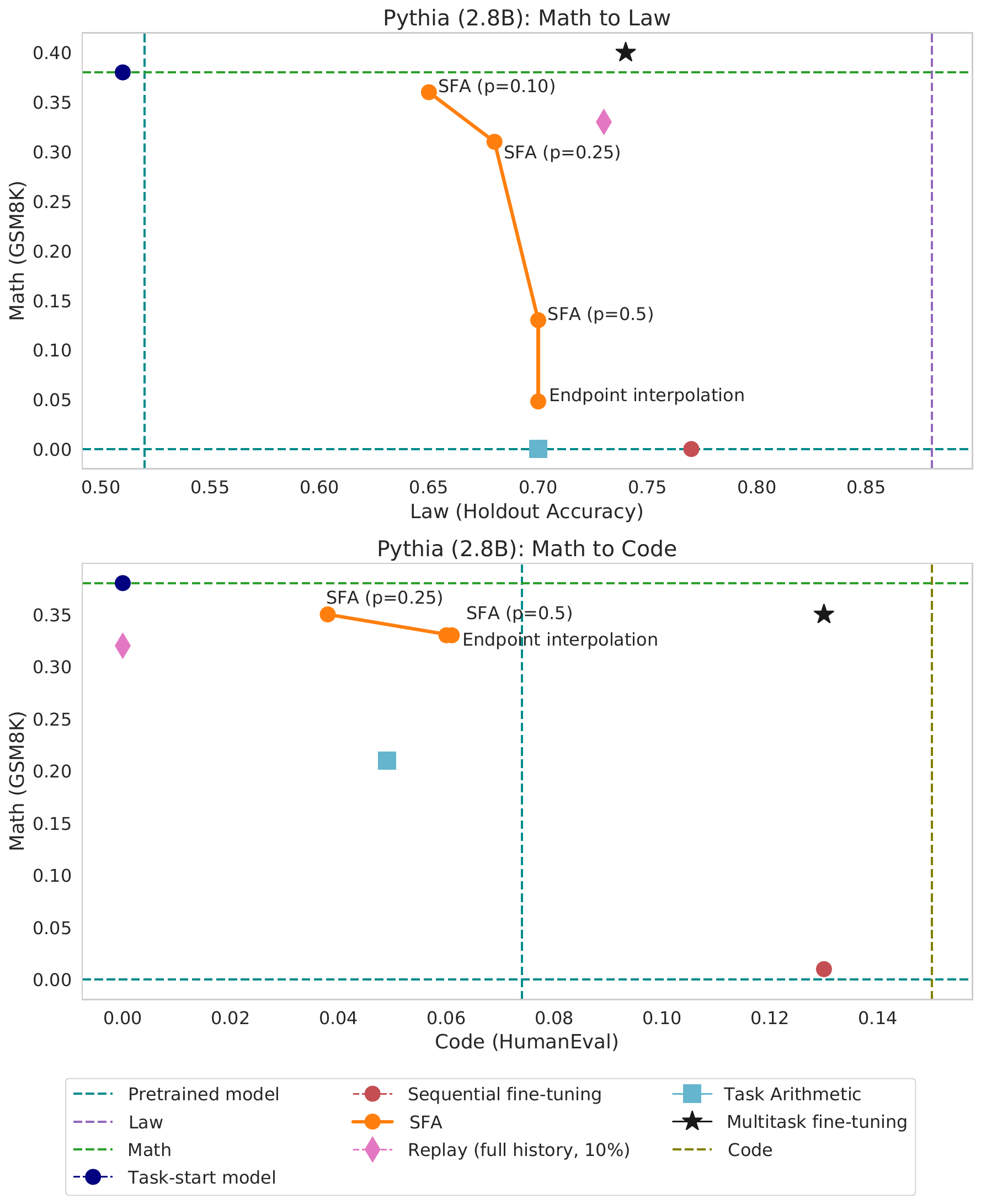}
     \caption{A comparison of Pythia (2.8B)'s performance on multiple domains (Math, Law and Math, Code) using various fine-tuning and model merging techniques similar to \cref{fig:llama7b_mathlaw}. On Math to Law, SFA $p=0.25$ can be seen as having comparable performance to using a data buffer, while outperforming Task Arithmetic. Likewise, in Math to Code, SFA with varying $p$ outperform using a data buffer and Task Arithmetic.}
     \label{fig:pythia2.8b}
     \end{center}
\end{figure}

We evaluate successive instruction fine-tuning tasks with domain shifts, including Math to Code and Math to Law, using the datasets in \cref{general:tasks}.
By restricting ourselves to pairs of dissimilar tasks, we can clearly quantify the trade off between learning the second task and forgetting the first one by visualizing the results on a plane that measures the accuracy of the first task on the y-axis and the accuracy of the second task on the x-axis. First, we confirm that forgetting occurs when fine-tuning on successive language tasks (\cref{appendix:forgetting_results}).
We present our results for sequentially learning Math and Law with Llama 2 (7B) in \cref{fig:llama7b_mathlaw}, Math and Law with Qwen2.5 (1.5B) in \cref{fig:qwen1.5b_mathlaw}, and Math and Law, as well as Math and Code with Pythia (2.8B) in \cref{fig:pythia2.8b}.

We first fine-tune our model Llama 2 (7B) in \cref{fig:llama7b_mathlaw}, Qwen2.5 (1.5B) in \cref{fig:qwen1.5b_mathlaw},  and Pythia (2.8B) in \cref{fig:pythia2.8b}) on MetaMathQA to obtain the initial model (dark blue circle).  
Note the base model performance on the first (second) task is represented by dark green for Llama 2 (7B), dark red for Qwen2.5 (1.5B), and blue for Pythia (2.8B) horizontal (vertical) dashed lines. 
This initial model improves upon the base model on our Math benchmark and is thus higher on the y-axis (performance on first task) while not being significantly different or being worse on the x-axis (performance on the second task which it has not been trained on yet). 
We then fine-tune the initial model on the second task to obtain the sequential fine-tuning model (red circle).
In \cref{fig:llama7b_mathlaw,fig:qwen1.5b_mathlaw} the second task is Law while in \cref{fig:pythia2.8b} the second task is either Law or Code. 
The sequential fine-tuning model performs well on the second task (higher on the x-axis) while forgetting almost everything it has learned about the first task (base model level on the y-axis).
This movement down and to the right of the initial model (dark blue circle) to the sequential fine-tuning model (red circle) on the task 1 - task 2 performance plane in \cref{fig:llama7b_mathlaw,fig:qwen1.5b_mathlaw,fig:pythia2.8b} is emblematic of catastrophic forgetting of an earlier task as the model learns a new task.
For reference, the performance of just fine-tuning the base model on the second task is represented by the vertical purple for Law, or green for Code dashed line. For our upper baseline, we show the results of simultaneously fine-tuning the base model  on a mixture of both tasks to obtain the multitask fine-tuning model (black star). 
This model sits at the upper right of the plane as it does not exhibit forgetting and performs well on both tasks.
However, as stated before, in our continual learning setting where data streams in as a sequence of tasks, this is infeasible. As such, we implement a common continual learning method: rehearsal with a data buffer. Rehearsal is a common technique for mitigating forgetting in continual learning~\citep{prabhu2023computationallybudgetedcontinuallearning,garg2024ticclipcontinualtrainingclip}. It involves maintaining a buffer of past task data and interleaving it with new task data during fine-tuning \citep{doi:10.1080/09540099550039318}, where the size of the buffer is a hyperparameter. In our baseline setup, each batch is constructed using streams that ensure it maintains x\% of samples from the current task with 100-x\% of samples randomly drawn from the buffer of past tasks, ensuring continual exposure to prior knowledge while prioritizing adaptation to new data. 
For simplicity, the stream randomly draws from the full training data of past tasks. The 5\% and 10\% settings below denote replay sampling proportions, not limits on retained data; we refer to this as replay (full history). Fixed-memory experiments are identified separately.
 We demonstrate its effectiveness by further training our initial model (dark blue circle, fine-tuned on Math) on a mixture of $90\%$ (95\%) task 2 data and $10$\% (5\%) of Math data sampled randomly from the full Math dataset.
The resulting continual learning (CL) with data buffer model (pink diamond in \cref{fig:llama7b_mathlaw,fig:qwen1.5b_mathlaw,fig:pythia2.8b}) effectively improves on the initial model on task 2 (higher Law performance, i.e. x-axis) while mitigating forgetting (maintains high Math performance i.e. y-axis). As we increase the proportion of Math data from $5$\% to $10\%$, we see higher performance on Math (\cref{fig:qwen1.5b_mathlaw}). 
Replay (full history) requires access to earlier training data. Fixed-capacity replay can limit storage; replay fractions alone do not determine memory or compute costs (\cref{app:storage_compare}).
 Furthermore, maintaining a subset of past data can also threaten data privacy and security~\citep{li2024continual}. 
Finally, our domain fine-tuning scenario is set up for a variety of models, with baselines that include no intervention, multitask fine-tuning, and rehearsal with a data buffer. 

\subsection{SFA on Cross Domain Data}
\label{SFA-method}

\paragraph{Qwen2.5 domain adaptation.}
For Qwen2.5, SFA at $p=.25$ retains 35\% GSM8K accuracy, compared with 1\% for equal-weight endpoint interpolation, while the three Law accuracies differ by at most two percentage points (Table~\ref{table:qwen-math-law}). We use these results to illustrate the observed trade-offs of the methods.

Next, we test the ability of SFA to retain past domain performance when learning a new domain by implementing it and comparing it to a variety of other merging techniques. Recall that in SFA, we take a model that has already been fine-tuned on Task A, and while fine-tuning on Task B, every $pT$ steps we average the weights with the final model after fine-tuning on Task A and continue fine-tuning on Task B.
We evaluate SFA with varying averaging frequency $p$ during cross-domain sequential fine-tuning.
\cref{fig:llama7b_mathlaw,fig:pythia2.8b} show that as $p$ decreases, signifying more frequent averaging with the initial model, 
we observe stronger retention of past domain knowledge (orange curve). By adjusting the averaging frequency ($p$), we control the balance between past and new knowledge retention given that the model has progressed on the new task. This is evident, because as $p$ decreases, the performance on Math (y-axis) increases, indicating stronger retention of task 1. The corresponding cost in new-task performance varies across models, domains, and averaging intervals.
For Math followed by Law, using $p=0.25$ allows SFA to retain earlier knowledge while continuing to learn the new domain. 
Crucially, our method is able to achieve such performance without requiring a data buffer, but just two model checkpoints: the initial and current one throughout fine-tuning.

Next, we examine whether averaging during training provides a benefit over the other model merging techniques in this setting.

We implement Task Arithmetic ~\citep{ilharco2023editing} (blue square), TIES ~\citep{yadav2023tiesmerging} (green triangle), and WiSE-FT \citep{wortsman2022robust}, 
and observe substantial retention gains in several settings, particularly Math-to-Law.

In the Math-then-Law setting, the equal-weight Task Arithmetic and the reported TIES configurations retain little Math performance, while SFA can retain substantially more. 
Our figure values for Pythia (2.8B) can be found in \cref{table:pythia-math-law} (Math and Law), and~\cref{table:pythia-math-code} (Math and Code). Results for Llama 2 (7B) can be found in~\cref{table:llama-math-law} (Math and Law), and \cref{table:llama-math-code} (Math and Code). Finally, results for Qwen2.5 (1.5B) can be found in \cref{table:qwen-math-law} (Math and Law).

To see how our method scales as the number of domains increases, we also continue fine-tuning and applying SFA on our model for 3 domains (\cref{fig:pythia2.8b-3domains}). In these graphs, we take a high performing SFA model ($p$ of 0.25) on Math and Law, and Math and Code from~\cref{fig:pythia2.8b}, and continue fine-tuning the model with SFA on the final domain (Code and Law respectively). We find that by using SFA (specifically adjusting $p$), we are able to maintain high performance on the previous 2 domains while also learning an additional domain. As such, SFA is a useful forgetting mitigation technique for continual learning given a sequence of domains. In the case of Math-Code to Law, SFA with $p$ of 0.25 yields performance comparable to rehearsal (pink diamond). The figure results of Pythia (2.8B) fine-tuning on Math-Code to Law, and Math-Law to Code can be found in~\cref{table:pythia-math-law-code-3}.

\begin{figure*}[t]
\centering
\includegraphics[width=.95\textwidth]{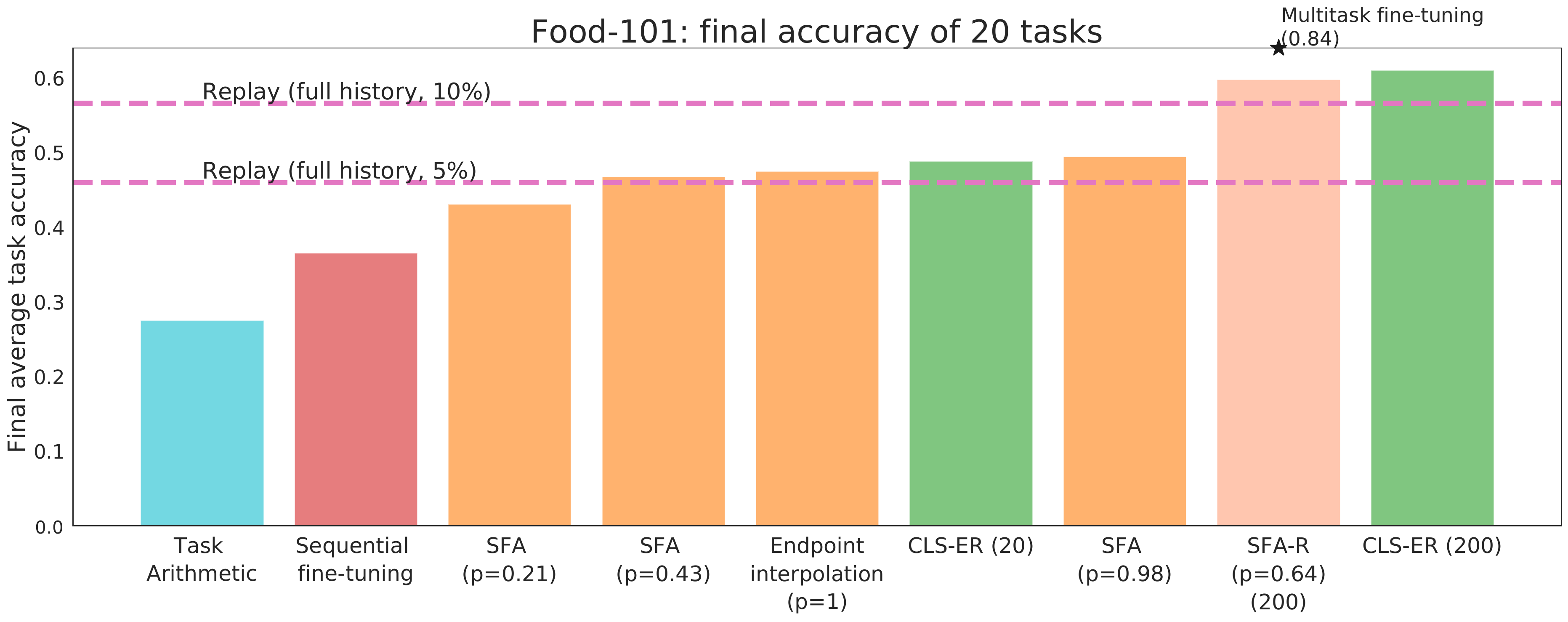}
\par
\includegraphics[width=.95\textwidth]{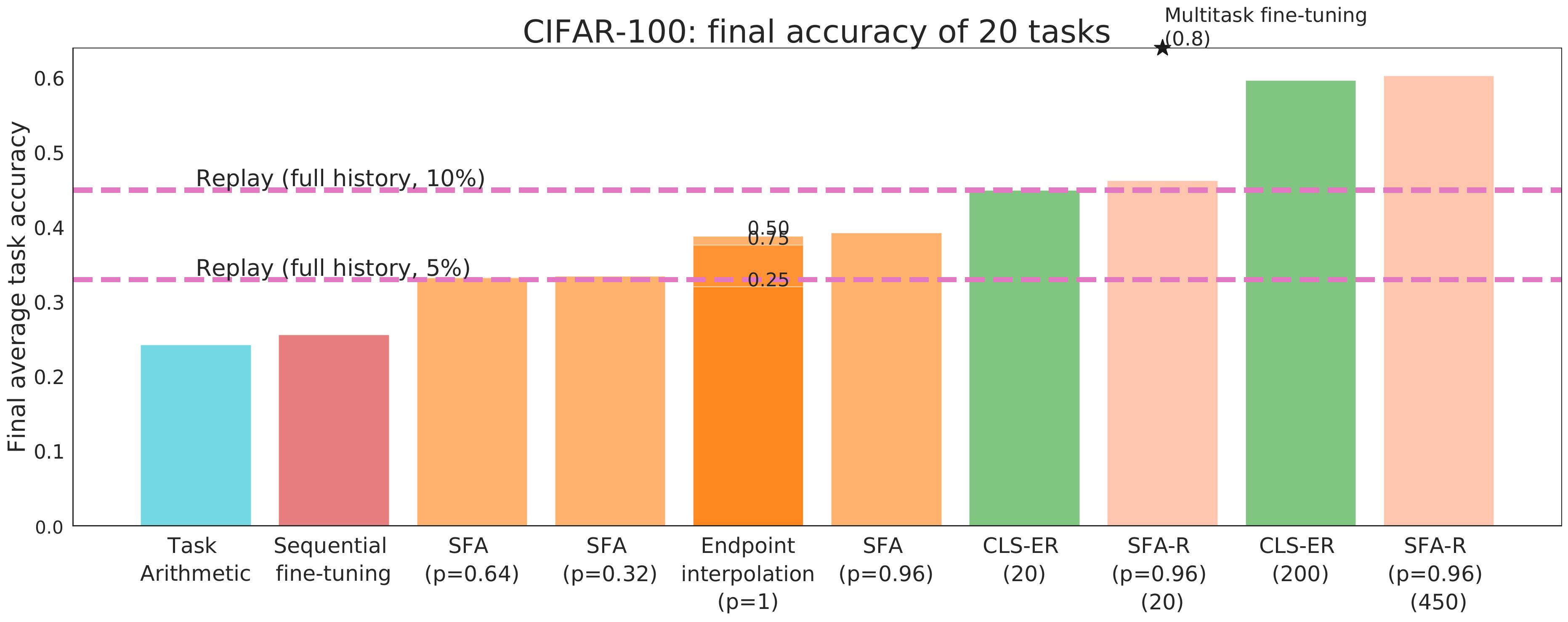}
\caption{20-task vision experiments with ViT-base: Food-101 (top) and CIFAR-100 (bottom). Replay-free SFA improves on the displayed sequential, Task Arithmetic, and endpoint configurations. Dashed lines use replay (full history) at the indicated sampling proportions. SFA-R is a replay-augmented variant of SFA. Multitask reference accuracies are annotated above the plotted range.}
\label{fig:cv_final}
\end{figure*}

\subsection{Mitigating Forgetting Across Sequences of Image Tasks}
\label{image_tasks_SFA}
Building on SFA’s success with dissimilar task pairs in the language domain, we next evaluate its performance on image classification tasks to provide a comprehensive empirical analysis across settings. Specifically, we demonstrate SFA’s effectiveness in a continual learning scenario, where a ViT model is fine-tuned on a long sequence of image classification tasks drawn from both Food-101 and CIFAR-100 (\cref{fig:cv_final}). We provide an upper baseline by simultaneously fine-tuning the initial model on all  tasks to obtain a multitask fine-tuning model which performs well on average (black star). However, as previously stated, this is infeasible given a continual learning setting where task data appears sequentially.
Next, we fine-tune without intervention (red bar) which results in low average accuracy due to the catastrophic forgetting of earlier tasks. To mitigate this forgetting, we implement 2 variations of Rehearsal with a data buffer: one that combines 5\% past task data and 95\% current task data, and another with 10\% past task data and 90\% current task data (pink dashed horizontal lines). As in the previous examples, we sample from the full training data of past tasks to supply the specified 5\% or 10\% replay proportion. 
We also compare with CLS-ER~\citep{arani2022learningfastlearningslow}, using fixed replay capacities of 20--200 examples. On Food-101, the plotted results show SFA alone approaching CLS-ER with a small buffer, and SFA with fixed memory approaching CLS-ER at some larger capacities. On CIFAR-100, matching the stronger CLS-ER settings requires adding more replay examples to SFA.\\   
Similarly to \cref{averaging:results}, we next apply commonly used model merging methods to compare our original SFA to. Task Arithmetic \citep{ilharco2023editing} (blue bar), and WiSE-FT \citep{wortsman2022robust} (orange bar). As we explain in \cref{averaging:continual}, generalized WiSE-FT can be equivalent to SFA with $p=1$, because it only involves merging the final trained and initial model. Finally, we compare these baselines to our method, SFA, which merges a partially-trained and an initial model, using varying averaging frequency $p$. Selected averaging configurations outperform sequential fine-tuning and the displayed Task Arithmetic baseline. Finally, SFA ($p=0.98$ for Food-101, and $p=0.96$ for CIFAR-100) where averaging occurs near the end and after training, achieves higher performance than WiSE-FT or SFA ($p=1$) where averaging occurs only once at end.
This suggests a benefit from continuing to train after averaging. However, the performance gap between WiSE-FT and SFA is less severe than when tested against dissimilar language tasks in \cref{SFA-method}.   
One possible explanation for the smaller gains in vision is that these tasks share more knowledge or require a different amount of learning before averaging becomes useful.

\begin{figure*}[t]
\begin{center}

  \includegraphics[clip,width=\linewidth]{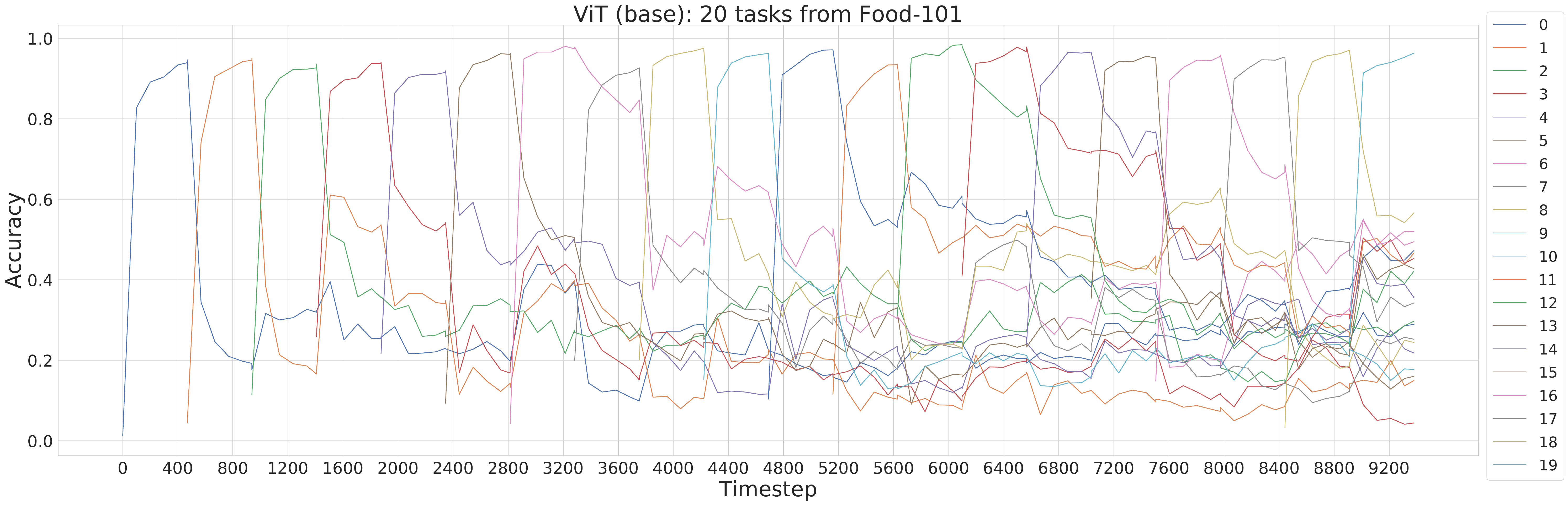}%

  \includegraphics[clip,width=\linewidth]{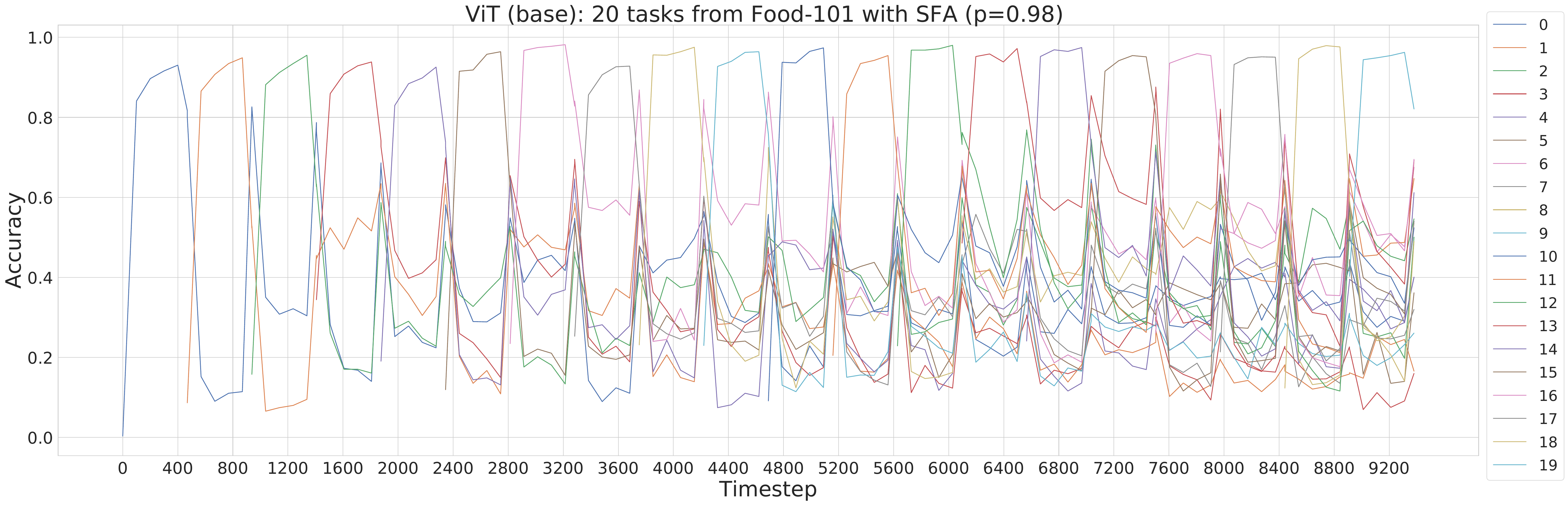}%

\caption{A comparison of sequentially fine-tuning ViT (base) on 20 tasks (Food-101) with (\textbf{bottom}) and without SFA (\textbf{top}). Each new task is introduced with a different colored curve across gradient timesteps (x-axis) resulting in changes to both current and past task accuracies (y-axis). The use of SFA can be seen to improve cumulative past task performance at averaging steps. }
\label{fig:food101_across_t}
\end{center}
\end{figure*}

In order to better understand why the optimal $p$ is so high in this scenario, we investigate the effects of averaging throughout this training cycle. We track the accuracy (y-axis) of both past and current tasks as a model fine-tunes on a sequence of tasks from Food-101 (\cref{fig:food101_across_t}). We compare a model fine-tuning without intervention (top graph) to fine-tuning using SFA (p=0.98) (bottom graph). As is shown, without intervention, past task performance continues to decrease as new tasks are introduced. Meanwhile, SFA boosts past task performance when averaging occurs (performance spikes in graph). Furthermore, as can be seen, each task requires a substantial portion of training steps to raise its performance on the given task. This may explain why SFA ($p=0.98$) outperforms SFA ($p<0.98$), as averaging before substantial learning on the given task has taken place may hinder current task performance. We contrast this to \cref{SFA-method}, where averaging earlier on in training can be beneficial for more quickly-learned tasks. The language and vision sweeps both exhibit retention gains, but favor different schedules. The controlled language experiments in the main text test this behavior with explicit selection and multi-seed comparisons.

\subsection{Averaging Weights}
\begin{figure}[t]
     \begin{center}
    \includegraphics[clip,width=\linewidth]{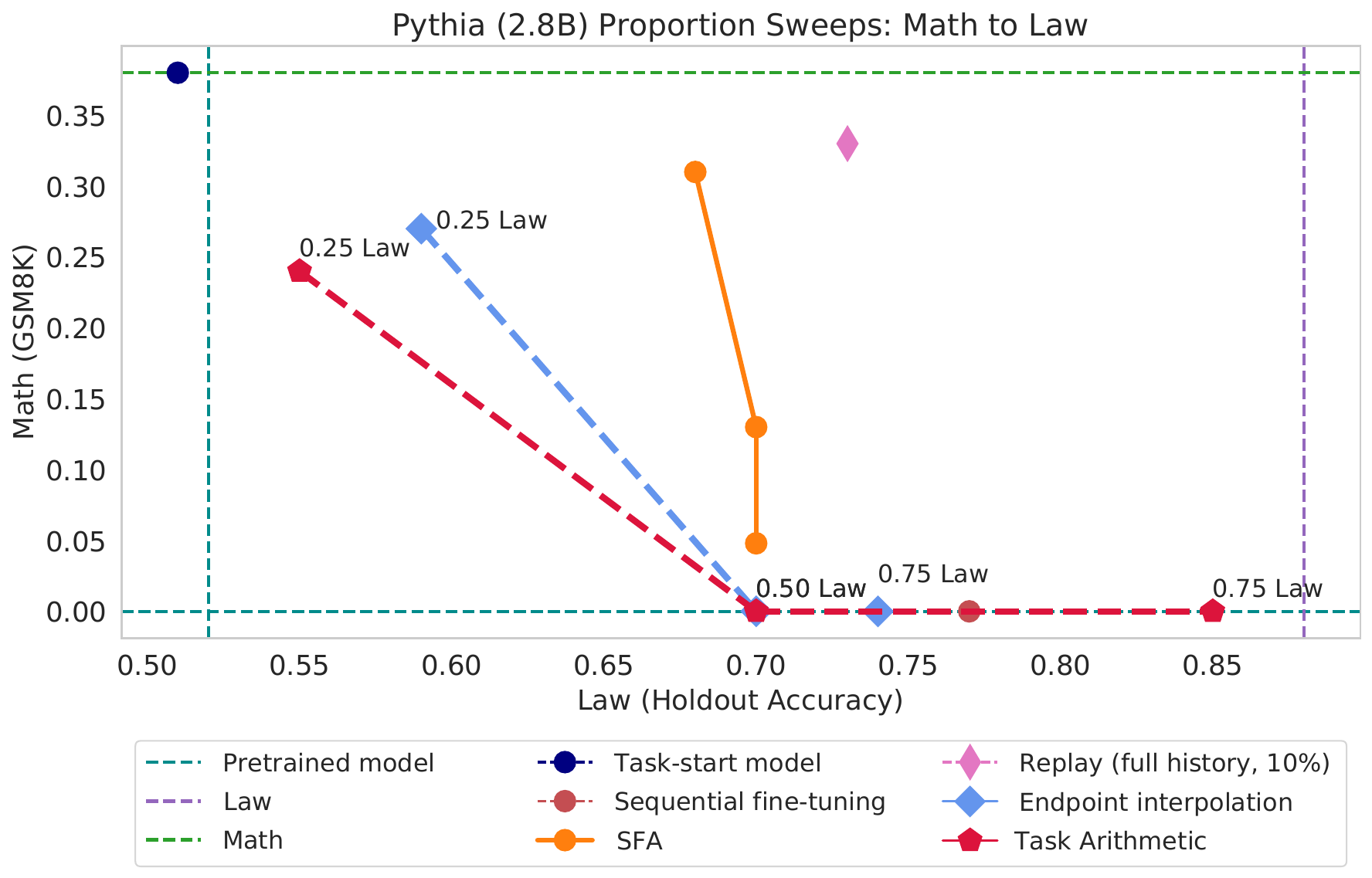}
     \caption{A comparison of varying the Task Arithmetic model weights, and $\beta$ on SFA ($p$=1), with SFA (varying $p, \beta=0.5$) for Pythia (2.8B).  We reproduce the results varying $p$ in SFA (orange curve) from \cref{fig:pythia2.8b} and add 2 sweeps showing change in performance on Pythia (2.8B) when the weights for the current and past checkpoints are varied for SFA ($p=1$) (dashed blue) and the domain-specific models are merged in Task Arithmetic (dashed red). }
     \label{fig:pythia2.8b-taskarithsfasweep}
     \end{center}
\end{figure}

In order to better understand whether SFA's performance advantages are based on its continual averaging, we explore the impact of modifying weights solely at the final stage. Rather than using default averaging which weights each model equally, we vary the weight hyperparameter of each model and compare this to using SFA.

Recall that SFA combines parameters from the initial and current model during fine-tuning. We posit that the initial model represents expertise in past tasks/domains, while the current model embodies new task/domain knowledge. 
Our default parameter weighting (0.50 for each) provides a balance. 
We explore if, instead of varying $p$, the frequency of averaging in SFA, we can get similar flexibility by first fine-tuning the model on a new task ($p=1$, or WiSE-FT) and then averaging the final model with the previous task model using different relative weights (vary $\beta$).
In \cref{fig:pythia2.8b-taskarithsfasweep,fig:llama7b_mathlawtaskarithsfasweep}, we show that SFA with $p<1$ and $\beta=0.5$ (orange curve) performs the same if not better than a sweep of weighting parameter $\beta$ for SFA ($p=1$, or WiSE-FT) (blue curve).  Likewise, for CIFAR-100 in \cref{fig:cv_final}, we show that varying $\beta$ for SFA ($p=1$, or WiSE-FT) is not as effective as SFA ($p=0.96$), implying that averaging during fine-tuning, with additional fine-tuning afterwards offers additional performance benefits.
These sweeps motivate examining what happens when the averaged model continues learning. The controlled experiments extend this comparison by accounting separately for the averaging schedule and cumulative strength.
 
We extend this analysis to Task Arithmetic, another model merging technique. 
In~\cref{fig:pythia2.8b-taskarithsfasweep,fig:llama7b_mathlawtaskarithsfasweep} we report the results sweeping over the weight values for averaging (red curve), and observe that Task Arithmetic, like SFA ($p=1$, or WiSE-FT) with varying $\beta$, fails to achieve the cross-domain performance improvements that SFA demonstrates. Specifically, it also shows even worse combined performance on task 1 (Math, y-axis) and task 2 (Law, or Code, x-axis). Furthermore, in the Math-Law setting, for weights on Law $\geq0.50$, it also fails to retain Math. 

\FloatBarrier
\section{Further Analysis of Quadratic Anchoring}
\label{sec:continual_merging}
\subsection{Intuition for Model Merging}
In this section, we connect model merging to penalty-based regression methods to offer intuition for why merging is so effective. Specifically, there exist many penalty-based methods in continual learning where the forgetting of past tasks is mitigated by constraining training weights. The penalty is often used to prevent weights from straying from model weights that perform well on past tasks. Examples include L1 and L2 penalties and EWC \citep{Kirkpatrick_2017}. Typically, these methods add a penalty to an existing loss objective for every gradient step. 
A direct implementation accesses the reference parameters at every optimizer step. Periodic SFA accesses them less often.  For plain SGD, we can relate per-step SFA exactly to anchored L2 regularization under a learning-rate reparameterization.
Consider, starting with $\theta_{0}$, the model trained on the previous task and $\theta_t$, the model currently being trained on the new task. Calculating the loss with an L2 penalty takes the form: 
\begin{equation}
    L(\theta_{t}) = L_\text{task}(\theta_{t}) + \frac{\lambda}{2} || \theta_{t} - \theta_{0} ||^{2} .
\end{equation} 
Updating the model once using the gradient of this loss is equivalent to:
\begin{equation}
    \theta_{t+1} = \theta_{t} - \eta (\nabla_{\theta_t} L_\text{task} + \lambda (\theta_{t} - \theta_{0})).
\end{equation}
This can be rewritten as: 
\begin{equation}
    \theta_{t+1} =  (1-\eta\lambda)\theta_{t} + (\eta\lambda)\theta_{0} - \eta\nabla_{\theta_t} L_\text{task}.
\end{equation}
Now we can compare this to a case of SFA where averaging occurs after each gradient step. Following the setup in \cref{alg:sfa} given some $T$, for each gradient step, current model parameters are first updated using only task loss, before being averaged with the initial model:  
\begin{align}
   \theta_{t+1}^{*} &= \theta_{t} - \alpha \nabla_{\theta_t} L_\text{task}. \\
    \theta_{t+1} &= (1-\beta)\theta_{t+1}^* + \beta (\theta_{0}).
\end{align}
We can combine these 2 steps to get the following form:\begin{equation}
     \theta_{t+1} = (1-\beta) (\theta_{t} - \alpha \nabla_{\theta_t} L_\text{task}) + \beta (\theta_{0}).
\end{equation}
This is equivalent to 
\begin{equation}
   \theta_{t+1} = (1-\beta)\theta_{t} + (\beta)\theta_{0} - \alpha \nabla_{\theta_{t}} L_\text{task} (1-\beta).
\end{equation}

The two updates agree for plain SGD when $\beta=\eta\lambda$ and $\alpha=\eta/(1-\eta\lambda)$, for learning rate $\eta>0$ and penalty strength $\lambda\geq0$ with $\eta\lambda<1$. Thus, averaging after every gradient step can reproduce anchored L2 regularization under an appropriate learning-rate reparameterization. 

A complementary interpretation applies to each merge, regardless of how the model was trained between merges. SFA balances remaining close to the current parameters against returning toward the earlier checkpoint. Specifically, for parameters $z$ immediately before merging, the averaged model exactly minimizes the quadratic objective
\begin{equation}
\begin{aligned}
 &(1-\beta)z+\beta\theta_0\\
 &=\operatorname*{arg\,min}_{\theta}\Bigl\{
 \tfrac12\|\theta-z\|_2^2
 +\tfrac{\beta}{2(1-\beta)}
 \|\theta-\theta_0\|_2^2
 \Bigr\},\\
 &\hfill 0\leq\beta<1.
\end{aligned}
\end{equation}

Differentiating this objective gives $(\theta-z)+\frac{\beta}{1-\beta}(\theta-\theta_0)=0$, whose solution is the stated average. Its Hessian is $I/(1-\beta)$, which is positive definite for $0\leq\beta<1$, so the minimizer is unique. At $\beta=1$, the merge returns $\theta_0$ directly; the displayed finite-penalty objective applies only below this endpoint.

Increasing $\beta$ places greater weight on the reference, while smaller values preserve more of the current model's update. This interpretation explains how averaging constrains movement away from an earlier solution, without guaranteeing that parameter proximity preserves its predictions. The distance--accuracy relationship examined in \cref{appendix:l2_distance_acc} provides empirical context for this intuition. The Gaussian-prior discussion in \cref{appendix:bayesian_l2} motivates the quadratic penalty, while \cref{appendix:ewc_model_merging_fisher} gives a related split-update interpretation for EWC.

\begin{figure}[t]
 \begin{center}
    \includegraphics[clip,width=\linewidth]{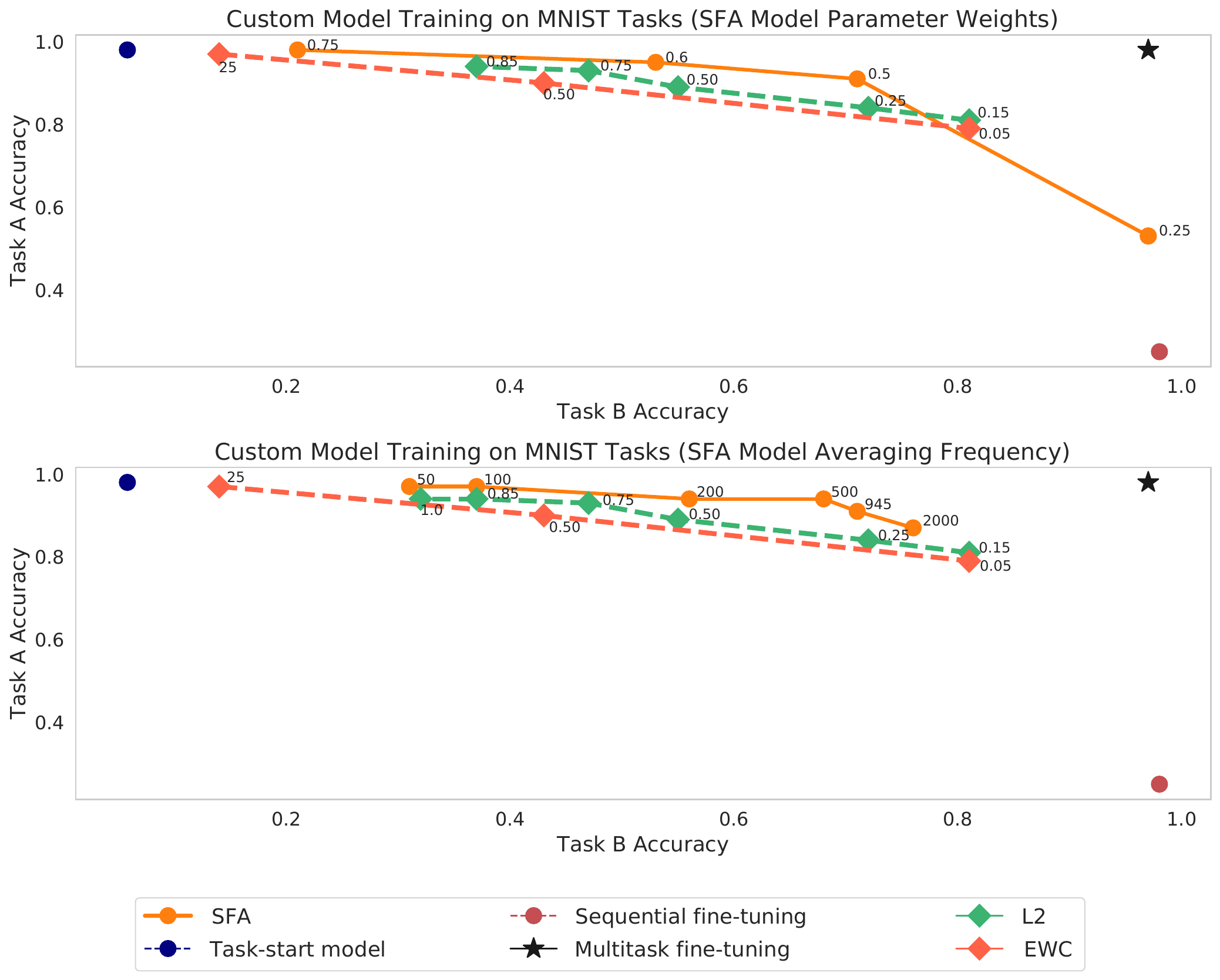}
     \caption{SFA compared against other continual learning methods, where the two tasks (Task A and B) were created by splitting MNIST by label. The accuracy after single-task training, sequential training, and multitask training is also shown. The lines for EWC and L2 are created by varying the coefficient corresponding to each method (and are the same for the top and bottom plots).
     \textbf{(Top)} visualizes SFA performance under varying $\beta$ coefficient, which determines how much weight is being placed on the initial model. 
     \textbf{(Bottom)} visualizes SFA with varying averaging frequency in steps before averaging.}
     \label{fig:custom_continual_compare_avg}
     \end{center}
\end{figure}
\subsection{Performance Comparison of SFA and L2-regression}
We illustrate this connection using a small experiment in which SFA, L2 regularization, and EWC balance performance on two sequential tasks. In~\cref{fig:custom_continual_compare_avg}, we train a small, custom neural network on 2 sequential MNIST tasks (Task A and Task B) separated by label introduced in ~\citet{ewc}. Task A involves labelling the five even digits, whereas Task B labels the five odd digits. The blue dot refers to the model after training on Task A, whereas the red dot is additionally trained on Task B without intervention. As such, performance rapidly drops on Task A as the model optimizes for Task B. The solid orange curve refers to SFA where, in \cref{fig:custom_continual_compare_avg} (top) we vary the averaging weight $\beta$ from \cref{alg:sfa} and in \cref{fig:custom_continual_compare_avg} (bottom) we vary the frequency of averaging in number of batches. As such, placing a higher $\beta$ or lower number of batches before averaging results in a model that performs better on Task A, and vice versa. The green dotted line shows L2-regression where $\lambda$ (weight on L2 penalty) varies, with a higher $\lambda$ performing better on Task A (and vice versa). Finally, an orange dotted line shows EWC with varying $\lambda$ (weight on EWC penalty) with a higher weight performing better on Task A (and vice versa). L2-regression outperforms EWC with a better trade off between performance on Task A and B. Selected SFA points improve the observed trade-off over the displayed L2 and EWC sweeps.

\FloatBarrier
\section{Model-family and Vision Protocols}
\label{app:experimental-details}
\subsection{Models}
\label{model_descriptions}
We fine-tune a combination of encoder-decoder and decoder only models. Specifically, we measure forgetting on T0\_3B (3B) and T0pp (11B) \citep{sanh2021multitask}, two models already pretrained and fine-tuned on many tasks, when sequentially fine-tuning on instruction tasks (\cref{appendix:forgetting_results}). We also fine-tune Pythia (2.8B)~\citep{biderman2023pythia}, Qwen2.5 (1.5B) \citep{qwen2.5} and Llama 2 (7B)~\citep{touvron2023llama} on tasks from different domains (Math, Law, Code) to measure performance on sequential learning, in addition to a variety of merging techniques (\cref{SFA-method}). Finally, we fine-tune vit-base-patch16-224-in21k~\citep{vit-in21k} on image datasets.\\
Mainly, we use Composer~\citep{mosaicml2022composer} for fine-tuning and evaluation. For additional evaluation metrics, we also use Language Model Evaluation Harness~\citep{eval-harness}. Finally, we create some model merging baselines using mergekit~\citep{goddard2024arcee}.
\subsection{Hyperparameters}
\label{hyperparameters}

\begin{figure}[t]
     \begin{center}
    \includegraphics[clip,width=\linewidth]{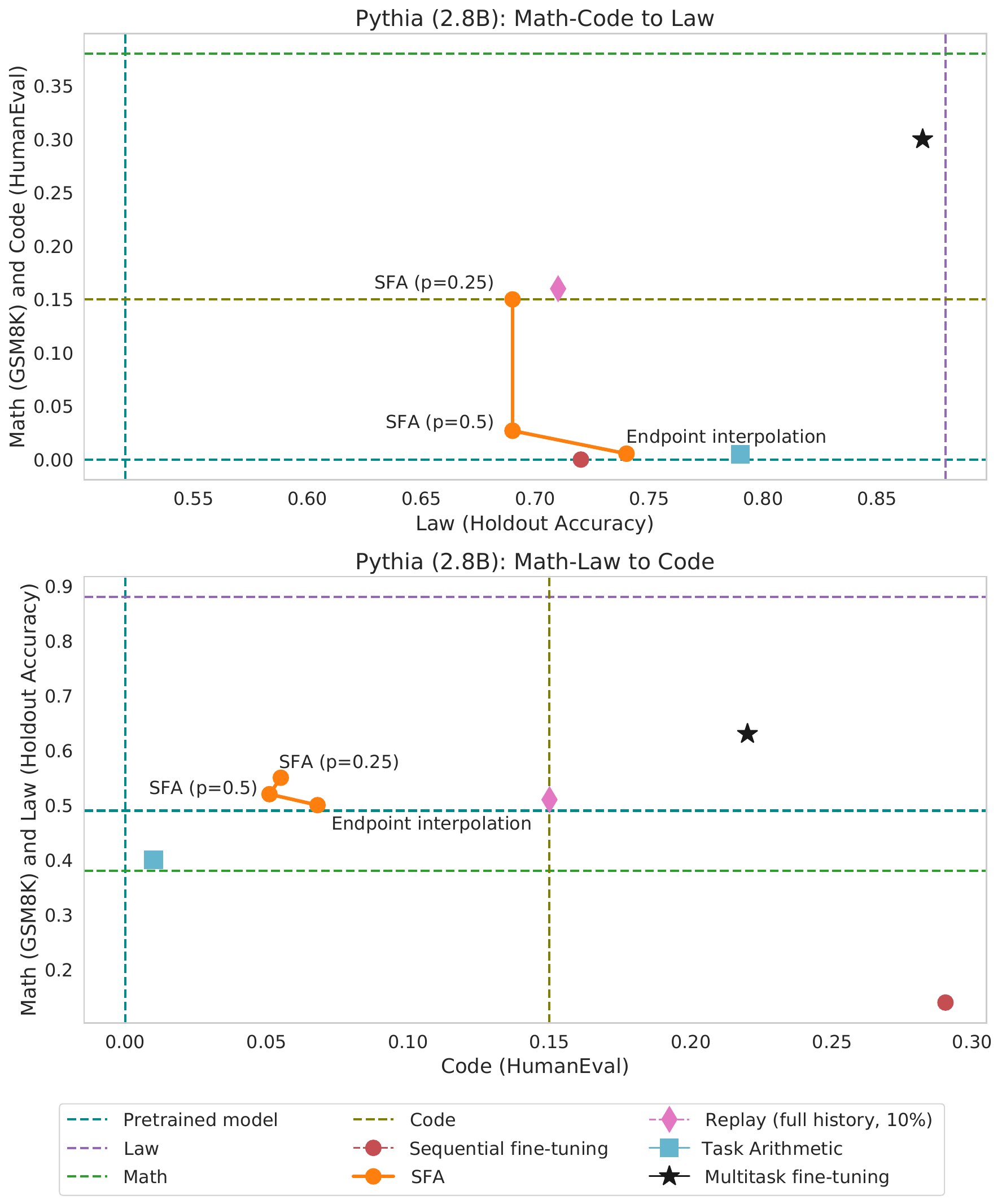}
     \caption{A comparison of Pythia (2.8B)'s performance when training on more than 2 domains (e.g. Math-Law and Code, Math-Code and Law) using various fine-tuning and model merging techniques similar to \cref{fig:pythia2.8b}. On Math-Code to Law, SFA $p=0.25$ can be seen as having comparable performance to using a data buffer, while outperforming Task Arithmetic. While, SFA with varying $p$ on Math-Law to Code outperforms Task Arithmetic, but performs worse than using a data buffer.}
     \label{fig:pythia2.8b-3domains}
     \end{center}
\end{figure}

\begin{figure}[t]
     \begin{center}
    \includegraphics[clip,width=\linewidth]{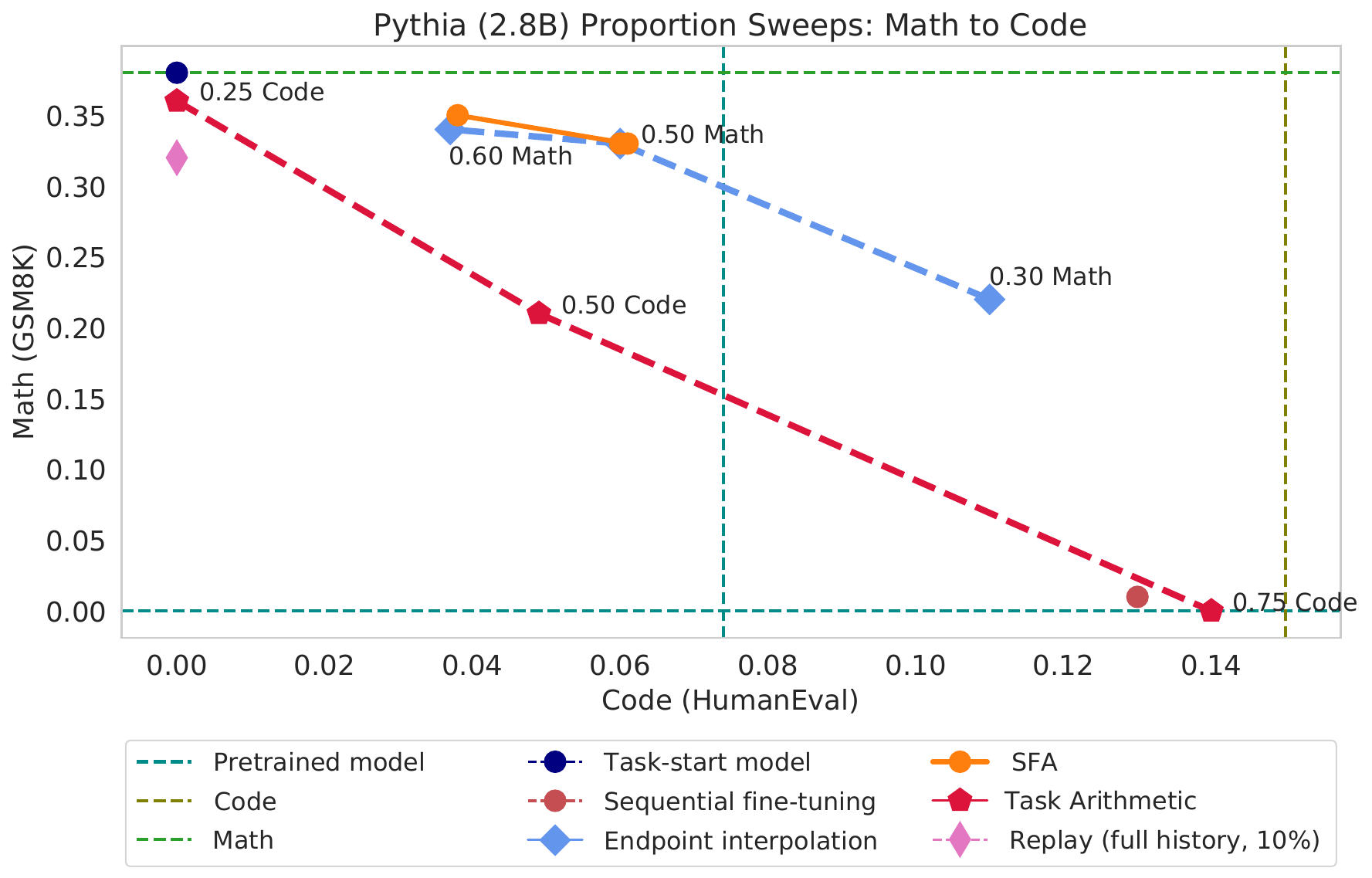}
     \caption{A comparison of varying the Task Arithmetic model weights, and $\beta$ on SFA ($p$=1), with SFA (varying $p, \beta=0.5$) for Pythia (2.8B) on math then code, similar to~\cref{fig:pythia2.8b-taskarithsfasweep}.  (\cref{table:pythia-math-code}).}
     \label{fig:pythia2.8b-taskarithsfasweep_mathcode}
     \end{center}
\end{figure}

Among our experiments, we typically first run a standard sweep on various hyperparameters and select the best performing ones, or record the full sweep to analyze our results with. For example, we run sweeps to determine averaging frequency \textit{p} (\cref{fig:llama7b_mathlaw,fig:qwen1.5b_mathlaw,fig:cv_final,fig:pythia2.8b,fig:pythia2.8b-3domains,fig:custom_continual_compare_avg}), and the model merging weight hyperparameter of baselines including SFA ($p=1$), Task Arithmetic, EWC, and L2 (\cref{fig:pythia2.8b-taskarithsfasweep,fig:llama7b_mathlawtaskarithsfasweep,fig:pythia2.8b-taskarithsfasweep_mathcode,fig:custom_continual_compare_avg}). Finally, when actually fine-tuning, we also initially run a task-specific learning rate sweep (typically from 1e-4 to 1e-6) to ensure robust results. Among our language domains of Math, Code and Law, we find that a learning rate of 1.0e-5 performs well across models and sequential task pairs (an exception being Qwen2.5 (1.5B) fine-tuning on Code for which 1.0e-6 is better). For our image classification tasks, we use a learning rate of 5e-5 for both CIFAR-100 and Food-101. 

\subsection{Instruction Datasets}
\label{instruction-datasets}
We use language generation tasks described in~\citep{scialom2022finetuned} to measure forgetting. These tasks are based on pre-existing datasets that we also reference here:
Text Simplification (Simpl) (Wiki-Auto \citep{jiang-etal-2020-neural}), Inquisitive Question Generation (InqQG) (Eli5~\citep{fan-etal-2019-eli5}), Headline Generation with Constraint (HGen) (Gigaword~\citep{graff2003english,Rush_2015}), Covid QA (CQA) (COVID-QA~\citep{moller-etal-2020-covid}), and Twitter Stylometry (TwSt) (Tweets Dataset~\citep{DVN/JBXKFD_2017}).

Note: We retrieve the data for COVID-fact from~\citep{scialom2022finetuned}'s existing codebase. We reference it using \citep{scialom2022finetuned} due to a lack of other citation in the paper.
\subsection{Image and Language Datasets}
\label{general:tasks}
 In order to measure and mitigate forgetting, we fine-tune our models on both a stream of image classification tasks, and 3 distinct language domains: Law, Math and Code. \\
In our classical continual learning setting, we construct a stream of 20 tasks from Food-101 \citep{bossard14}, as well as a stream of 20 tasks from CIFAR-100 \citep{Krizhevsky09learningmultiple}. For Food-101, we construct our tasks by grouping 5-labels together for all labels except 100. For CIFAR-100, we group 5-labels together for all labels.\\
 For each language domain, we fine-tune our model on a dataset featuring domain-specific knowledge, %
 as well as unique instruction tasks. 
 For Law, we combine CaseHOLD \citep{zheng2021does}, Terms of Service (ToS) \citep{Lippi_2019, tos_binary}, and Overruling \citep{zheng2021does} to create a more general Law dataset. 
 For Math, we use MetaMathQA \citep{yu2023metamath}, and for Code we use MagiCoder110k \citep{wei2023magicoder}. We believe that required task knowledge across these 3 domains is distinct with distinct task formats.
 As such, we purposefully aim to test our models' ability to generalize across a wide range of knowledge to measure the validity of our method under substantial domain shifts.

\begin{table*}[t]
\caption{Evaluation metrics for each task and domain used in our work.}
\label{appendix:eval-metrics}

\begin{center}
\begin{small}

\begin{tabular}{p{0.43\textwidth}p{0.49\textwidth}}
\toprule
Task/Domain & Eval Metric \\
\midrule
Food-101 & Food-101 holdout set \\
CIFAR-100 & CIFAR-100 holdout set \\
Text Simplification (Simpl)  & Text Simplification (Simpl) holdout set \\
Inquisitive Question Generation (InqQG) & Inquisitive Question Generation (InqQG) holdout set\\
Twitter Stylometry (TwSt) & Twitter Stylometry (TwSt) holdout set\\
Headline Generation with Constraint (HGen) & Headline Generation with Constraint (HGen) holdout set\\
COVID-fact & COVID-fact holdout set \\
Covid QA (CQA) & Covid QA (CQA) holdout set \\
Law & CaseHOLD, ToS, Overruling holdout sets \\
Math & GSM8K~\citep{cobbe2021training} \\
Code & HumanEval~\citep{chen2021evaluating} \\
\bottomrule
\end{tabular}

\end{small}
\end{center}

\end{table*}
\subsection{Bayesian Interpretation}
\label{appendix:bayesian_l2}
{
An isotropic Gaussian prior $\theta\sim\mathcal{N}(\theta_0,\tau^2 I)$ motivates anchoring to $\theta_0$. For a Gaussian regression likelihood $y\mid X,\theta\sim\mathcal{N}(f(X,\theta),\sigma^2 I)$, with fixed variances $\sigma^2,\tau^2>0$,
\begin{equation}
\begin{aligned}
p(\theta\mid X,y)&\propto\exp\Bigl[-\frac{\|y-f(X,\theta)\|_2^2}{2\sigma^2}\\
&\qquad-\frac{\|\theta-\theta_0\|_2^2}{2\tau^2}\Bigr].
\end{aligned}
\end{equation}
The MAP estimate therefore minimizes
\begin{equation}
\|y-f(X,\theta)\|_2^2+\frac{\sigma^2}{\tau^2}\|\theta-\theta_0\|_2^2.
\end{equation}

}
\subsection{L2 distance and SFA}
\label{appendix:l2_distance_acc}
\begin{figure}[t]
     \begin{center}
    \includegraphics[width=\linewidth]{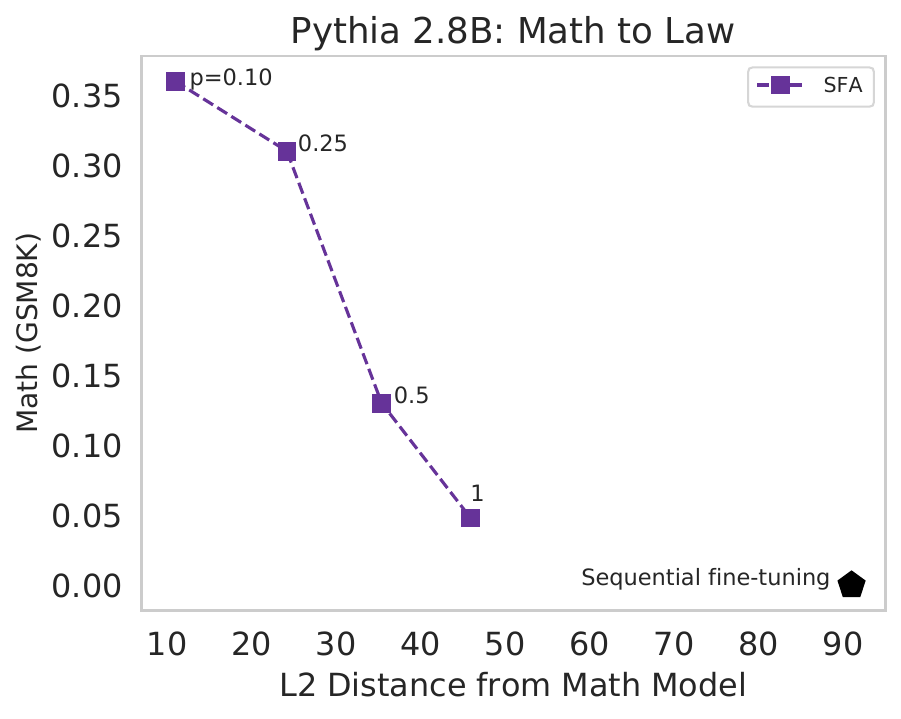}
     \caption{An analysis of the negative correlation between accuracy on Math and the $L2$ distance of the final model (fine-tuned on Math, then Law) from the original model (fine-tuned on Math only). The fine-tuning on Law is done using SFA with varying values of $p$ that determine the merging frequency. For reference we also mark sequential fine-tuning which leads to much higher L2 distance due to no merging, and accuracy just below that achieved with SFA merging once at the end of fine-tuning on law ($p=1$).} 
     \label{fig:pythia2.8b_l2distance}
     \end{center}
\end{figure}

To further explore this intuition of SFA and its relation to constraining parameter weights in the context of our experimental setup, we also show how accuracy and L2 distance are correlated when fine-tuning on different language domains. We use the setup described in~\cref{fig:pythia2.8b} where our model first fine-tunes on Math, then Law. As~\cref{fig:pythia2.8b_l2distance} shows, when proportion of fine-tuning before averaging $p$ decreases on SFA (purple curve), the L2 distance to the initial Math model decreases, while the accuracy on Math increases. This is in direct contrast to sequential fine-tuning without intervention (black pentagon), because of its much higher L2 distance to the initial model. As such, $p$ directly relates to L2 distance, as well as performance on previous tasks, because averaging frequency constrains how much model parameters can change from their initial positions. The values for this figure can be found in~\cref{table:pythia-math-law-l2}.

\subsection{EWC Approximated by Model Merging}
\label{appendix:ewc_model_merging_fisher}

Consider fine-tuning a model with an EWC penalty \citep{Kirkpatrick_2017} where $\theta_{0}$ and $\theta_t$ are the weights of the initial and fine-tuning model respectively. Let $\eta>0$ be the learning rate and $F_0$ the fixed diagonal Fisher estimate, with $F_0^{(j)}\geq0$. We set the penalty coefficient $\lambda=1$ for this derivation.
\begin{equation}
L(\theta_{t}) = L_\text{task}(\theta_{t}) + \sum_{j} \frac{1} {2} F_{0}^{(j)} ( \theta_{t}^{(j)} - \theta_{0}^{(j)})^{2}. \\
\end{equation}
Consider a split update: first take a gradient step on the task loss, then a separate gradient step on the EWC penalty evaluated at the updated parameters:
\begin{align}
\theta_{t+1}^{*} &= \theta_{t} - \eta \nabla_{\theta_t} L_\text{task}.\\
\theta_{t+1} &= (I - \eta F_{0}) \theta_{t+1}^* + \eta  F_{0}\theta_{0}.
\end{align}

Thus a separate explicit gradient step on the EWC penalty is a coordinate-wise interpolation when $0\leq\eta F_0^{(j)}\leq1$. This is not the same as one gradient step on the combined loss at $\theta_t$: the split update differs from that step by $+\eta^2 F_0\nabla_{\theta_t}L_{\mathrm{task}}$. The interpretation therefore concerns the separate correction, rather than an exact equivalence to ordinary EWC training.
This is reminiscent of Fisher model merging from \citet{matena2022merging} where merging an initial and fine-tuning model has the form:

\begin{gather}
\theta^{*(j)} = \frac{\lambda_{0} F_0^{(j)} \theta_{0}^{(j)} + \lambda_{t} F_t^{(j)} \theta_{t}^{(j)}} {\lambda_{0} F_0^{(j)} + \lambda_{t} F_t^{(j)}}.\\
\text{which can be rewritten as:}\notag\\
\theta^{*(j)} = \begin{aligned}[t]
&\left(1- \frac{\lambda_{0} F_0^{(j)}} {\lambda_{0} F_0^{(j)} + \lambda_{t} F_t^{(j)}}\right) \theta_{t}^{(j)}\\
&+ (\frac{\lambda_{0} F_0^{(j)}}{\lambda_{0} F_0^{(j)} + \lambda_{t} F_t^{(j)}}) \theta_{0}^{(j)}.
\end{aligned}
\end{gather}

For nonnegative weights and diagonal Fisher entries, this is a convex combination wherever $\lambda_0F_0^{(j)}+\lambda_tF_t^{(j)}>0$. If this denominator is zero, the formula does not determine that coordinate and a separate fallback is required. Unlike the EWC approximation, this uses the Fisher information of both the initial and current model for merging.

\subsection{Forgetting under Sequential fine-tuning}
\label{appendix:forgetting_results}

\begin{figure*}[t]
    \begin{center}
    \includegraphics[width=\linewidth]{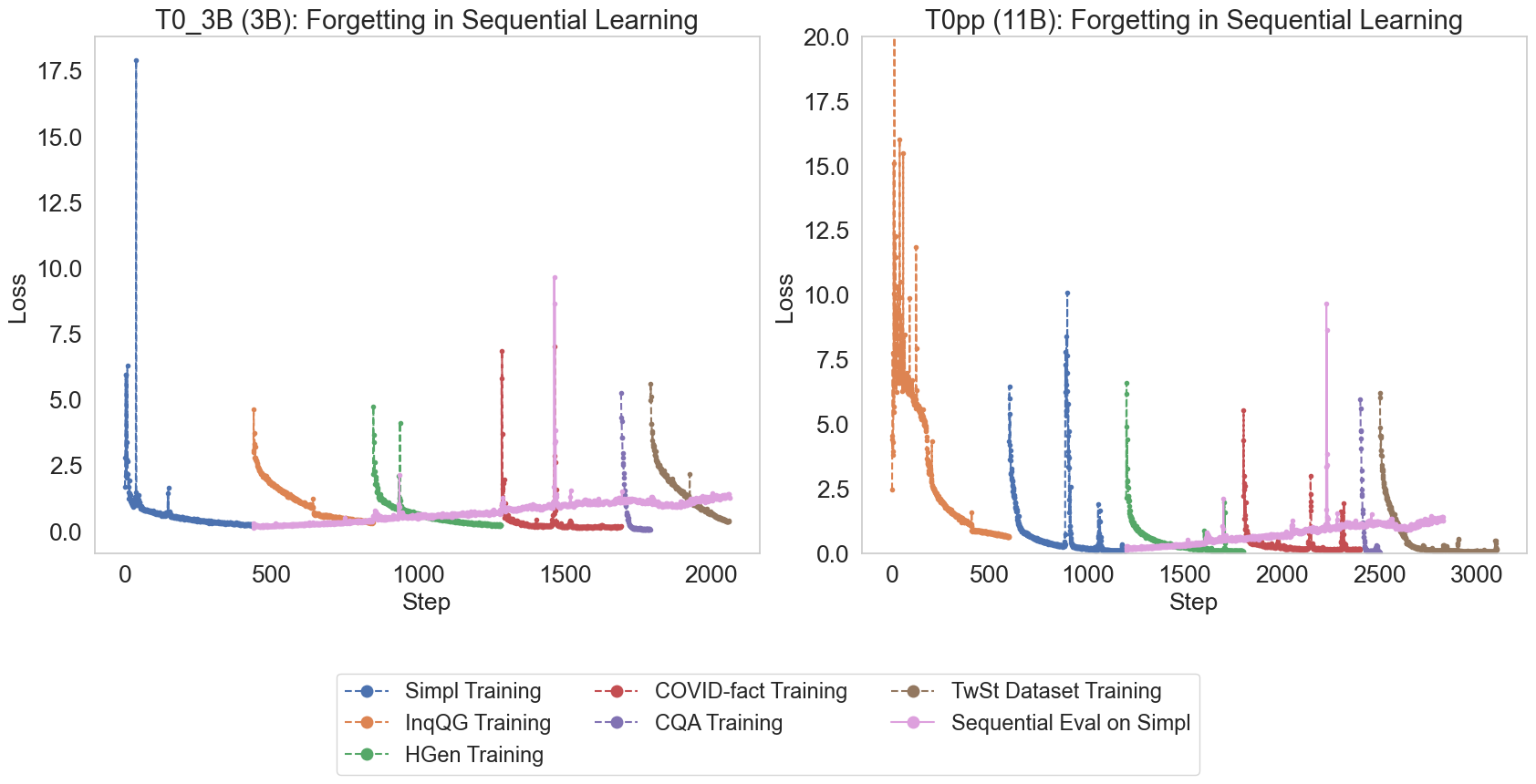}
    \caption{The fine-tuning of T0\_3B (3B) and T0pp (11B) on a stream of language generation tasks. Training loss on each subsequent task decreases as the model learns it, while evaluation loss on Simpl continues to increase, indicating that forgetting is present.}
    \label{fig:t03b_full}
    \end{center}
\end{figure*}

We confirm that fine-tuning on a sequence of different tasks leads to performance degradation on previously learned tasks. 
This forgetting phenomenon occurs across different task domains and for different model sizes. 
In this work, we focus on catastrophic forgetting of capabilities acquired during instruction fine-tuning instead of base pretrained model capabilities.
This is because, as we will show, forgetting of skills learned during instruction fine-tuning can be quite severe and experiments at this scale are more feasible. 
We fine-tune our models on a sequence of instruction, language generation datasets that test general knowledge to measure forgetting. Specifically, we use~\citet{scialom2022finetuned}'s: Text Simplification (Simpl), Inquisitive Question Generation (InqQG), Headline Generation with Constraint (HGen), COVID-fact, Covid QA (CQA), and Twitter Stylometry (TwSt). Many of these tasks incorporate existing datasets which we describe in~\cref{instruction-datasets}.\\
In our first experiments, we fine-tune the T0\_3B (3B) and T0pp (11B) models (see ~\cref{model_descriptions} for model descriptions) on the sequence of tasks described in \cref{general:tasks} while measuring forgetting on the first task.
The results are shown in \cref{fig:t03b_full}. 
The model is first trained on Simpl which leads to a decrease in validation loss shown in blue. 
Subsequently, the model is trained on a sequence of other tasks; the decrease in validation loss on these tasks is shown in different colors.
During this process, we continue to monitor the validation loss on Simpl, displayed in pink.
As models fine-tune on new tasks, their performance on Simpl consistently declines as loss increases.
Forgetting occurs at both 3B and 11B in these sequences (\cref{fig:t03b_full}).

\begin{figure}[t]
     \begin{center}
    \includegraphics[width=\linewidth]{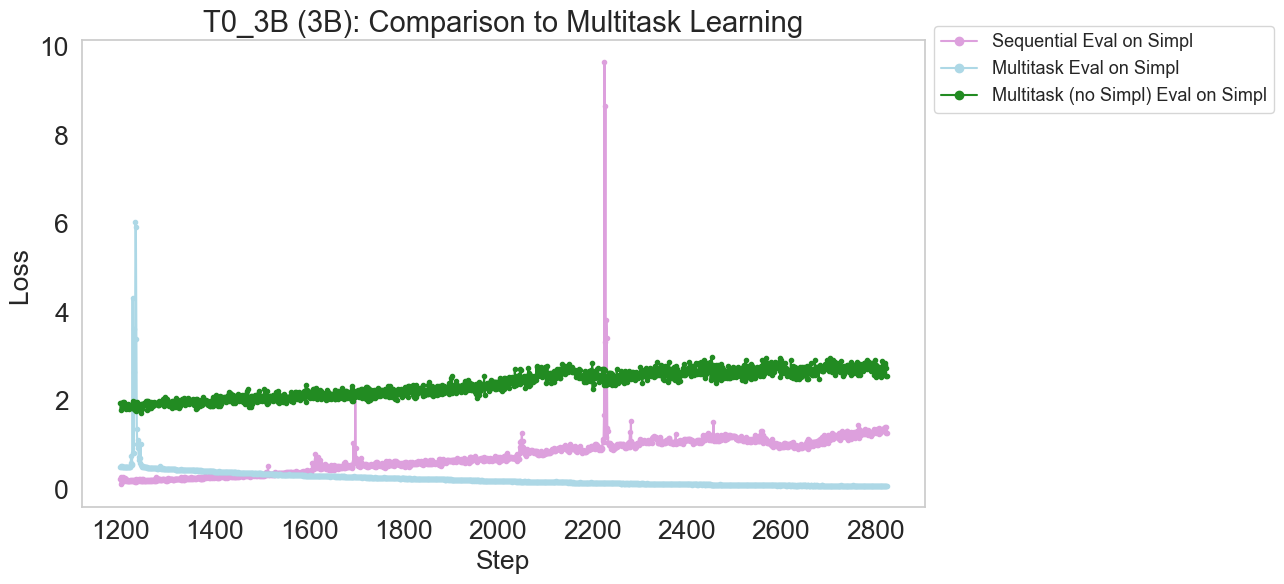}
     \caption{The Simpl loss curve of T0\_3B (3B) from~\cref{fig:t03b_full} is compared to a multitask model training on all tasks, and a multitask model training on all tasks except Simpl. As T0\_3B (3B) continues to fine-tune on each new task, the loss on Simpl becomes the same order of magnitude as that of a model that is never exposed to Simpl.}
     \label{fig:t0_wikiauto11b}
     \end{center}
\end{figure}

But how severe is this forgetting?
We quantify this by comparing a model that was trained on and has then forgotten Simpl to a model that has never seen Simpl.
In \cref{fig:t0_wikiauto11b}, the pink line shows validation loss on Simpl for a model trained on a sequence of fine-tuning tasks starting with Simpl. 
As the model learns new tasks, its performance deteriorates.
After 2000 steps, the sequentially fine-tuned model's loss on Simpl is the same order of magnitude as that of the multitask model trained on all tasks except Simpl.
Thus if a model that has learned Simpl is finetuned on other tasks for as little as 2000 steps, its performance degrades to that of a model that has never seen Simpl.
This indicates significant forgetting, as the model loses the ability to respond to tasks it previously was able to. 

\subsection{Table Results of Domain Fine-tuning}
\begin{table*}[t]
\caption{Results of Pythia (2.8B) models fine-tuning on Math and Law}
\label{table:pythia-math-law}

\begin{center}
\begin{small}

\begin{tabular}{p{0.49\textwidth}ccccc}
\toprule

Pythia (2.8B) & case\_hold & tos &overruling  & GSM8K (0-shot) \\
\midrule
Pythia (2.8B) Original & 0.25 & 0.85 & 0.45 & 0\\
metamathqa    & 0.19 & 0.87 & 0.48 & 0.38\\
law         & 0.74 & 0.93 & 0.97 & 0 \\
Sequential fine-tuning: metamathqa, Law & 0.76 & 0.95 & 0.59 & 0\\
Endpoint interpolation: metamathqa, Law  (p=1) & 0.74 & 0.88 & 0.49 & 0.048 \\
Endpoint interpolation: metamathqa, Law  (p=1, 0.75 law, 0.25 math) & 0.78 & 0.93 & 0.52 & 0\\ %
Endpoint interpolation: metamathqa, Law  (p=1, 0.25 law, 0.75 math) & 0.42 & 0.87 & 0.49 & 0.27\\ %
SFA: metamathqa, Law  (p=0.5)& 0.69& 0.89 & 0.52 & 0.13\\
SFA: metamathqa, Law  (p=0.25)& 0.67 & 0.87 & 0.49 & 0.31\\
SFA: metamathqa, Law  (p=0.10) & 0.59 & 0.87 & 0.49 & 0.36\\ %
Task Arithmetic (0.5 Law, 0.5 math) & 0.68 & 0.87 & 0.55 & 0 \\
Task Arithmetic (0.75 law, 0.25 math) & 0.73 & 0.88 & 0.95 & 0 \\ 
Task Arithmetic (0.25 law, 0.75 math) & 0.30 & 0.87 & 0.49 & 0.24 \\  %
Multitask fine-tuning & 0.76 & 0.87 & 0.58 & 0.40\\
Replay (full history, 10\%) & 0.72 & 0.93 & 0.54 & 0.33\\
\bottomrule
\end{tabular}

\end{small}
\end{center}

\end{table*}
\begin{table*}[t]
\caption{The L2 distance of Pythia (2.8B) models from previous checkpoints of models fine-tuning on Math and law}
\label{table:pythia-math-law-l2}

\begin{center}
\begin{small}

\begin{tabular}{p{0.49\textwidth}ccccc}
\toprule

Pythia (2.8B) & L2-Distance \\
\midrule
metamathqa, Law  - metamathqa & 90.99\\
metamathqa, Law  (p=1) - metamathqa &  45.50 \\
metamathqa, Law  (p=0.5) - metamathqa & 35.38\\
metamathqa, Law  (p=0.25) - metamathqa & 24.16\\
metamathqa, Law  (p=0.10) - metamathqa & 10.94\\ %
Task Arithmetic (0.5 Law, 0.5 math)- metamathqa & 101.77 \\
Multitask fine-tuning - metamathqa & 178.96\\
Replay (full history, 10\%) - metamathqa & 82.21\\
\bottomrule
\end{tabular}

\end{small}
\end{center}

\end{table*}

\begin{table*}[t]
\caption{Results of Llama 2 (7B) models fine-tuning on Math and law}
\label{table:llama-math-law}

\begin{center}
\begin{small}

\begin{tabular}{p{0.49\textwidth}ccccc}
\toprule

Llama 7B & case\_hold  & tos  &overruling  & GSM8K (0-shot) \\
\midrule
Llama 2 (7B) Original & 0.32 & 0.13 & 0.49 & 0 \\ %
metamathqa  & 0.21  & 0.38 & 0.49 & 0.42\\ %
law & 0.81 & 0.51 & 0.94 & 0\\ %
Sequential fine-tuning: metamathqa, Law  & 0.64 & 0.86  & 0.93 & 0\\ %
Endpoint interpolation: metamathqa, Law  (p=1) & 0.61 & 0.59 & 0.90 & 0.0015 \\ %
Endpoint interpolation: metamathqa, Law  (p=1, 0.75 law, 0.25 math) & 0.64 & 0.83 & 0.94 & 0 \\ %
Endpoint interpolation: metamathqa, Law  (p=1, 0.25 law, 0.75 math) & 0.55 & 0.16 & 0.79 & 0.096 \\ %
SFA: metamathqa, Law  (p=0.75)&  0.53 & 0.13 & 0.97 & 0\\ %
SFA: metamathqa, Law  (p=0.5)& 0.50 & 0.13 & 0.90 & 0.12\\ %
SFA: metamathqa, Law  (p=0.25)& 0.53 & 0.13 & 0.95 & 0.46\\ %
SFA: metamathqa, Law  (p=0.17)& 0.48 & 0.13 & 0.63 & 0.48\\ %
Task Arithmetic (0.5 law, 0.5 math) & 0.68 & 0.13 & 0.96 & 0 \\ %
Task Arithmetic (0.75 law, 0.25 math) & 0.79 & 0.18 & 0.97 & 0 \\  %
Task Arithmetic (0.25 law, 0.75 math) & 0.44 & 0.13 & 0.56 & 0 \\  %
TIES & 0.37 & 0.13 & 0.61 & 0.014\\ %
Multitask fine-tuning  & 0.86 & 0.27 & 0.97 & 0.54\\ %
Replay (full history, 10\%) & 0.46 & 0.13 &0.96 & 0.49\\ %

\bottomrule
\end{tabular}

\end{small}
\end{center}

\end{table*}

\begin{table*}[t]
\caption{Qwen2.5-1.5B Math$\to$Law accuracy. These single-run configurations illustrate retention-adaptation trade-offs.}
\label{table:qwen-math-law}
\label{tab:legacycompact}

\begin{center}
\begin{small}

\begin{tabular}{p{0.49\textwidth}ccccc}
\toprule

Qwen2.5 1.5B & case\_hold  & tos  &overruling  & GSM8K (0-shot) \\
\midrule
Qwen2.5 (1.5B) Original & 0.47 & 0.28 & 0.90 & 0 \\ %
metamathqa  & 0.26  & 0.19 & 0.69 & 0.60\\ %
law & 0.77 & 0.91 & 0.95 & 0\\ %
Sequential fine-tuning: metamathqa, Law  & 0.78 & 0.92  & 0.94 & 0\\ %
Endpoint interpolation: metamathqa, Law  (p=1) & 0.81 & 0.88 & 0.93 & 0.01 \\ %
SFA: metamathqa, Law  (p=0.5)& 0.77 & 0.89 & 0.94 & 0.12\\ %
SFA: metamathqa, Law  (p=0.25)& 0.79 & 0.89 & 0.94 & 0.35\\ %
Task Arithmetic & 0.77 & 0.88 & 0.96 & 0 \\ %
TIES & 0.57 & 0.88 & 0.94 & 0\\ %
Multitask fine-tuning  & 0.83 & 0.90 & 0.96 & 0.59\\ %
Replay (full history, 5\%) & 0.78 & 0.93 &0.93 & 0.41\\ %
Replay (full history, 10\%) & 0.79 & 0.93 &0.95 & 0.51\\ %

\bottomrule
\end{tabular}

\end{small}
\end{center}

\end{table*}

\begin{table*}[t]
\caption{Results of Pythia (2.8B) models fine-tuning on Math and code}
\label{table:pythia-math-code}

\begin{center}
\begin{small}

\begin{tabular}{p{0.49\textwidth}ccccc}
\toprule
Pythia (2.8B) & HumanEval (5-shot)  & GSM8K (0-shot) \\
\midrule
original Pythia (2.8B) & 0.074 & 0 \\
metamathqa    & 0.0 & 0.38\\
Magicoder-Evol-Instruct-110K & 0.15 & 0 \\
Sequential fine-tuning: metamathqa, Magicoder-Evol-Instruct-110K & 0.13 & 0.01 \\
Endpoint interpolation: metamathqa, Magicoder-Evol-Instruct-110K (p=1) & 0.06 & 0.33 \\
Endpoint interpolation: metamathqa, Magicoder-Evol-Instruct-110K (p=1, 0.3 math, 0.7 Code) & 0.11 & 0.22 \\
Endpoint interpolation: metamathqa, Magicoder-Evol-Instruct-110K (p=1, 0.6 math, 0.4 Code) & 0.037 & 0.34 \\
Endpoint interpolation: metamathqa, Magicoder-Evol-Instruct-110K (p=1, 0.7 math, 0.3 Code) & 0.018 & 0.38 \\
SFA: metamathqa, Magicoder-Evol-Instruct-110K (p=0.5)& 0.061 & 0.33 \\
SFA: metamathqa, Magicoder-Evol-Instruct-110K (p=0.25)& 0.038 & 0.35 \\
Task Arithmetic (0.5 code, 0.5 math) & 0.049 & 0.21\\
Task Arithmetic (0.75 code, 0.25 math) & 0.14 & 0\\
Task Arithmetic (0.25 code, 0.75 math) & 0 & 0.36\\
Multitask fine-tuning & 0.13 & 0.35\\
Replay (full history, 10\%) & 0 & 0.32\\
\bottomrule
\end{tabular}

\end{small}
\end{center}

\end{table*}

\begin{table*}[t]
\caption{Results of Llama 2 (7B) models fine-tuning on Code then Math, as indicated by the row order.}
\label{table:llama-math-code}

\begin{center}
\begin{small}

\begin{tabular}{p{0.49\textwidth}ccccc}
\toprule

Llama 2 (7B) & HumanEval (5-shot)  & GSM8K (0-shot) \\
\midrule
Llama 2 (7B) Original & 0.15 & 0 \\
metamathqa    & 0 & 0.55\\
Magicoder-Evol-Instruct-110K & 0.35  & 0 \\
Sequential fine-tuning: Magicoder-Evol-Instruct-110K, metamathqa & 0.046 & 0.54 \\
Endpoint interpolation: Magicoder-Evol-Instruct-110K, metamathqa (p=1) &  0.18 & 0.49 \\
SFA: Magicoder-Evol-Instruct-110K, metamathqa (p=0.75)& 0.22 & 0.41\\
SFA: Magicoder-Evol-Instruct-110K, metamathqa (p=0.5)& 0.17 & 0.44\\
SFA: Magicoder-Evol-Instruct-110K, metamathqa (p=0.25)& 0.22 & 0.36\\
Task Arithmetic & 0.19 & 0.44\\
TIES & 0.27 & 0.090\\
Multitask fine-tuning & 0.09 & 0.40\\ %
\bottomrule
\end{tabular}

\end{small}
\end{center}

\end{table*}
\begin{table*}[t]
\caption{Results of Pythia (2.8B) models fine-tuning on Math, Law and Code for 2 orders}
\label{table:pythia-math-law-code-3}

\begin{center}
\begin{small}

\setlength{\tabcolsep}{2pt}
\begin{tabular}{p{0.62\textwidth}ccc}
\toprule

Pythia (2.8B) & case\_hold & tos &overruling  \\
\midrule
Pythia (2.8B) Original & 0.25 & 0.85 & 0.45 \\ %
metamathqa    & 0.19 & 0.87 & 0.48 \\
law         & 0.74 & 0.93 & 0.97 \\
Magicoder-Evol-Instruct-110K & 0.22 & 0.28 & 0.52 \\
Sequential fine-tuning: metamathqa, Law, Magicoder-Evol-Instruct-110K & 0.30 & 0.87 & 0.51 \\
Endpoint interpolation: metamathqa, Law, Magicoder-Evol-Instruct-110K (p=1) & 0.50 & 0.88 & 0.59\\
SFA: metamathqa, Law, Magicoder-Evol-Instruct-110K  (p=0.5)& 0.55& 0.88 & 0.57 \\
SFA: metamathqa, Law, Magicoder-Evol-Instruct-110K (p=0.25)& 0.57 & 0.88 & 0.67 \\
Sequential fine-tuning: metamathqa, Magicoder-Evol-Instruct-110K, Law & 0.73 & 0.93 & 0.49 \\
Endpoint interpolation: metamathqa, Magicoder-Evol-Instruct-110K, Law (p=1) & 0.75 & 0.87 & 0.60 \\
SFA: metamathqa, Magicoder-Evol-Instruct-110K, Law (p=0.5) & 0.70 & 0.88 & 0.49 \\
SFA: metamathqa, Magicoder-Evol-Instruct-110K, Law (p=0.25) & 0.68 & 0.88 & 0.51\\
Task Arithmetic (0.33 math, 0.33 Law, 0.33 code) & 0.63 & 0.87 & 0.88\\ 
Multitask fine-tuning & 0.80 & 0.88 & 0.93 \\
Replay (full history, 10\%; Math--Code--Law) & 0.75 & 0.89 & 0.49 \\ 
Replay (full history, 10\%; Math--Law--Code) & 0.69 & 0.89 & 0.56 \\ 
\bottomrule
\end{tabular}
\begin{tabular}{p{0.62\textwidth}cc}
\toprule

Pythia (2.8B) & GSM8K (0-shot) & HumanEval (5-shot) \\
\midrule
Pythia (2.8B) Original & 0 & 0.0\\ %
metamathqa    & 0.38 & 0\\
law          & 0 & 0 \\
Magicoder-Evol-Instruct-110K &  0 & 0.15 \\
Sequential fine-tuning: metamathqa, Law, Magicoder-Evol-Instruct-110K & 0.01 & 0.14\\
Endpoint interpolation: metamathqa, Law, Magicoder-Evol-Instruct-110K (p=1) & 0.34 & 0.068 \\
SFA: metamathqa, Law, Magicoder-Evol-Instruct-110K  (p=0.5) & 0.37 & 0.051\\
SFA: metamathqa, Law, Magicoder-Evol-Instruct-110K (p=0.25) & 0.39 & 0.055\\
Sequential fine-tuning: metamathqa, Magicoder-Evol-Instruct-110K, Law & 0.0 & 0.00\\
Endpoint interpolation: metamathqa, Magicoder-Evol-Instruct-110K, Law (p=1) & 0.011 & 0\\
SFA: metamathqa, Magicoder-Evol-Instruct-110K, Law (p=0.5) & 0.054 & 0\\
SFA: metamathqa, Magicoder-Evol-Instruct-110K, Law (p=0.25)  & 0.30 & 0.0012\\
Task Arithmetic (0.33 math, 0.33 Law, 0.33 code) & 0 & 0.01 \\ 
Multitask fine-tuning & 0.38 & 0.22 \\
Replay (full history, 10\%; Math--Code--Law) & 0.30 & 0.029\\ 
Replay (full history, 10\%; Math--Law--Code) & 0.30 & 0.15\\ 
\bottomrule
\end{tabular}

\end{small}
\end{center}

\end{table*}

\FloatBarrier
\subsection{Comparing Storage Overhead}
\label{app:storage_compare}

SFA preserves information from earlier tasks through a reference checkpoint rather than a collection of training examples. Its additional storage therefore depends on model size, not on the number or size of past datasets. This offers a predictable storage budget throughout continual learning and becomes advantageous when retaining a substantial collection of past examples, particularly for high-resolution inputs.

Let $P$ be the number of model parameters, $b_w$ the bytes per stored parameter, and $D(M)$ the storage required for $M$ retained examples, including labels or stored logits when applicable. For full-parameter models, excluding common optimizer states and activations, the storage requirements are:
\begin{center}
\begin{tabular}{@{}p{0.28\linewidth}p{0.32\linewidth}p{0.29\linewidth}@{}}
\toprule
Method & Total parameter/data storage & Extra over one model \\
\midrule
SFA & $2Pb_w$ & $Pb_w$ \\
Replay (buffer) & $Pb_w+D(M)$ & $D(M)$ \\
SFA-R & $2Pb_w+D(M)$ & $Pb_w+D(M)$ \\
CLS-ER & $3Pb_w+D(M)$ & $2Pb_w+D(M)$ \\
\bottomrule
\end{tabular}
\end{center}

SFA requires less additional storage than replay whenever the retained examples occupy more space than its reference checkpoint, $D(M)>Pb_w$. For ViT-Base with 86 million parameters stored in FP16, this checkpoint occupies approximately 172 MB. By comparison, 2,000 uncompressed $224\times224$ RGB images stored as uint8 pixels occupy approximately 301 MB, before labels and metadata. In this example, SFA reduces additional storage by approximately 43\%, while retaining no training examples. Relative to CLS-ER, SFA removes both an auxiliary parameter copy and the replay buffer, reducing retained-state storage by $Pb_w+D(M)$.

\begin{figure*}[t]
     \begin{center}
    \includegraphics[clip,width=\linewidth]{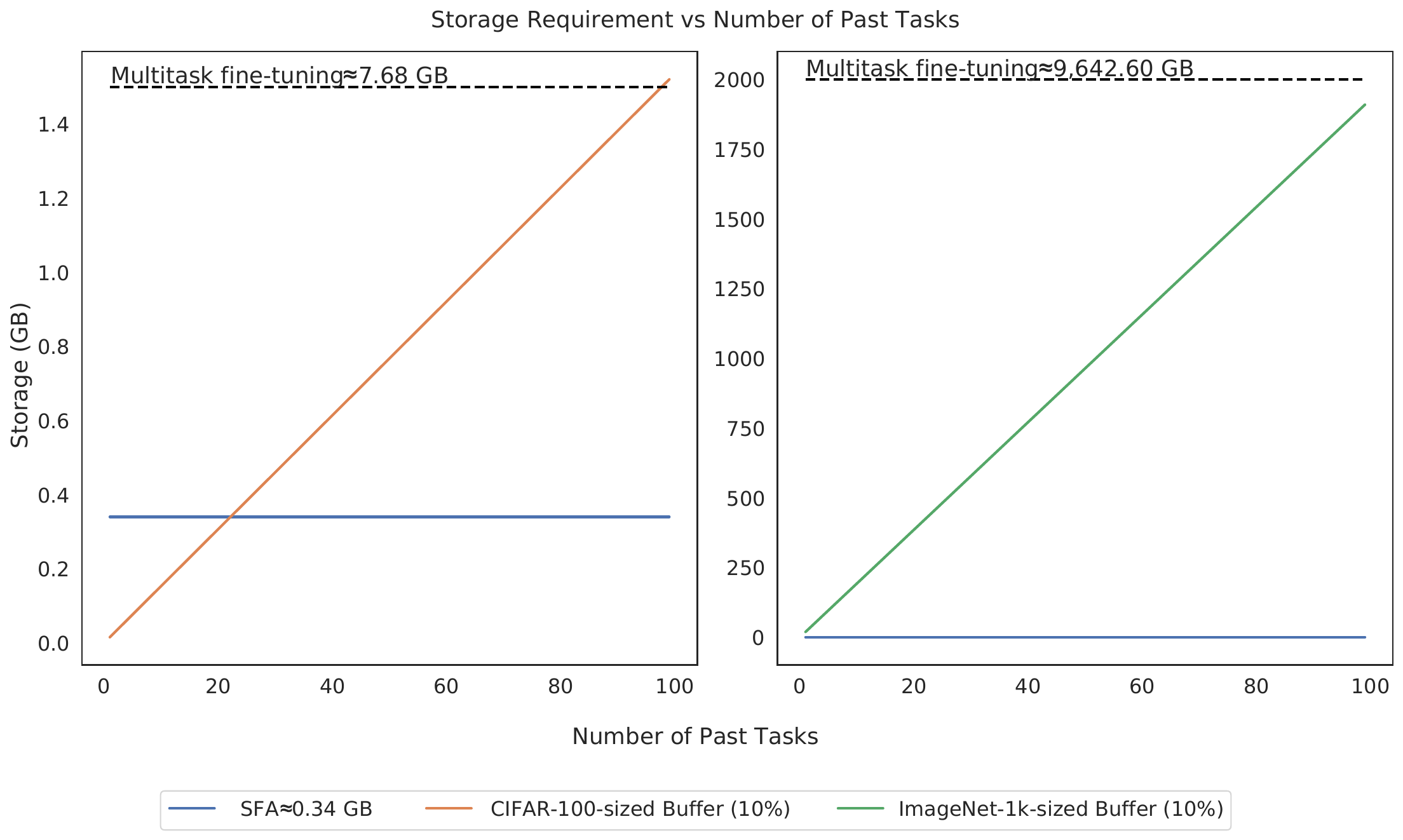}
     \caption{
SFA's constant parameter-storage requirement compared with growing replay collections in hypothetical CIFAR-100-sized (\textbf{left}) and ImageNet-1k-sized (\textbf{right}) task streams. Replay retains 10\% of each task without a capacity limit. The SFA curve includes both working and reference parameters, while replay curves show retained data only; adding replay's working model would increase its total storage. Images are assumed uncompressed, and these analytical examples are distinct from the experimental task streams. 
}
     \label{fig:memory_usage}
     \end{center}
\end{figure*}

The advantage can also grow across a task stream when replay retains a proportion of each task's data. In the illustrative setting of \cref{fig:memory_usage}, retaining 10\% of each CIFAR-100-sized dataset adds 15.36 MB per task, exceeding the 172 MB reference after 12 tasks. SFA's reference storage remains unchanged. These calculations illustrate the benefit of bounded parameter storage when the retained data collection grows; they use hypothetical whole-dataset tasks rather than the five-class tasks in our vision experiments.

Replay can instead enforce a fixed buffer capacity, and small buffers or compressed inputs may occupy less space than the reference. SFA's storage advantage is therefore most relevant to larger retained datasets and approaches that maintain multiple model copies alongside replay. In adapter training, SFA further limits its reference to the trainable adapter parameters, leaving the shared frozen backbone unchanged.

Beyond storage, SFA requires only parameter interpolation at each merge, without additional forward or backward passes over past examples. Under matched current-task exposure, it avoids the additional rehearsal work required by replay-based methods.

\begin{figure}[t]
     \begin{center}
    \includegraphics[clip,width=\linewidth]{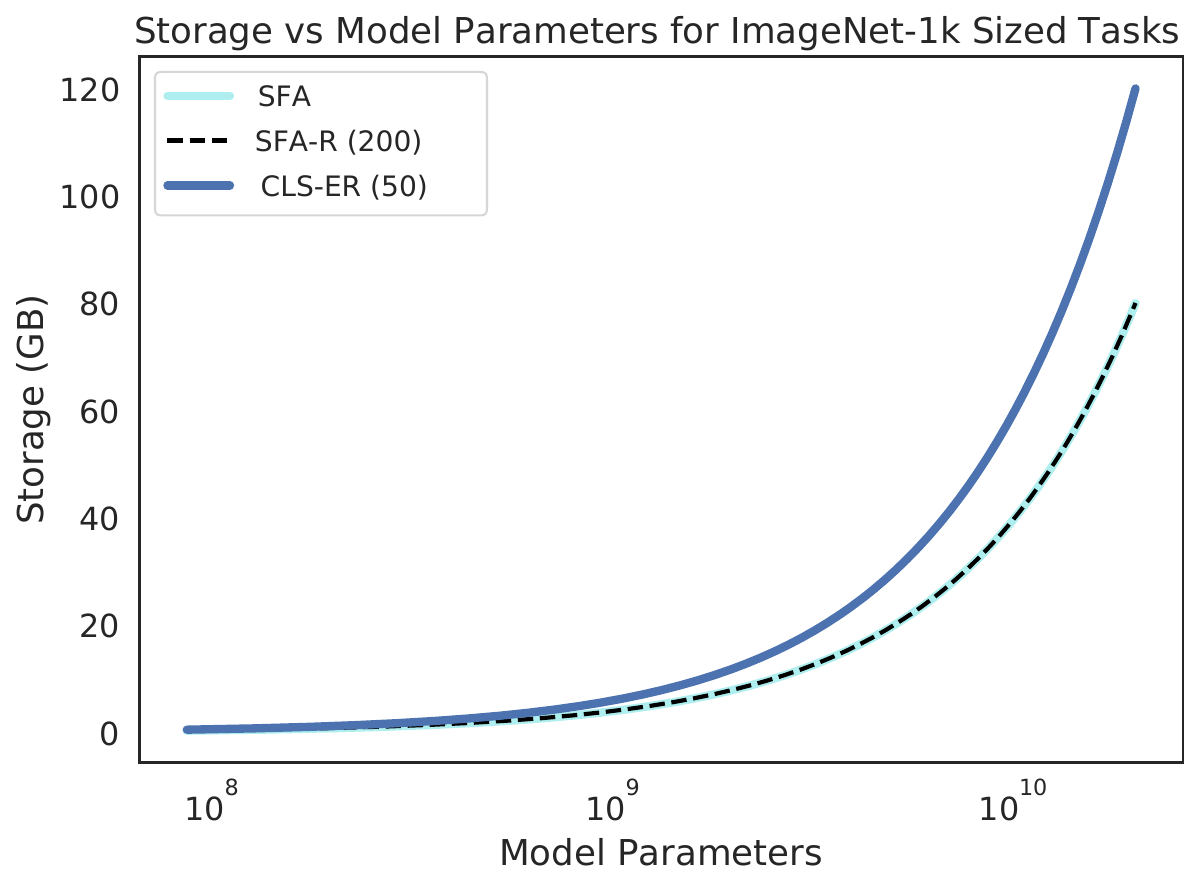}
     \caption{
Analytical parameter-plus-buffer storage for SFA and CLS-ER with FP16 parameters and uncompressed $224\times224$ RGB images stored as uint8 pixels. Parenthesized legend values denote the number of retained images. SFA requires one reference alongside the working model, whereas CLS-ER maintains two auxiliary model copies and a replay buffer. 
}
     \label{fig:memory_params}
     \end{center}
\end{figure}

\FloatBarrier
\subsection{SFA Without Task Boundaries}

The main SFA rule assumes known training-task boundaries. We also evaluate a scheduled-anchor variant, SFA-WTB, that averages every $C$ optimizer steps and refreshes the anchor every $A$ steps, without consulting task labels. When both events coincide, the working model is averaged first and the anchor is then replaced by that result (\cref{alg:sfpa}). This introduces two schedules rather than eliminating schedule selection. In the special case $A=C$, its parameter updates reduce to the slow/fast interpolation pattern of Lookahead, with interpolation coefficient $1-\beta$ and the same inner optimizer~\citep{zhang2019lookahead} and synchronization schedule; optimizer-state handling must also match for full implementation equivalence.

\begin{algorithm}
\caption{Sequential Fine-tuning with Averaging (Without Task Boundaries) (SFA-WTB)}\label{alg:sfpa}
{
\begin{algorithmic}[1]
\REQUIRE Initial weights $\theta_0$, positive integer intervals $C,A$, learning rate $\alpha>0$, weight $0\leq\beta\leq1$, positive integer updates $T$
\STATE $a\gets\theta_0$
\FOR{$t=0,\ldots,T-1$}
\STATE $\theta_{t+1}^{*}\gets\theta_t-\alpha\nabla_{\theta_t}L_t$
\STATE $\theta_{t+1}\gets\theta_{t+1}^{*}$
\IF{$(t+1)\bmod C=0$}
\STATE $\theta_{t+1}\gets\beta a+(1-\beta)\theta_{t+1}^{*}$
\ENDIF
\IF{$(t+1)\bmod A=0$}
\STATE $a\gets\theta_{t+1}$
\ENDIF
\ENDFOR
\RETURN $\theta_T$
\end{algorithmic}
}
\end{algorithm}

Here $a$ denotes the changing reference, distinct from the initial weights $\theta_0$. The periodic schedule is unchanged: unlike Algorithm~\ref{alg:sfa}, this variant does not force an extra merge at $T$ when $T$ is not a multiple of $C$.

\begin{figure*}[t]
\centering
         \includegraphics[clip,width=0.49\textwidth]{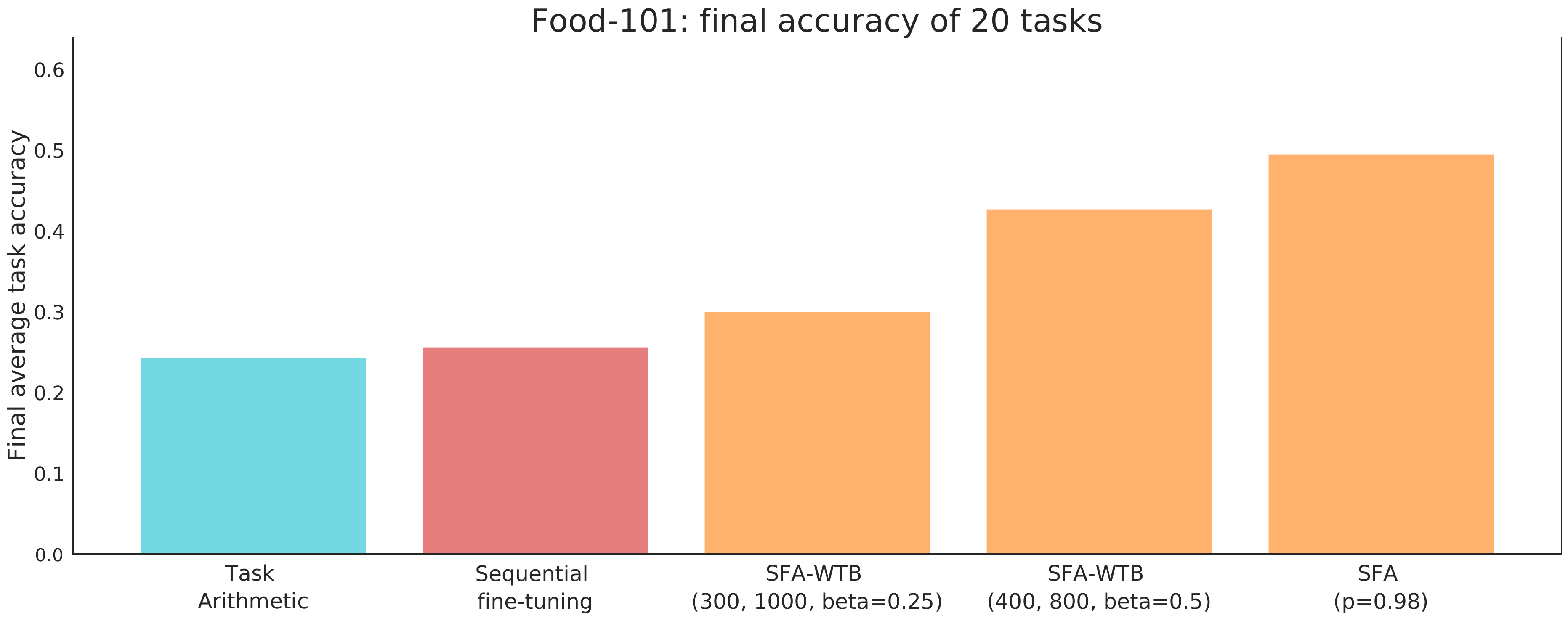}
         \includegraphics[clip,width=0.49\textwidth]{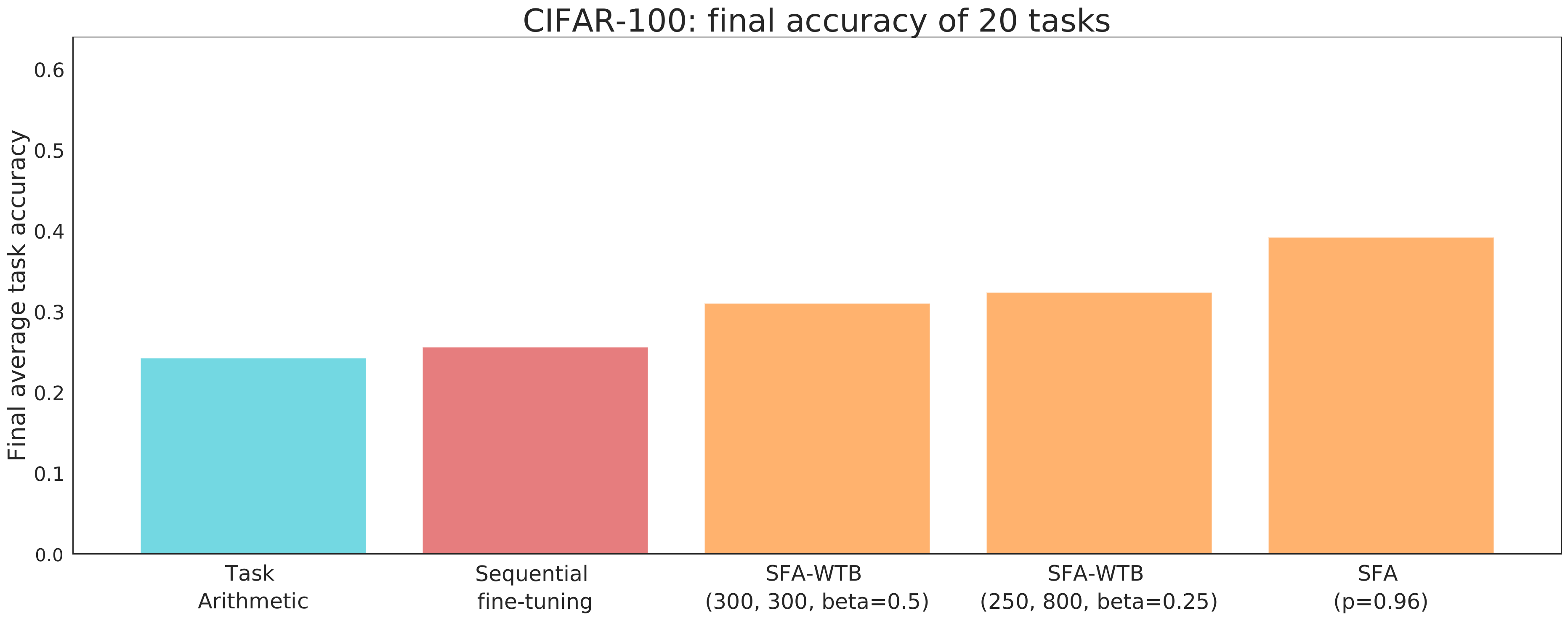}
     \caption{A comparison of SFA with and without using task boundaries to control model averaging and updating frequencies on a sequence of 20 tasks from Food-101 (\textbf{left}) and CIFAR-100 (\textbf{right}). While SFA is still effective without access to task boundaries, their incorporation enhances performance.}
     \label{fig:cv_sfawtb}
\end{figure*}

\cref{fig:cv_sfawtb} evaluates SFA-WTB on the same 20-task vision sequences after sweeping $C$, $A$, and $\beta$. In these experiments the boundary-aware version performs better. A plausible explanation is that completed-task anchors preserve earlier skills more effectively than arbitrarily scheduled anchors.